\documentclass[preprint, 12pt, authoryear]{elsarticle}

\usepackage[utf8]{inputenc}
\usepackage[T1]{fontenc}
\usepackage{lmodern}
\usepackage{microtype}

\usepackage{amsmath,amssymb,amsfonts}
\usepackage{textcomp}

\usepackage{graphicx}
\usepackage{xcolor}
\usepackage{array}
\usepackage{booktabs}
\usepackage{longtable}
\usepackage{tabularx}
\usepackage{ltablex}
\keepXColumns
\usepackage{multirow}
\usepackage{makecell}
\usepackage{calc}
\usepackage{enumitem}
\usepackage{verbatim}

\usepackage[colorlinks=true,
            linkcolor=blue!50!black,
            citecolor=blue!50!black,
            urlcolor=blue!50!black,
            breaklinks=true]{hyperref}
\usepackage{url}
\usepackage{seqsplit}

\PassOptionsToPackage{hyphens}{url}
\makeatletter
\g@addto@macro\UrlBreaks{\do\/\do\-\do\_\do\.\do\=\do\?\do\&\do\+\do\;\do\:}
\makeatother
\providecommand{\tightlist}{\setlength{\itemsep}{0pt}\setlength{\parskip}{0pt}}
\providecommand{\textquotesingle}{\char13}
\providecommand{\labelenumi}{(\arabic{enumi})}
\providecommand{\ul}[1]{\underline{#1}}

\DeclareUnicodeCharacter{00FC}{\"u}
\DeclareUnicodeCharacter{00E9}{\'e}
\DeclareUnicodeCharacter{00FA}{\'u}

\journal{Computers, Environment and Urban Systems}

\begin{document}

\begin{frontmatter}

\title{NORA: A Harness-Engineered Autonomous Research Agent for Spatial Data Science}

% --- Authors and affiliations ----------------------------------------------
\author[utk]{Bing Zhou\corref{cor1}}
\ead{bzhou11@utk.edu}
\author[emory]{Xiao Huang}
\author[emory]{Huan Ning}
\author[utk]{Qiusheng Wu}
\author[tamu]{Diya Li}
\author[tiktok]{Ziyi Zhang}

\cortext[cor1]{Corresponding author.}

\affiliation[utk]{organization={Department of Geography and Sustainability,
            University of Tennessee, Knoxville},
            city={Knoxville}, state={TN}, country={USA}}
\affiliation[emory]{organization={Department of Environmental Science,
            Emory University},
            city={Atlanta}, state={GA}, country={USA}}
\affiliation[tamu]{organization={Department of Geography,
            Texas A\&M University},
            city={College Station}, state={TX}, country={USA}}
\affiliation[tiktok]{organization={TikTok},
            city={San Jose}, state={CA}, country={USA}}

% --- Abstract --------------------------------------------------------------
\begin{abstract}
The automation of scientific research workflows has emerged as a transformative
frontier in artificial intelligence, yet existing autonomous research agents
remain largely domain-agnostic, lacking the specialized reasoning, method
selection, and data acquisition capabilities required for rigorous spatial
data science. This paper introduces NORA (Night Owl Research Agent), a
harness-engineered, multi-agent autonomous research system purpose-built for
GIScience and spatial data science. NORA orchestrates the complete research
lifecycle through a skills-first architecture comprising 21 domain-specialized
workflow skills, 9 specialist sub-agents, and custom Model Context Protocol
(MCP) servers. Central to the system's design are two novel domain-specialized
skills: a spatial analysis skill unit that encodes decision frameworks for
exploratory spatial data analysis, spatial regression, and diagnostics; and
a spatial data download skill that supports reproducible acquisition from
authoritative geospatial data sources. We formalize the concept of harness
engineering for scientific research agents, demonstrating how lifecycle hooks,
safety gates, generator--evaluator separation, human-in-the-loop, and state
persistence ensure reliable and reproducible autonomous research. We evaluate
NORA through case studies by 6 domain specialists and 3 LLM reviewers across
seven dimensions (novelty, quality, rigor, etc.). Results demonstrate that
domain-specialized harness engineering substantially improves the efficiency
and quality of research output compared to general-purpose agent
configurations.
\end{abstract}

% --- Keywords --------------------------------------------------------------

\end{frontmatter}

% =============================================================================
%  Body of the manuscript (converted from Manuscript_arxiv.docx).
% =============================================================================

\section{Introduction}\label{introduction}

The past decade has witnessed an accelerating convergence between artificial intelligence and geographic information science, giving rise to a vibrant research domain variously termed GeoAI (Janowicz et al. 2020), spatial data science (Singleton and Arribas-Bel 2021). This convergence has produced powerful tools for spatial prediction and geospatial knowledge extraction. Yet the process of conducting spatial research itself, which contains formulating hypotheses, acquiring geospatial data, selecting spatially appropriate analytical methods, interpreting results in geographic context, and communicating findings through scholarly manuscripts, remains a predominantly manual, labor-intensive endeavor requiring deep domain expertise at every stage.

Recent advances in large language models (LLMs) and autonomous agents have demonstrated the feasibility of automating substantial portions of scientific research. The AI Scientist (Lu et al. 2026) achieved the milestone of generating a manuscript that passed peer review at a top-tier machine learning workshop, while systems such as Mind2Report (Cheng et al. 2026) have shown that cognitive agentic workflows can synthesize expert-level reports from noisy web sources. In parallel, the Geographic Information Science (GIScience) community has begun exploring LLM-powered spatial agents: LLM-Geo (Li and Ning 2023), LLM-Find (Ning et al. 2025), and GIS Copilot (Akinboyewa et al. 2025) have demonstrated that LLMs can generate and execute geoprocessing workflows through natural language.

However, existing autonomous research agents have largely been conceptualized around the norms of computer science and AI. Although its pipeline offers a useful foundation, it lacks domain-specific elements and reasoning that suit the needs of GIScience and spatial data science (Figure 1). In these fields, research is often constrained not only by model design but also by the discovery, acquisition, and assessment of heterogeneous spatial data, whose availability, scale, temporal resolution, and geographic coverage fundamentally shape the feasibility and validity of analysis. General-purpose AI research agents lack the domain-specific reasoning required for rigorous spatial research: not enforce coordinate reference system (CRS) consistency, test for spatial autocorrelation, select spatially appropriate analytical methods---such as geographically weighted regression (Fotheringham et al. 2002) or multiscale GWR (Fotheringham et al. 2017) instead of ordinary least squares when spatial heterogeneity is present---or ensure that geospatial data provenance meets the reproducibility standards increasingly demanded by GIScience journals (Kedron et al. 2021; Nüst and Pebesma 2021). Moreover, GIScience research is also commonly motivated by societally urgent questions involving hazards, health, and environmental change, for which explanatory insight, actionable decision support, and methodological transparency may be as important as predictive accuracy. Lastly, existing GeoAI agents have primarily focused on task-level spatial operations rather than orchestrating the full arc of a research project from conception to comprehensive reporting.

\begin{figure}[htbp]
\centering
\includegraphics[width=\linewidth]{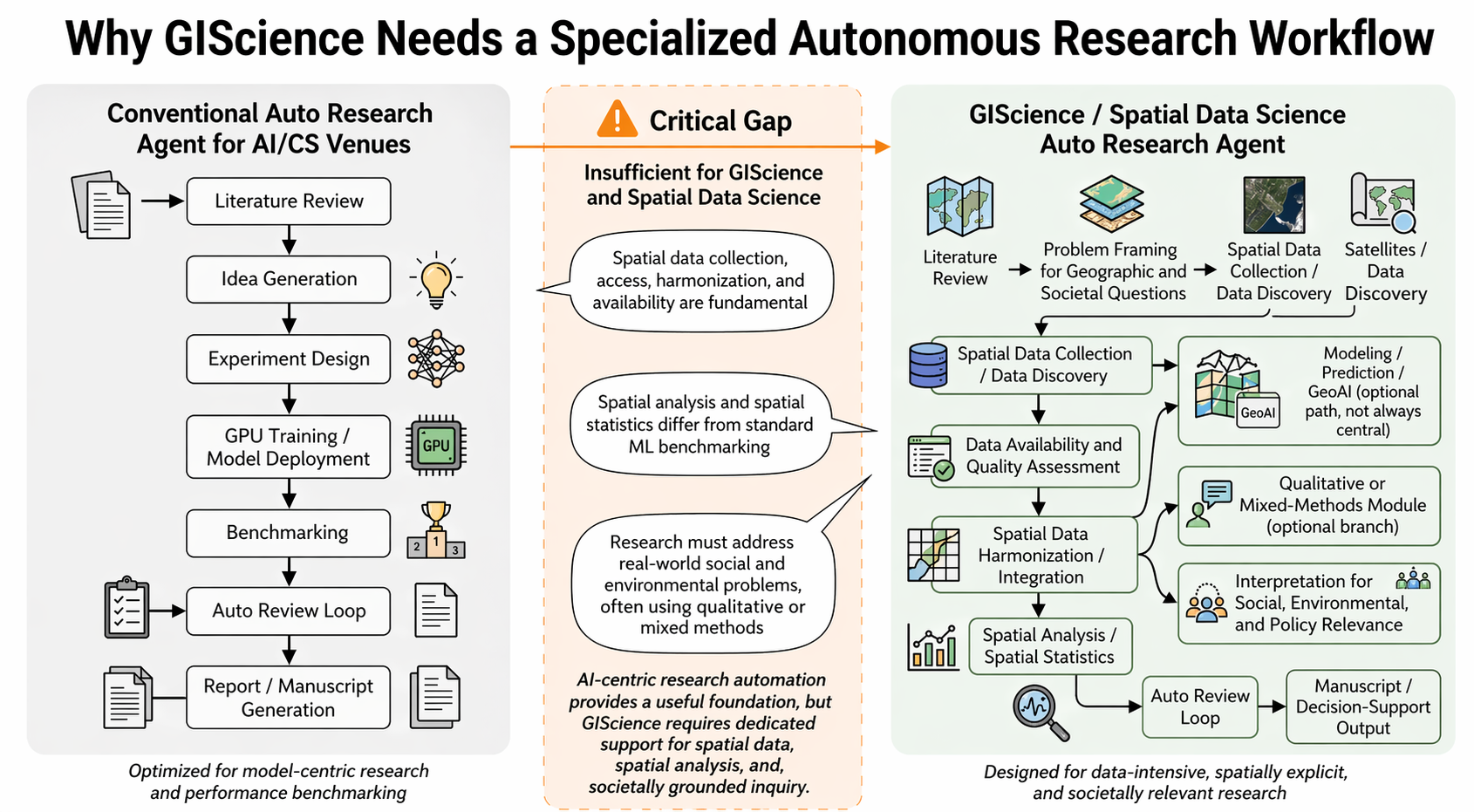}
\caption{Critical gaps between conventional auto research agents for AI venues and GIScience and spatial data science venues.}
\label{fig:1}
\end{figure}

This paper addresses these gaps by introducing NORA (Night Owl Research Agent), a harness-engineered, multi-agent autonomous research system designed specifically for spatial data science research. NORA is not a general-purpose LLM wrapper or a simple coding assistant. It is a domain-specialized research automation platform whose architecture, skill decomposition, and quality assurance mechanisms are purpose-built for the unique demands of GIScience and spatial data science scholarship. The contributions of this paper are as follows:

(1) We propose NORA with a domain-specialized architecture to orchestrate the complete spatial research lifecycle from literature review through manuscript generation. We introduce two novel skill units---a spatial analysis skill for method selection, diagnostics, and spatial reasoning; and a spatial data download skill supporting reproducible acquisition from authoritative geospatial sources---and argue that such domain decomposition is essential for trustworthy automated spatial research.

(2) We formalize the concept of harness engineering in the context of autonomous research systems for spatial data science, demonstrating how lifecycle hooks, structured state persistence, generator-evaluator separation, human-in-the-loop, and safety gates ensure reliable, reproducible, and auditable research workflows.

(3) We evaluate the performance of NORA by conducting three case studies that target the IJGIS publication with various focuses and difficulties. Details are provided on how NORA orchestrates the entire workflow to demonstrate the contributions of each core architectural unit. The final report generated is evaluated by 6 domain experts and 3 LLM reviewers across seven evaluation dimensions to uncover the current capability boundaries of LLM-based auto research agents.

(4) NORA could be the entryway toward standardized workflow and strengthen reproducibility. Once a tool proves reliable and effective, it begins to shape the broader research workflow itself by encouraging more consistent, transparent, and structured practices. In the context of geospatial research, this could support the development of common benchmarking frameworks that allow methods, datasets, and results to be evaluated under shared standards. Equally important, the tool can automatically document each step of the process, including data acquisition, preprocessing, parameter settings, model training, evaluation, and visualization. The automatic record improves the traceability of results and makes it easier for researchers to replicate, verify, and extend the work.

\section{Related Work}\label{related-work}

\subsection{LLMs in Scientific Workflows and Autonomous Research Agents}\label{llms-in-scientific-workflows-and-autonomous-research-agents}

The vision of automating scientific discovery has deep roots, from early systems such as DENDRAL for mass spectroscopy interpretation to Waltz and Buchanan's (2009) articulation of closed-loop automated science. Recent advances in large language models (LLMs) and autonomous agents have brought this vision substantially closer to realization. Systems such as \emph{The AI Scientist} (Lu et al. 2026) demonstrated the feasibility of end-to-end research automation, spanning ideation, experimentation, manuscript writing, and even automated peer review, with reported workshop-level acceptance at ICLR 2025. Its automated reviewer suggests that LLM-based quality assessment is approaching human-level reliability. Other efforts further reinforce this trajectory. \emph{Mind2Report} (Cheng et al. 2026) introduced a cognitive deep research workflow with dynamic memory and multi-dimensional reflection for synthesizing commercial reports, outperforming systems such as OpenAI's o3 Deep Research and Google's Gemini Deep Research in quality, reliability, and coverage.

Beyond full end-to-end autonomy, LLMs have also been embedded into specific components of scientific workflows. Tool-augmented language models such as \emph{Toolformer} (Schick et al. 2024) showed that LLMs can learn to call external resources. In the geospatial domain, \emph{Spatial-Agent} (Bao et al. 2026) formalized geo-analytical question answering as a concept transformation problem using GeoFlow Graphs grounded in spatial information theory, demonstrating how LLM-based agents can be adapted to domain-specific analytical settings. At the same time, the notion of the agent harness---the surrounding infrastructure that makes AI agents reliable, controllable, and robust---has emerged as a critical discipline. Anthropic (2025) formalized effective harness patterns for long-running agents, including lifecycle hooks, context compaction, and tool orchestration, highlighting that agent capability depends not only on model intelligence but also on the engineering systems that structure and constrain its operation.

However, autonomous LLM agents can reason, call tools, and generate research artifacts, yet they often lack rigorous provenance, standardized workflows, and discipline-specific guardrails that conventional scientific workflow systems (e.g. KNIME) offer. Additionally, existing autonomous research systems are also largely domain-agnostic. This generality is both a strength and a limitation. While it enables broad applicability, it cannot by itself ensure the level of domain-specific rigor. It is precisely these gaps that motivate the harness engineering approach formalized in this paper.

\subsection{Autonomous GIS}\label{autonomous-gis}

Recently, a research agenda towards Autonomous GIS has been proposed (Li and Ning 2023), envisioning AI-powered geographic information systems that can independently generate and execute geoprocessing workflows. Following this, several LLM-based spatial agent frameworks have emerged (Li et al. 2025). For example, LLM-Find (Ning et al. 2025) developed an autonomous GIS agent framework for geospatial data retrieval, introducing the "data source handbook" concept, which is structured knowledge package teaching LLMs how to retrieve data from specific sources. GIS Copilot (Akinboyewa et al. 2025) extended this work to integrate LLMs into existing GIS platforms such as QGIS. GeoFlow (Bhattaram et al. 2025) demonstrated agentic workflow automation for geospatial tasks, improving task success by 6.8\% over prior work while reducing token costs. GeoColab (Wu et al. 2025) introduced a multi-agent collaborative framework for geospatial code generation. GeoAnalystBench (Zhang et al. 2025) established a benchmark of 50 Python-based spatial analysis tasks, finding that proprietary models achieve 95\% workflow validity while smaller open-source models achieve only 48.5\%. GeoBenchX (Krechetova and Kochedykov 2025) further benchmarked LLM agents on multistep geospatial tasks, finding that even the best models struggle with complex chains of spatial reasoning.

These GeoAI agents represent important advances in automating spatial operations. However, they focus on task-level execution rather than orchestrating a complete research project. None of them addresses the full lifecycle from research question formulation through peer-reviewed manuscript generation, and none of them incorporates the methodological guardrails that spatial research demands.

\subsection{Reproducibility and Replicability in GIScience}\label{reproducibility-and-replicability-in-giscience}

Reproducibility and replicability (R\&R) have emerged as central concerns in GIScience. Kedron et al. (2021) examined R\&R opportunities and challenges for geospatial research, while Kedron and Li (2024) further analyzed R\&R from a computational and spatial perspective, identifying factors including training data selection, model design uncertainty, and the inherent spatial heterogeneity of geospatial processes. Nüst and Pebesma (2021) highlighted the gap between reproducibility aspirations and practice in computational geography, emphasizing the need for research compendia that include code, data, and containerized computing environments. Leading journals have adopted data and code sharing policies aligned with FAIR principles (Findable, Accessible, Interoperable, Reusable), yet compliance remains inconsistent (Kedron et al. 2026).

Critically, Kedron et al. (2024) observed that reproducibility solutions disproportionately focus on data analysis, with insufficient attention to data acquisition. This is the phase preceding analysis that profoundly shapes downstream results. This observation directly motivates NORA\textquotesingle s dedicated spatial data download skill unit, which enforces provenance documentation, source authority ranking, and validation checks throughout the data acquisition process.

\section{System Design and Architecture}\label{system-design-and-architecture}

\subsection{Architecture Overview}\label{architecture-overview}

NORA comprises four architectural layers (User Layer, Skills Layer, Persistence Layer, and Infrastructure Layer) that collectively orchestrate the research (Figure 2). The detailed execution workflow and file generation instructions are illustrated in Appendix A.

\begin{figure}[htbp]
\centering
\includegraphics[width=\linewidth]{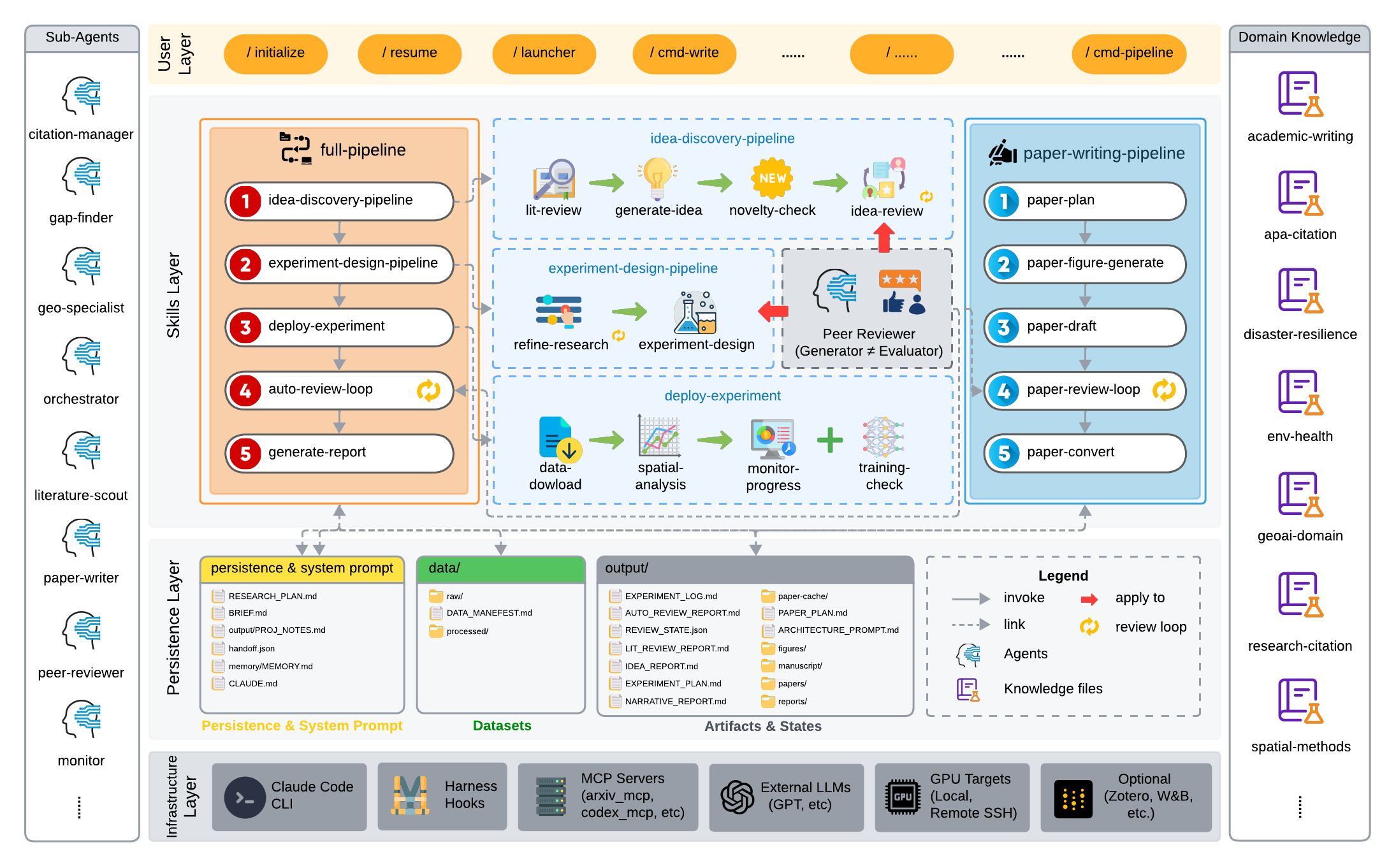}
\caption{NORA Architectural Design.}
\label{fig:2}
\end{figure}

\textbf{User Layer} serves as the primary interface between NORA and the user, enabling interaction through structured command execution. These commands encapsulate coordinated sets of skills while providing guidance that improves user experience.

\textbf{Skills Layer} constitutes the core of the framework, orchestrating and executing the functions required for automated research tasks. It comprises 21 Markdown-defined skills that encode procedural knowledge for specific research operations. Rather than functioning as rigid scripts, these skills operate as decision frameworks: they specify relevant considerations, required guardrails, and expected outputs, while allowing the executing agent to determine the most appropriate sequence of actions given the current research context. This layer is further supported by 9 sub-agents dedicated to domain-specific tasks, which are invoked by the skills when specialized reasoning or execution is required.

\textbf{Persistence Layer} stores system prompts, instructional files, state records, and output artifacts. Together, these components guide agent behavior and enable reliable transfer of contextual information across extended, multi-stage research workflows.

\textbf{Infrastructure Layer} provides external tools and computational environment. NORA is implemented within the Claude Code command-line interface (CLI), while allowing expandability. At this layer, MCP servers and external large language models provide specialized tool access and support reviewer--generator separation during the review process. The layer also manages connections to local and remote GPUs and facilitates interoperability with other software.

\subsection{System Design Principles}\label{system-design-principles}

NORA\textquotesingle s architecture is guided by three design principles: (1) \textbf{Skills-first modularity.} All research workflow logic resides in self-contained Markdown skill files. This design enables maximum flexibility within defined constraints: the agent reads a skill to understand the workflow, then determines the exact action sequence based on the context, which contrasts with template-based approaches that impose fixed execution paths regardless of context. This principle is crucial for enabling the system to leverage the rapidly evolving capabilities of backbone LLMs. (2) \textbf{Domain-specialized reasoning.} Rather than relying on general LLM knowledge of spatial methods, NORA encodes explicit decision frameworks for spatial analysis: when to use geographically weighted regression versus spatial lag models, how to select spatial weights matrices, and when spatial cross-validation is required, etc. The geo-specialist agent embodies deep expertise in spatial statistics, remote sensing, and GeoAI, providing method recommendations grounded in disciplinary conventions. (3) \textbf{Harness Engineering.} NORA\textquotesingle s harness is the scaffolding that governs the agent\textquotesingle s runtime: lifecycle hooks, persistent state, human checkpoints, and declarative configuration. It externalizes cross-cutting concerns, namely, permissioning, logging, checkpointing, and recovery. Harness engineering plays a crucial role because long-horizon autonomous research exceeds any single context window; structured file-based handoffs defeat "context anxiety" (Anthropic, 2025), enforce generator--evaluator separation, and make multi-session runs deterministically resumable. We therefore elevate it to a first-class design principle: skills encode intent and the harness encodes guarantees. More implementation details are provided in Section 4. The screenshot of the demonstration webpage and CLI interface is shown in Figure 3(a) and (b).

\begin{figure}[htbp]
\centering
\includegraphics[width=\linewidth]{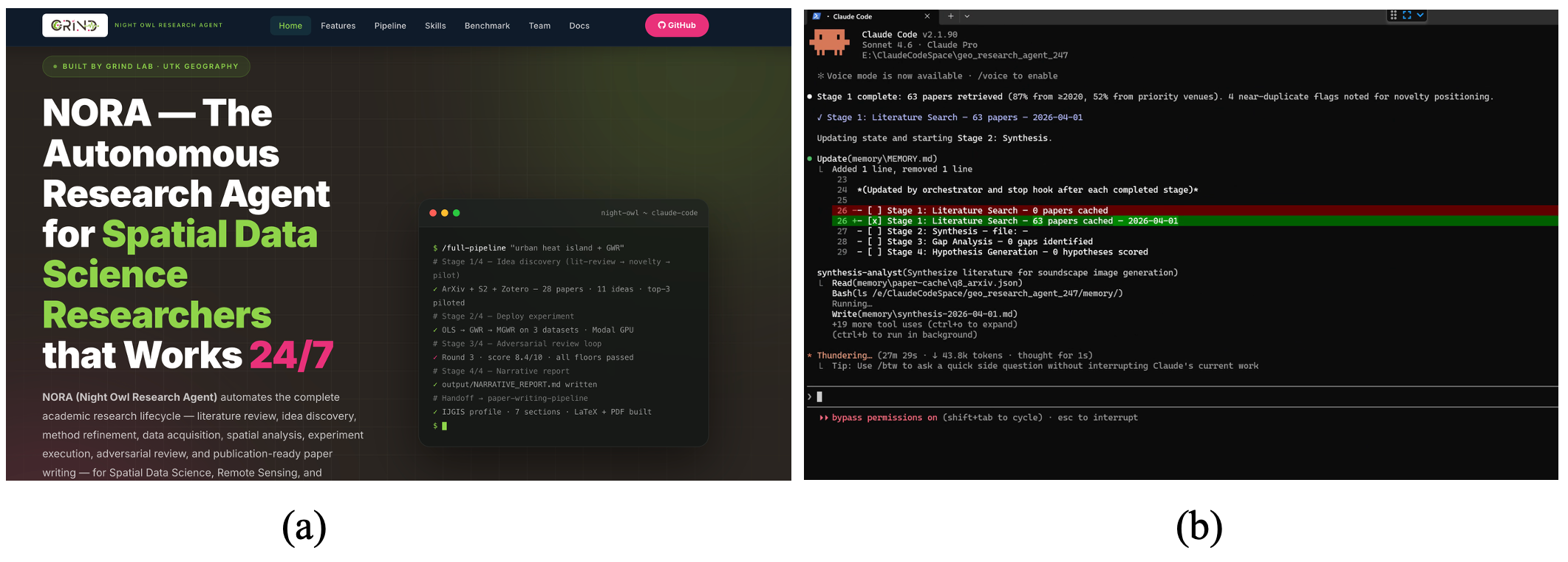}
\caption{Demonstration webpage and CLI interface.}
\label{fig:3}
\end{figure}

\subsection{Agent and Skill Design Principles}\label{agent-and-skill-design-principles}

Agent and skills design for autonomous AI research systems should follow a modular, tool-grounded, and provenance-first philosophy. For long-running scientific workflows, agents and skills should be embedded in an external harness rather than relying on the base model to maintain coherence on its own over extended sessions (Anthropic, 2025).

For skill design: instead of single monolithic prompts, the system should decompose research into reusable, domain-aware skills, such as literature review, data acquisition, analysis, visualization, and manuscript drafting, with explicit input-output contracts (Wang et al., 2023). Scientific credibility depends on transparency, each skill execution should also emit structured metadata and workflow provenance so that data sources, parameter settings, intermediate outputs, and decision points can be inspected, audited, and reproduced later (Nüst \& Pebesma, 2021). Detailed principles and items that should be included in skills are demonstrated in Appendix B.

For agent design, agents should interleave reasoning with action and invoke external tools or skills only through well-defined interfaces, since ReAct-style and tool-augmented language models are more effective and interpretable than text-only generation for multi-step problem solving and factual or computational tasks (Schick et al., 2023; Yao et al., 2023). Agents will also be invoked when subtasks, such as peer review, require context isolation to ensure that LLMs do not over-positively react to the content they generate (Anthropic, 2025). Detailed principles and items that should be included in agents are demonstrated in Appendix C.

\subsection{Research Lifecycle Workflow}\label{research-lifecycle-workflow}

\begin{figure}[htbp]
\centering
\includegraphics[width=\linewidth]{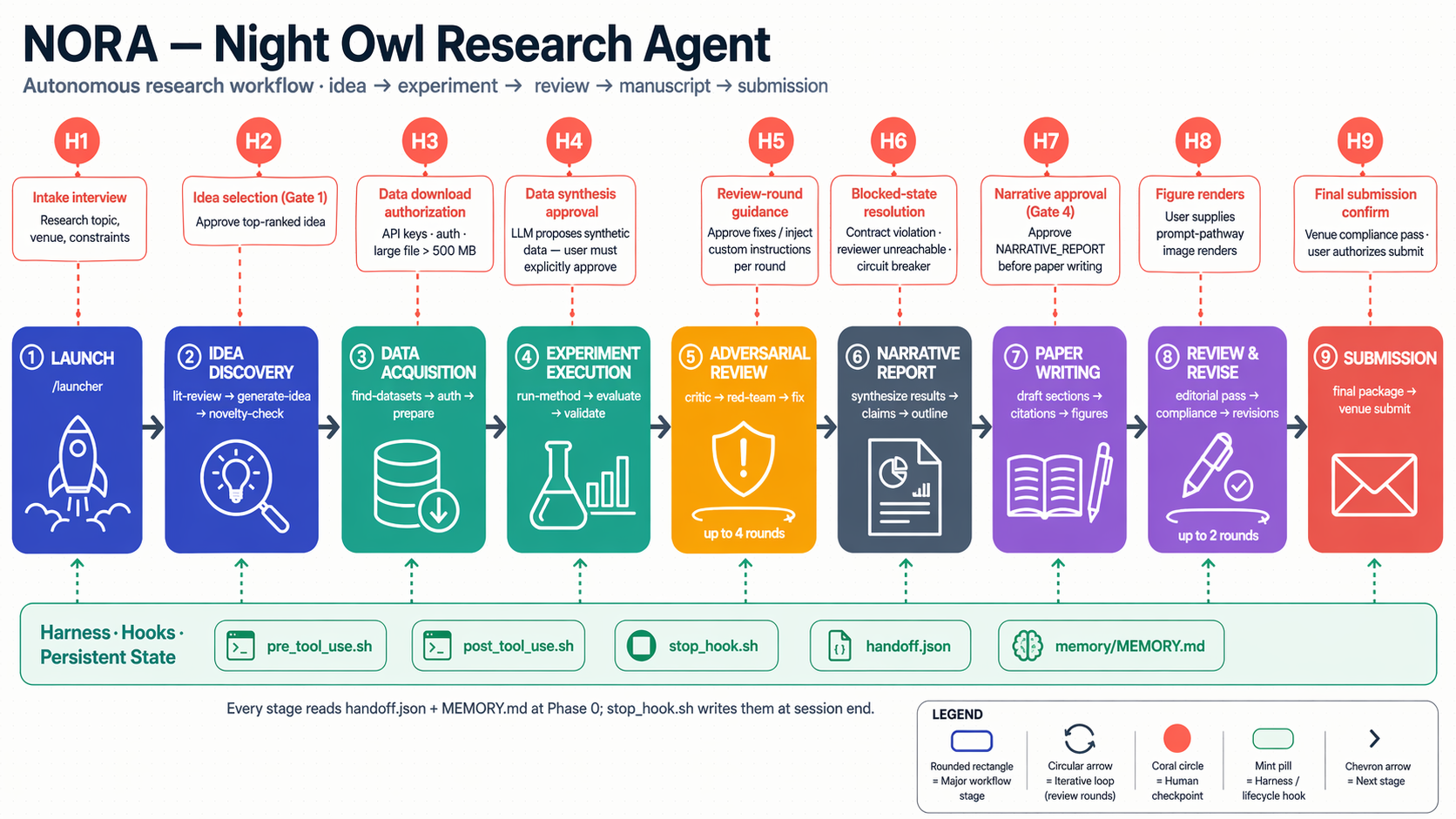}
\caption{NORA's research lifecycle workflow diagram with human checkpoint design.}
\label{fig:4}
\end{figure}

The NORA workflow is an end-to-end research pipeline that moves from idea formation to final submission through nine major stages and critical human checkpoints (Figure 4). It also allows the user to start and resume at any intermediate stage. Normally, it begins with Launch, where the user provides the research topic, target venue, and major constraints. In Idea Discovery, the system reviews literature, generates candidate ideas, and checks novelty before the user approves the strongest direction. Data Acquisition then gathers the needed datasets, with explicit human authorization required for sensitive access such as API keys, authentication, or very large downloads. The workflow also includes a checkpoint to approve proposed synthetic data before use.

Next, Experiment Execution runs the core analytical or modeling workflow and evaluates outputs. The process then enters Adversarial Review, an iterative quality-control stage where the system stress-tests the work, critiques weaknesses, and revises it for up to four rounds. If the process becomes blocked, a dedicated human intervention step resolves issues such as policy constraints or unreachable reviewers. After this, the Narrative Report synthesizes findings into a coherent interpretation, and the user must approve this report before writing begins.

In Paper Writing, the system converts the approved narrative into manuscript sections with citations and figures. Review and Revise add a second iterative loop, refining the draft through editorial improvement and compliance checks. Finally, Submission packages the manuscript and requires explicit final approval before submission. Beneath all stages sits a harness layer of hooks and persistent memory files, ensuring every stage can resume reliably and maintain context across the entire workflow.

\subsection{Configuration and Adaptability}\label{configuration-and-adaptability}

NORA\textquotesingle s behavior is governed by a central configuration file. Journal-specific templates encode venue requirements, including word limits, abstract structure, figure specifications, and required methodological elements. This configurability enables NORA to adapt to different research domains within spatial data science and to target different publication venues without architectural changes.

\section{Harness Engineering for Scientific Spatial Research}\label{harness-engineering-for-scientific-spatial-research}

Harness engineering is the discipline of designing the systems, constraints, and feedback loops that surround AI agents to make them reliable in production (Anthropic, 2025). A harness encompasses the complete infrastructure governing agent operation: tool access controls, safety guardrails, feedback mechanisms, observability layers, and state persistence mechanisms. In the context of scientific research agents, we extend this definition to encompass the engineering of constraints, validation gates, and provenance mechanisms that ensure autonomous research outputs meet the methodological standards of the target scientific community.

For spatial data science specifically, harness engineering must address challenges that do not arise in generic agent contexts: ensuring CRS consistency across heterogeneous datasets, checking spatial autocorrelation in model evaluation, enforcing appropriate spatial method selection given data characteristics, and maintaining the provenance chain from raw geospatial data through published claims. NORA implements harness engineering through 5 mechanisms:

\subsection{Lifecycle hooks}\label{lifecycle-hooks}

This hook-based approach ensures that the state is never lost between sessions, even when sessions terminate unexpectedly, by executing shell scripts. The pre-tool-use hook (pre\_tool\_use.sh) validates every tool invocation before execution, blocking destructive operations (rm -rf, force push), preventing writing files to system directories, and logging all tool calls for audit. The stop hook (stop\_hook.sh) executes on session termination, performing three critical functions: (1) updating persistent memory (MEMORY.md) with current pipeline state, token usage, and section quality scores; (2) writing a structured handoff file (handoff.json) capturing pipeline position, review state, experiment results, and recovery instructions; and (3) triggering notifications.

\subsection{Generator-evaluator separation}\label{generator-evaluator-separation}

NORA enforces that the entity generating content never evaluates its own output. This separation operates at multiple levels. For example, the paper-writer agent drafts sections in one context, while the peer-reviewer agent scores them in a separate context without access to writer reasoning. The mechanism is implemented via MCP (e.g., configurable as GPT via the Codex MCP). Claude Code subagent alternatives are also provided to ensure the main context is quarantined from the review agent when external LLMs are configured erroneously. The review loop implements a scoring mechanism inspired by but extending the automated review approach of The AI Scientist (Lu et al. 2026). Five dimensions are scored on a 0--10 scale: novelty, rigor, literature coverage, clarity, and impact. Acceptance criteria are adjustable, and default to a weighted average score over 7.5, and all five floors are met. The scoring system supports few-shot calibration anchors that link each score level to GIScience-specific examples provided by users. These anchors will ground the scoring in concrete domain expectations.

\subsection{Structured state persistence}\label{structured-state-persistence}

Rather than relying on LLM context windows and avoiding the risk of information loss during context compression in long-running tasks, NORA persists all critical states to structured files. The handoff.json file captures eight top-level sections: pipeline position (stage, last completed step, next step), review state (per-criterion scores, decision, action items), last experiment results (best model), paper draft state (sections accepted/pending, last score), token budget, and recovery hints (files to read on resume, skill to invoke). On session restart, the orchestrator agent reads handoff.json and deterministically resumes from the exact interruption point, never re-running completed stages or accepted sections.

\subsection{Human-in-the-loop}\label{human-in-the-loop}

In a domain-specific research pipeline, it is crucial to integrate human checkpoints to ensure the reliability and validity of the generation even when the system is reliable and execute the entire process automatically. The pipeline runs autonomously but pauses at nine deliberate checkpoints (Figure 3): (1) the /launcher intake interview; (2) idea selection after idea-discovery-pipeline; (3) data-download authorization for authentication, API keys, or files above threshold; (4) explicit approval before any fallback to synthetic data; (5) per-round guidance in auto-review-loop; (6) blocked-state resolution on contract violation; (7) narrative-report approval before the paper pipeline; (8) figure-render approval for prompt-pathway diagrams; and (9) final submission confirmation via submit-check. These gates guarantee that autonomous research systems do not blindly self-authorize irreversible decisions. For example, data provenance, fabrication risk, and computing expenditure all require accountable human judgment.

\section{Domain-Specialized Skills for Spatial Research}\label{domain-specialized-skills-for-spatial-research}

\subsection{Rationale for Domain-Specific Skill Decomposition}\label{rationale-for-domain-specific-skill-decomposition}

General-purpose research agents treat all domains equivalently, relying on the LLM\textquotesingle s parametric knowledge to determine appropriate methods. This approach is insufficient for spatial data science for three reasons. First, spatial analysis requires explicit methodological selection. Second, geospatial data acquisition is challenging (Goodchild 2007). Third, spatial research has discipline-specific validity criteria such as Moran\textquotesingle s I of residuals, spatial cross-validation, etc. NORA addresses these requirements through two dedicated skill units that encode domain-specific procedural knowledge: the spatial analysis skill unit and the spatial data download skill unit. NORA also supports easy expansion by adding skills that cover other aspects of spatial data science.

\subsection{Spatial Analysis Skill Unit}\label{spatial-analysis-skill-unit}

\subsection*{Components}\label{spatial-analysis-components}
The spatial analysis skill unit is a structured decision framework that ensures methods are chosen based on research questions rather than convenience. It includes six connected components. First, it classifies nine analytical objectives (description, explanation, prediction, clustering, accessibility, spatiotemporal change, and causal inference) and links them to appropriate methods. Second, it provides data-readiness guidance for CRS selection, geometry validity, duplicates, dataset integration, and variable preparation, including outlier handling, skewness, multicollinearity, and standardization. Third, it guides the selection of spatial weights and neighbors. Fourth, it offers analytical decision frameworks for regression, clustering, prediction, accessibility, network analysis, and interpolation. For regression, it moves from OLS to residual Moran's I testing and then to spatial lag, spatial error, GWR, or MGWR when appropriate. Fifth, it requires key diagnostics and robustness checks, including residual spatial autocorrelation, heteroscedasticity, MAUP sensitivity, and boundary effects when appropriate. Sixth, it imposes guardrails against common mistakes, such as using latitude/longitude for distance calculations, clustering raw counts, relying on random rather than spatial cross-validation, or overstating causality. This distinguishes NORA from general-purpose agents by embedding explicit, operational spatial reasoning into the workflow.

\subsection{Spatial Data Download Skill Unit}\label{spatial-data-download-skill-unit}

The spatial data download skill unit addresses an overlooked but crucial part of spatial research: reproducible data acquisition. Responding to Kedron et al.'s argument that geospatial reproducibility efforts focus too heavily on analysis instead of data collection, this skill enforces a six-phase protocol. First, it assesses data needs across ten dimensions, including geography, time, resolution, variables, format, licensing, quality, and size, and maps them to sixteen geospatial data categories. Second, it guides source discovery through a seven-tier hierarchy that prioritizes authoritative and stable providers such as government agencies, international organizations, research repositories, open data projects, APIs, and academic sources. Third, it evaluates candidate sources using nine criteria, including authority, coverage, usability, freshness, stability, and citability. Access controls classify sources as public, API-based, login-restricted, institutional, paid, click-through, or CAPTCHA-protected, forcing human approval when needed. Fourth, it supports multiple download methods, including direct HTTP, APIs, archive extraction, cloud-optimized access, and OpenStreetMap queries, with retry and rate-limiting safeguards. Fifth, it validates downloaded files and checks geographic, temporal, and structural integrity. Sixth, it records every dataset in a DATA\_MANIFEST.md file with detailed provenance fields.

\section{Experiment Design and Evaluation Framework}\label{experiment-design-and-evaluation-framework}

\subsection{Case Study Design}\label{case-study-design}

We design case studies in which NORA attempts to produce research reports for leading venues in spatial data science, such as the \emph{International Journal of Geographical Information Science (IJGIS).} For each case study, NORA receives a research brief that specifies the topic, target venue, geographic scope, and key datasets. The system then executes the full pipeline autonomously with human checkpoints enabled during execution.

Evaluating an autonomous research agent requires moving beyond the task-level metrics typical of GeoAI benchmarks to assess research-level quality. To this end, we design a multi-faceted evaluation comprising three complementary components: case studies with pipeline analysis, expert and LLM evaluation, and ablation studies.

\subsection{Expert Evaluation Protocol}\label{expert-evaluation-protocol}

Three domain experts were invited to test NORA and provide qualitative analysis. The experts were selected to cover complementary areas of spatial data science, including spatial statistics, remote sensing, GeoAI, and geographic information systems.

The three experts were asked to monitor the entire research report generation process and record critical behaviors of NORA based on the following guiding questions:

\begin{enumerate}
\def\labelenumi{(\arabic{enumi})}
\item
  Did the agent correctly understand the research objective, scope, target venue, and constraints?
\item
  Did the agent generate ideas that were relevant and non-trivial?
\item
  Did the agent identify the most relevant literature?
\item
  Did the agent identify datasets appropriate for the research question?
\item
  Did the agent correctly clean, transform, and integrate the data?
\item
  Were the chosen methods appropriate for the research question?
\item
  Did the agent execute the workflow correctly and consistently?
\item
  Did the agent demonstrate real domain reasoning rather than generic AI output?
\item
  Did the agent interpret the results correctly and cautiously?
\item
  Were conclusions supported by the evidence?
\item
  Did the agent know when to pause and ask for approval?
\item
  At which stage was the agent strongest and weakest?
\end{enumerate}

To validate the contribution of each core architectural unit, we design ablation studies in which individual components are removed or degraded while the remainder of the system operates normally. The detailed design and study results are recorded in Appendix E.

\subsection{Evaluation Metrics}\label{evaluation-metrics}

For quantitative assessment, we extend the evaluation metrics from OpenReview, 2024, and focus on the following seven dimensions. Each dimension will be rated from 1 to 10, with 10 being the strongest.

\begin{enumerate}
\def\labelenumi{(\arabic{enumi})}
\item
  \textbf{Idea novelty:} The extent to which the manuscript proposes an original research question, perspective, method, or application that goes beyond well-established or routine work.
\item
  \textbf{Literature coverage:} The degree to which the manuscript identifies, cites, and meaningfully engages with the most relevant prior work in the field.
\item
  \textbf{Significance:} The degree to which the study addresses an important problem and has the potential to make meaningful scientific, practical, or societal impacts.
\item
  \textbf{Code quality:} The extent to which the code is correct, efficient, readable, modular, well-documented, and reproducible for supporting the reported analyses.
\item
  \textbf{Figure quality:} The extent to which figures are accurate, readable, well-designed, and effective in expressing the main findings of the study.
\item
  \textbf{Writing quality:} The degree to which the manuscript is clear, coherent, well-structured, precise, and appropriate in tone for scholarly communication.
\item
  \textbf{Rigor:} The extent to which the study demonstrates methodological soundness, analytical depth, appropriate validation, and careful interpretation of results.
\end{enumerate}

\section{Results and Analysis}\label{results-and-analysis}

\subsection{Case Study 1: Multi-Vintage Street-View First-Floor Elevation}\label{case-study-1-multi-vintage-street-view-first-floor-elevation}

\textbf{Case and experiment.} This case study asks whether historical Google Street View (GSV) panoramas, treated as a multi-vintage archive rather than a single snapshot, can raise the coverage and decision-utility of community-scale first-floor elevation (FFE) measurement for a barrier-island flood-resilience study in the USA (North Wildwood, NJ). The researcher supplied the direction "use historical street-view images to improve the coverage of first-floor-elevation measurement for North Wildwood, NJ" together with two anchor papers (Ning et al., 2022; Gao et al., 2024) and a 121-image test set with 254 annotated house-front doors for detector benchmarking. The main idea of this method is to measure the elevation of the door bottom as a proxy for the first-floor elevation. NORA executed the full workflow, from literature review to a report with embedded figures, in a single multi-stage session. The study sits in a small but active lineage of street-view first-floor-elevation studies.

The work evaluates a parcel-anchored pipeline over the full residential stock of North Wildwood as represented in OpenStreetMap, namely 1,078 OSM building parcels assigned to 2,199 unique panoramas captured between 2008 and 2022, with Gao et al.\textquotesingle s 2024 YOLOv5 door detector used as a fixed, non-retrained measurement instrument benchmarked against zero-shot Grounding-DINO (Liu et al., 2025). The headline numeric finding, a +15.6 percentage-point coverage gain from latest-only to any-vintage extraction (29.9 \% $\rightarrow$ 45.5 \%), is conditional on that OSM cohort, and here a notable upstream failure surfaced: NORA did not recognize that OSM building coverage for North Wildwood is materially incomplete relative to a cadastral-parcel inventory, and accordingly missed roughly 80 of the retrievable panoramas. The under-sampling propagates directly into every downstream statistic: the 0-of-29 coverage on FEMA V-zone parcels is most plausibly a lower bound on the true stilted-foundation coverage gap.

\textbf{Human interventions.} Roughly a dozen genuine interventions occurred across the multi-session workflow, of which nine were pipeline-critical. The researcher (1) directed NORA to read the provided reference papers and use the APIs those papers imply for downloading GSV imagery, rather than writing a bespoke client; (2) asked NORA to run an explicit method comparison between a pre-trained YOLO detector and a prompt-based zero-shot model (Grounding-DINO); (3) pointed NORA to the correct code snippets in the relevant GitHub repositories so that the downloader used the supported library calls rather than a reimplementation; (4) provided the no-retraining instruction for the door detector, later reinforced with the clarification that Gao 2024\textquotesingle s checkpoint had already been trained and any retraining would be redundant; (5) supplied information about the pre-installed GPU environment required to run Grounding-DINO; (6) cleared the Windows paging file and enabled GPU use when the first Grounding-DINO inference attempt failed under memory pressure; (7) denied NORA\textquotesingle s request for an OpenTopography academic API key for an external DEM (as used by the reference paper) and redirected NORA to use the EGM96 elevation embedded in each panorama\textquotesingle s metadata as the per-pano ground reference; (8) requested that NORA send detection thumbnails to Codex for adversarial examination of the YOLO and Grounding-DINO outputs; and (9) requested that the necessary figures be inserted into the final report. Each of these interventions kept the pipeline on a scientifically defensible path, and two of them, the redirect away from an external DEM and the library-and-weights pointers, replaced extra work.

\textbf{Case summary.} Extracting first-floor elevation from street-view imagery is non-trivial work that demands intensive skills in programming, data acquisition, and back-and-forth visual and data inspection. From a programming standpoint, NORA significantly reduced the human workload by automating development-environment setup, multi-source data downloading, detector benchmarking, and end-to-end testing; otherwise tedious tasks were automated at low cost. The work still required domain-specific knowledge that NORA did not possess out of the box: it could not independently identify the correct data sources or pre-trained weights without being pointed at them, it did not use perspective-cropped and full-panorama imagery appropriately for the FFE geometry, and it lacked a solid data-inspection instinct for failure modes characteristic of the domain (most visibly, OSM parcel incompleteness relative to a cadastral cohort). Human researchers develop such instincts from accumulated past failures, and NORA currently cannot build equivalent experience on its own because it cannot perform the right quality-control routines without full context on the user\textquotesingle s purpose; it works well when the user supplies timely corrective feedback. Its relatively weak visual capability also limited its performance on image-intensive analyses. In this case, NORA served as a helpful and efficient assistant, yet the user still had to steer the direction at several critical steps.

\subsection{Case Study 2: Atlanta Crime-Fairness Theorem}\label{case-study-2-atlanta-crime-fairness-theorem}

\textbf{Case and experiment.} This case study investigates whether the choice of spatial algorithm affects not only predictive accuracy but also group-level fairness, using census-tract-level crime prediction in the City of Atlanta. The researcher supplied the direction: \emph{"Atlanta census-tract crime rate prediction with a spatial algorithm using ACS sociodemographic covariates"} and NORA executed the full research workflow from literature review to a submission-ready narrative report in a single session. The resulting study benchmarks seven place-based predictors (OLS, XGBoost, Spatial Lag, Spatial Error, GWR, MGWR, and a Spatial-XGBoost tree baseline that substituted for a GCN after a local torch library failure) against a proven lower bound linking residual Moran\textquotesingle s I, a protected-group clustering coefficient, and a disparate-impact gap. Empirical verification uses 19,494 APD Part-I incidents from 2025 joined at 94.9\% to 169 Atlanta city tracts, paired with ACS 5-year 2018-2022 covariates, plus a 1,000-dataset Monte-Carlo simulation for the theorem and its group-imbalance corollary.

\textbf{Workflow execution by NORA.} NORA launched the /full-pipeline skill, which chained four sub-skills: idea-discovery-pipeline $\rightarrow$ deploy-experiment $\rightarrow$ auto-review-loop $\rightarrow$ generate-report. Idea discovery produced twelve candidates, pruned to six by feasibility-and-novelty filtering, and merged the top three into a unified proposal after a deep novelty-check (score 7.8/10). The refine-research loop reached 8.95/10 over two rounds, and experiment-design produced a plan with five pre-registrations demanded by the internal peer-reviewer sub-agent: a formal theorem statement, block-spatial cross-validation, parameter-count caps, a race-in/race-out covariate ablation, and the demotion of SVC-NN from the contribution list. \emph{deploy-experiment} then ran seventy Atlanta benchmark fits plus the simulation grid, and the adversarial auto-review-loop iterated from 5.0 (ALMOST) in Round 1 to 6.5 (READY) in Round 2, where it corrected a discovered MGWR library-signature bug, added GWR local-coefficient surfaces, documented the analysis CRS, and restructured the paper\textquotesingle s narrative around Corollary 1 and a newly surfaced fold-4 extrapolation finding. Each stage wrote its own structured state to \emph{handoff.json} and \emph{REVIEW\_STATE.json} so the pipeline could survive context compaction, a direct instantiation of NORA\textquotesingle s harness-engineering principles of lifecycle hooks and structured state persistence.

\textbf{Contribution of the spatial analysis and data download skill units.} The spatial analysis skill unit was decisive at several branching points. It enforced an OLS $\rightarrow$ residual-Moran\textquotesingle s-I $\rightarrow$ spatial-lag/error $\rightarrow$ GWR/MGWR progression rather than a monolithic machine-learning sweep, mandated 5-fold block-spatial CV with a two-tract buffer over random splits, defended the use of rook contiguity and the symmetric weight matrix for Moran\textquotesingle s I computation, forced all distance-sensitive operations into EPSG:26967 (NAD83 Georgia West), and required MAUP-consistent residual and fairness diagnostics. The data download skill unit orchestrated three authoritative sources, i.e., TIGER 2020 tract polygons, the Census ACS 5-year API, and the Atlanta Police Department open data portal, and recorded each in data/DATA\_MANIFEST.md with provenance fields including retrieval date, variable list, spatial-join rate, and licensing. When the ArcGIS crime-layer endpoint probes failed, the skill\textquotesingle s validation step flagged the gap explicitly rather than silently proceeding with synthetic stand-in data, enabling the human researcher to supply an authoritative CSV that the protocol then validated, spatial-joined, and merged.

\textbf{Human interventions.} Seven genuine interventions occurred across the full pipeline. The researcher 1) supplied the initial research direction, 2) authorized the creation of a seeded RESEARCH\_PLAN.md, 3) provided the Census API key (stored in a git-ignored .env), 4) supplied the authoritative APD 2025 CSV mid-pipeline, 5) approved Gate 1 after the refined proposal, 6) chose the \emph{reframe} rather than \emph{rescue} option at the Round-1 adversarial-review checkpoint, and 7) requested the final deliverable in Word format. Two of these, i.e., the API key and the real-data CSV, materially improved the reproducibility of the study. Every other decision, including model selection, hyperparameter caps, bug diagnosis and repair, pre-registration tracking, and failure-pivot activation, was made autonomously by NORA under its generator-evaluator-separated architecture.

\textbf{Final report and generated manuscript.} Stage 4 produced NARRATIVE\_REPORT.md together with a matching NARRATIVE\_REPORT.docx (57 KB): twenty-one sections, a ten-row claim-to-evidence matrix citing verbatim numbers, a nine-row figure-and-table plan with five figures already rendered, honest failure-pivot documentation for two pre-registered criteria that did not meet threshold, and twenty verified citation seed keys with no fabricated BibTeX. The narrative is explicitly framed as the contract consumed by NORA\textquotesingle s downstream paper-writing-pipeline, which was left uninvoked in this session so that the case study remains confined to a single end-to-end loop.

\subsection{Case Study 3: Stochastic Nested Surface Depression Delineation}\label{case-study-3-stochastic-nested-surface-depression-delineation}

\textbf{Case Study and Experiment.} The researcher provided three foundational papers in deterministic surface-depression delineation (Wu et al., 2015; Wu and Lane, 2017; Wu et al., 2018), along with the open-source LiDAR Python package (Wu, 2021), and tasked NORA with generating new research ideas based on this lineage. From ten candidate ideas, NORA selected the question of whether vertical uncertainty in LiDAR-derived digital elevation models (DEMs) can be propagated through nested surface-depression hierarchies in the Prairie Pothole Region (PPR), and whether the resulting probabilistic measures of depression persistence improve predictions of independent surface-water observations. The proposed approach bridges deterministic level-set methods with the per-cell stochastic framework introduced in WhiteboxTools (Lindsay, 2018). NORA implemented the full workflow over the Cottonwood Lake Study Area in North Dakota, using a 3 m USGS 3DEP DEM, Sentinel-1 GRD VV and VH backscatter data (2018--2020), and the JRC Global Surface Water dataset (Pekel et al., 2016). The deterministic baseline identified 5,930 primitive depressions. A Gaussian random field ensemble (N = 25; $\sigma$ = 0.10 m, $\ell$ = 60 m) achieved methodological equivalence with WhiteboxTools under matched conditions, yielding a per-depression Spearman correlation of $\rho$ = 0.929. A key finding is a cell-versus-hierarchy decoupling effect. While cell-level persistence remains high (mean = 0.929 across reference depressions), strict parent--child label agreement within the level-set hierarchy shows a median of 0.0 across 577 candidate pairs. This indicates that hierarchical topology is substantially more sensitive to LiDAR vertical uncertainty than the existence of depressions themselves.

\textbf{Workflow execution by NORA.} NORA chained idea-discovery $\rightarrow$ deploy-experiment $\rightarrow$ auto-review-loop $\rightarrow$ generate report. The idea discovery stage produced ten candidate directions and applied an automated novelty assessment, which identified the uncertain merge-tree literature as the primary external risk. An adversarial pre-review then defined the minimum viable contribution and triggered a scope reduction from the 2,770 km$^{2}$ Pipestem subbasin to the Cottonwood Lake HUC12 under a wall-clock constraint. The proposal was subsequently refined with pre-registered success criteria and the Persistence-weighted Inundation Consistency (PwIC) metric formally specified. The experiment deployment stage executed a structured sequence of experiments (R001--R015), including synthetic validation tests, the deterministic baseline, a Gaussian random field (GRF) ensemble, and the WhiteboxTools baseline (Lindsay, 2018). The automated review stage improved to 7.5/10 (Round 2). The pre-registered target (PwIC\_primary $\geq$ 0.40) was not achieved (Sentinel-1 $\rho$ = 0.164; JRC GSW $\rho$ = 0.250). The contribution was then reframed around three supported findings: methodological equivalence with the Lindsay (2018) baseline, persistence-driven residual signal within size bins, and the decoupling between cell-level persistence and hierarchical structure. Finally, the report generation stage compiled all results.

\textbf{Human interventions.} Five substantive interventions were made by the researcher: (1) providing the foundational papers and LiDAR package, and requesting new research directions without specifying a hypothesis or study area; (2) approving the scope reduction from the Pipestem subbasin to Cottonwood Lake at the pre-review stage; (3) granting access to Google Earth Engine for data retrieval; (4) endorsing the final review decision (Path A), including withdrawal of the strawman claim and deferral of large-scale replication; and (5) requesting the final deliverables in Word format with embedded figures.

\textbf{Final report and generated manuscript.} The final stage produced a comprehensive narrative report, along with a manuscript in both Microsoft Word and PDF formats. Outputs include five core figures and three tables. All results are traceable to structured JSON files at line-level granularity, with both significant and null findings reported. Replication at the full Pipestem extent is explicitly deferred as future work under the validated wall-clock protocol established in the Cottonwood Lake pilot study.

\subsection{Expert and LLM Evaluation Results}\label{expert-and-llm-evaluation-results}

Six domain experts and three LLM reviewers evaluated the reports generated by NORA across the seven evaluation dimensions illustrated in section 6.3. The results are listed in Table 1.

\begingroup\renewcommand{\arraystretch}{1.05}
% Table 1: Reports rating from domain experts and LLM reviewers
\begin{tabularx}{\linewidth}{@{}>{\hsize=0.9609\hsize\centering\arraybackslash}X>{\hsize=1.2199\hsize\centering\arraybackslash}X>{\hsize=0.8399\hsize\centering\arraybackslash}X>{\hsize=1.4439\hsize\centering\arraybackslash}X>{\hsize=0.9459\hsize\centering\arraybackslash}X>{\hsize=1.0179\hsize\centering\arraybackslash}X>{\hsize=1.1039\hsize\centering\arraybackslash}X>{\hsize=1.1819\hsize\centering\arraybackslash}X>{\hsize=0.5659\hsize\centering\arraybackslash}X>{\hsize=0.7199\hsize\centering\arraybackslash}X@{}}
\toprule\noalign{}
\textbf{Case No.} & \textbf{Reviewers} & \textbf{Idea Novelty} & \textbf{Literature Coverage} & \textbf{Significance} & \textbf{Code Quality} & \textbf{Figure Quality} & \textbf{Writing Quality} & \textbf{Rigor} & \textbf{Average} \\
\midrule\noalign{}
\endhead
\bottomrule\noalign{}
\endlastfoot
\multirow{11}{=}{\centering\arraybackslash Case Study 1} & Reviewer 1 & 5.0 & 6.0 & 5.0 & 6.0 & 7.0 & 4.0 & 4.0 & 5.3 \\
& Reviewer 2 & 6.0 & 6.0 & 5.0 & 6.0 & 7.0 & 4.0 & 7.0 & 5.9 \\
& Reviewer 3 & 7.0 & 9.0 & 5.0 & 6.0 & 7.0 & 6.0 & 5.0 & 6.4 \\
& Reviewer 4 & 8.5 & 9.0 & 8.0 & 9.0 & 8.0 & 8.5 & 8.0 & 8.4 \\
& Reviewer 5 & 5.0 & 4.0 & 4.0 & 6.0 & 5.0 & 7.0 & 3.0 & 4.9 \\
& Reviewer 6 & 6.0 & 4.0 & 7.0 & 7.0 & 4.0 & 2.0 & 1.0 & 4.4 \\
& \textbf{Human Avg} & \textbf{6.3} & \textbf{6.3} & \textbf{5.7} & \textbf{6.7} & \textbf{6.3} & \textbf{5.3} & \textbf{4.7} & \textbf{5.9} \\
& Opus 4.7 & 8.0 & 6.0 & 7.0 & 5.0 & 5.0 & 6.0 & 8.0 & 6.4 \\
& GPT 5.5 & 8.0 & 7.0 & 8.0 & 6.0 & 6.5 & 7.0 & 7.0 & 7.1 \\
& Gemini 3.1 Pro & 8.0 & 8.0 & 8.0 & 7.0 & 8.0 & 9.0 & 9.0 & 8.1 \\
& \textbf{LLM Avg} & \textbf{8.0} & \textbf{7.0} & \textbf{7.7} & \textbf{6.0} & \textbf{6.5} & \textbf{7.3} & \textbf{8.0} & \textbf{7.2} \\
\multirow{11}{=}{\centering\arraybackslash Case Study 2} & Reviewer 1 & 6.0 & 7.0 & 7.0 & 6.0 & 6.0 & 5.0 & 6.0 & 6.1 \\
& Reviewer 2 & 5.0 & 6.0 & 6.0 & 5.0 & 6.0 & 4.0 & 5.0 & 5.3 \\
& Reviewer 3 & 5.0 & 4.0 & 7.0 & 8.0 & 9.0 & 6.0 & 7.0 & 6.6 \\
& Reviewer 4 & 6.0 & 9.0 & 9.0 & 8.5 & 7.0 & 8.5 & 9.0 & 8.1 \\
& Reviewer 5 & 6.0 & 5.0 & 5.0 & 6.0 & 4.0 & 5.0 & 3.0 & 4.9 \\
& Reviewer 6 & 5.0 & 5.0 & 6.0 & 8.0 & 6.0 & 5.0 & 4.0 & 5.6 \\
& \textbf{Human Avg} & \textbf{5.5} & \textbf{6.0} & \textbf{6.7} & \textbf{6.9} & \textbf{6.3} & \textbf{5.6} & \textbf{5.7} & \textbf{6.1} \\
& Opus 4.7 & 6.0 & 8.0 & 7.0 & 7.0 & 8.0 & 8.0 & 8.0 & 7.4 \\
& GPT 5.5 & 8.5 & 8.0 & 8.5 & 7.0 & 8.0 & 9.0 & 7.5 & 8.1 \\
& Gemini 3.1 Pro & 7.0 & 9.0 & 8.0 & 8.0 & 8.0 & 9.0 & 8.0 & 8.1 \\
& \textbf{LLM Avg} & \textbf{7.2} & \textbf{8.3} & \textbf{7.8} & \textbf{7.3} & \textbf{8.0} & \textbf{8.7} & \textbf{7.8} & \textbf{7.9} \\
\multirow{11}{=}{\centering\arraybackslash Case Study 3} & Reviewer 1 & 6.0 & 7.0 & 6.0 & 7.0 & 5.0 & 6.0 & 5.0 & 6.0 \\
& Reviewer 2 & 6.0 & 7.0 & 7.0 & 6.0 & 6.0 & 5.0 & 5.0 & 6.0 \\
& Reviewer 3 & 9.0 & 6.0 & 7.0 & 9.0 & 5.0 & 9.0 & 8.0 & 7.6 \\
& Reviewer 4 & 7.0 & 9.0 & 7.5 & 8.5 & 9.0 & 8.0 & 8.0 & 8.1 \\
& Reviewer 5 & 5.0 & 6.0 & 4.0 & 7.0 & 3.0 & 5.0 & 3.0 & 4.7 \\
& Reviewer 6 & 3.0 & 4.0 & 4.0 & 7.0 & 3.0 & 4.0 & 2.0 & 3.9 \\
& \textbf{Human Avg} & \textbf{6.0} & \textbf{6.5} & \textbf{5.9} & \textbf{7.4} & \textbf{5.2} & \textbf{6.2} & \textbf{5.2} & \textbf{6.0} \\
& Opus 4.7 & 7.0 & 8.0 & 6.0 & 8.0 & 7.0 & 9.0 & 9.0 & 7.7 \\
& GPT 5.5 & 8.5 & 8.5 & 7.5 & 8.0 & 7.5 & 8.0 & 8.0 & 8.0 \\
& Gemini 3.1 Pro & 9.0 & 9.0 & 8.0 & 8.0 & 8.0 & 9.0 & 9.0 & 8.6 \\
& \textbf{LLM Avg} & \textbf{8.2} & \textbf{8.5} & \textbf{7.2} & \textbf{8.0} & \textbf{7.5} & \textbf{8.7} & \textbf{8.7} & \textbf{8.1} \\
\end{tabularx}

The evaluation suggests that NORA can produce research manuscripts with moderate-to-strong scholarly quality, but the results also reveal clear areas where human oversight remains necessary. NORA's three generated manuscripts received an overall average score of 6.58/10 across six human reviewers and three LLM evaluators, suggesting a generally promising but uneven level of manuscript quality. Human reviewers were notably more conservative, with an average score of 6.01, compared with 7.73 from the LLM evaluators. This gap is important: while LLMs appear to recognize the structural coherence and apparent completeness of the manuscripts, human reviewers seem more sensitive to disciplinary expectations, especially regarding rigor, interpretation, and scholarly contribution. Across the three case studies, the combined average scores were relatively close, increasing modestly from 6.33 for Case Study 1 to 6.69 for Case Study 2 and 6.73 for Case Study 3, indicating that NORA's performance was reasonably consistent rather than dependent on a single strong case. Among the seven dimensions, code quality received the highest combined score \textbf{(}7.04), followed by literature coverage \textbf{(}6.83), suggesting that NORA is relatively effective at producing technically organized and literature-aware research outputs. However, rigor received the lowest combined score \textbf{(}6.17\textbf{)} and the lowest human-only score \textbf{(}5.17\textbf{)}, highlighting a key limitation: autonomous research agents may generate plausible workflows and polished narratives, but methodological validation and careful interpretation still require expert oversight. The largest human--LLM discrepancies occurred in rigor and writing quality, whereas scores for code quality were nearly identical. Overall, the results position NORA as a useful research acceleration system, but not a substitute for human scholarly judgment.

\section{Discussion}\label{discussion}

\subsection{Implications for Domain-specific Research Practice}\label{implications-for-domain-specific-research-practice}

NORA shows that domain-specialized autonomous research agents can substantially automate spatial data science workflows while preserving the methodological standards. Its ability to enforce core spatial analysis conventions through explicit, skill-based decision frameworks marks a meaningful advance beyond both general-purpose research agents and task-specific GeoAI tools. The system's harness engineering approach provides a principled foundation for building trustworthy autonomous research systems. By implementing quality assurance mechanisms, NORA makes reliability a property of the surrounding engineering infrastructure rather than relying solely on model capability. This distinction is especially important in spatial data science, because accessible and reliable datasets and the correct selection of spatial methods and evaluation metrics are crucial to ensure scientific rigor. NORA therefore occupies a distinct position within the broader landscape of AI research tools. Compared with The AI Scientist (Lu et al. 2026), NORA exchanges domain generality for spatial specialization. Compared with GIS Copilot (Akinboyewa et al. 2025), NORA moves beyond task-level spatial operations toward full research lifecycle orchestration.

\subsection{Responsible Automation and Human Oversight}\label{responsible-automation-and-human-oversight}

Based on the case studies, NORA can generate inspiring and novel research ideas and automate experiments. However, responsible automation requires careful identification of where autonomy is appropriate and where human judgment remains necessary. In NORA, several workflow stages are intentionally configured to invite human intervention. These checkpoints performed effectively in the case studies. Technically, these stages could be fully automated by allowing the LLM to select the ``best'' idea, dataset, or method based on predefined criteria. Nevertheless, preserving human checkpoints at these moments helps ensure that the research remains scientifically grounded, ethically aware, and aligned with domain expectations. Human oversight is also necessary when the automated workflow encounters failure or ambiguity. For example, when code execution fails, data acquisition is incomplete, or a required dataset is unavailable, the LLM may attempt to find an alternative data source, modify the analytical strategy, or even reframe the research question. Although such adaptability is useful, these decisions can substantially alter the scope, assumptions, and implications of the study. Human intervention is therefore important to determine whether the modified pathway remains valid.

NORA's generator--evaluator separation and verification gates further support responsible automation by preventing unsupported outputs from moving forward. The LLM-based review mechanism was intentionally strict: when a manuscript or experiment failed to meet the predefined evaluation threshold, the system stopped subsequent tasks rather than forcing completion. Human researchers can still intervene, adjust the threshold, or decide to continue despite a failed review. However, the default behavior prioritizes quality control over productivity. This design reflects a broader principle: autonomous research agents should accelerate scientific work, but they should not bypass accountability, methodological rigor, or human responsibility.

\subsection{Limitations and Future Directions}\label{limitations-and-future-directions}

Several limitations merit acknowledgment. First, NORA\textquotesingle s quality is bound by the capabilities of its underlying LLMs; as models improve, the system\textquotesingle s output quality will correspondingly advance. Currently, lower performance in non-methodological-focused research, and niche research topics constrain the sophistication of the generated research. Second, the system\textquotesingle s spatial analysis decision frameworks, while comprehensive, inevitably encode a particular methodological perspective; researchers with different analytical traditions may find the default decision paths inappropriately constraining. Third, the adversarial review loop, while more rigorous than self-evaluation, cannot fully substitute for human domain expertise, particularly for assessing conceptual novelty and practical significance. Fourth, the system has been evaluated primarily on spatial regression and remote sensing tasks; its performance on other spatial research paradigms (e.g., human geography, qualitative analysis) remains untested.

Several directions for future work emerge. First, extending NORA\textquotesingle s skill library to additional spatial research paradigms. Second, incorporating multimodal reasoning to enable direct interpretation of maps, satellite imagery, and spatial visualizations. Third, developing collaborative modes where NORA works alongside human researchers in real-time rather than autonomously. Fourth, extending the evaluation framework to include longitudinal studies tracking the impact of NORA-assisted research on research productivity and methodological quality. Fifth, implementing new MCP tools for complicated data harvesting and mechanisms for self-evolving agents and skills. Sixth, enhancing user experience by adapting to multiple CLI tools and building user interfaces.

\subsection{Pathway toward Reproducibility and Replicability}\label{pathway-toward-reproducibility-and-replicability}

NORA's automation capacity provides a promising pathway toward more reproducible and replicable geospatial research. Once an autonomous research agent becomes reliable enough to support end-to-end scientific workflows, its role extends beyond task assistance; it begins to standardize how research is designed, executed, documented, and evaluated. This is particularly important in spatial data science, where analytical outcomes are often sensitive to data sources, spatial resolution, coordinate reference systems, preprocessing decisions, model parameters, and validation strategies. By encoding these decisions into explicit skills and workflow protocols, NORA can encourage more consistent methodological practices across studies. Moreover, the system can automatically generate detailed records of the research process, including data acquisition procedures, cleaning and transformation steps, spatial analysis choices, model training configurations, evaluation metrics, visualization settings, and intermediate outputs. Such automatic provenance tracking reduces the likelihood that critical methodological details are omitted from manuscripts or supplementary materials. It also makes the research process more transparent, auditable, and easier to rerun by other scholars. In the longer term, systems like NORA could support common benchmarking frameworks for geospatial research, enabling methods, datasets, and findings to be compared under shared standards. In this sense, autonomous research agents may become infrastructure for strengthening scientific accountability rather than merely tools for improving efficiency.

\section{Conclusion}\label{conclusion}

This paper introduced NORA, a harness-engineered autonomous research agent purpose-built for spatial data science. NORA successfully orchestrates the complete research lifecycle, from literature synthesis to adversarial peer review. Crucially, we demonstrated that formalizing harness engineering creates a reliable, reproducible, and auditable foundation for automated scientific workflows.

Our findings underscore that domain decomposition is essential for trustworthy spatial research. Through expert review and ablation analysis, we showed that NORA's domain-specialized architecture is advantageous. The spatial analysis unit effectively prevents common GIScience errors, such as inappropriate method selection and spatial autocorrelation leakage, while the data download unit enforces strict, reproducible data provenance. Furthermore, maintaining a strict separation between generation and evaluation ensures robust result-to-claim verification, actively preventing the propagation of unsupported claims.

Ultimately, NORA represents a critical step toward automating the mechanical aspects of spatial research, including routine analysis, data acquisition, and manuscript formatting. By absorbing these operational tasks, the system frees researchers to concentrate on the creative, conceptual, and ethical dimensions of geographic knowledge production. As the community grapples with the responsible integration of artificial intelligence and the ongoing demand for reproducibility, domain-specialized agents like NORA offer a principled path forward. This approach demonstrates that the future of automated scientific inquiry relies not just on raw model capability, but on the careful engineering of guardrails and quality assurance mechanisms.

\section{Data and Code Availability}\label{data-and-code-availability}

The data and code used in this manuscript, as well as the research agent itself can be accessed through this link: \url{https://github.com/GRIND-Lab-Core/night_owl_research_agent}

\textbf{Declaration of generative AI and AI-assisted technologies in the manuscript preparation process}

During the preparation of this work, the author(s) used ChatGPT and Gemini in order to correct grammatical mistakes and optimize writing. After using this tool/service, the author(s) reviewed and edited the content as needed and take(s) full responsibility for the content of the published article.

% --- References ------------------------------------------------------------
% The body uses author-year prose citations rather than \cite{} commands, so
% \nocite{*} forces every entry in references.bib to appear in the printed
% bibliography.
\nocite{*}
\bibliographystyle{elsarticle-harv}
\bibliography{references}

% =============================================================================
%  Appendices (converted from Appendix_Full_Final.docx).
% =============================================================================

\clearpage
\appendix
\section{Appendix A: NORA File System Design Principles}\label{appendix-a-nora-file-system-design-principles}

NORA decomposes the research workflow into discrete skills that communicate exclusively through canonical files on disk rather than through shared memory, yielding a strictly file-mediated dataflow. User-authored inputs (RESEARCH\_\allowbreak{}PLAN.\allowbreak{}md, BRIEF.md) seed the pipeline; lit-review then produces output/\allowbreak{}LIT\_\allowbreak{}REVIEW\_\allowbreak{}REPORT.\allowbreak{}md and output/paper-cache/, which idea-discovery-pipeline consumes to emit output/\allowbreak{}IDEA\_\allowbreak{}REPORT.\allowbreak{}md. Subsequent stages chain in the same producer--consumer pattern: experiment-design writes output/\allowbreak{}EXPERIMENT\_\allowbreak{}PLAN.\allowbreak{}md, deploy-experiment appends to output/\allowbreak{}EXPERIMENT\_\allowbreak{}LOG.\allowbreak{}md, spatial-analysis emits output/spatial-analysis/, generate-report consolidates these into output/\allowbreak{}NARRATIVE\_\allowbreak{}REPORT.\allowbreak{}md, and the paper-writing pipeline (paper-plan $\rightarrow$ paper-figure-generate $\rightarrow$ paper-draft $\rightarrow$ paper-review-loop $\rightarrow$ paper-covert) transforms the narrative into output/\allowbreak{}PAPER\_\allowbreak{}PLAN.\allowbreak{}md, output/figures/, output/manuscript/, and finally output/papers/. Three cross-cutting artifacts thread the full graph: output/\allowbreak{}PROJ\_\allowbreak{}NOTES.\allowbreak{}md (append-only discovery log), memory/\allowbreak{}MEMORY.\allowbreak{}md (session state), and handoff.json (context-reset payload written by the stop hook). This canonical-path contract makes every transition inspectable, every stage independently resumable, and the entire workflow composable --- any skill can be re-invoked against its declared inputs without replaying upstream work.

\begin{figure}[htbp]
\centering
\includegraphics[width=\linewidth]{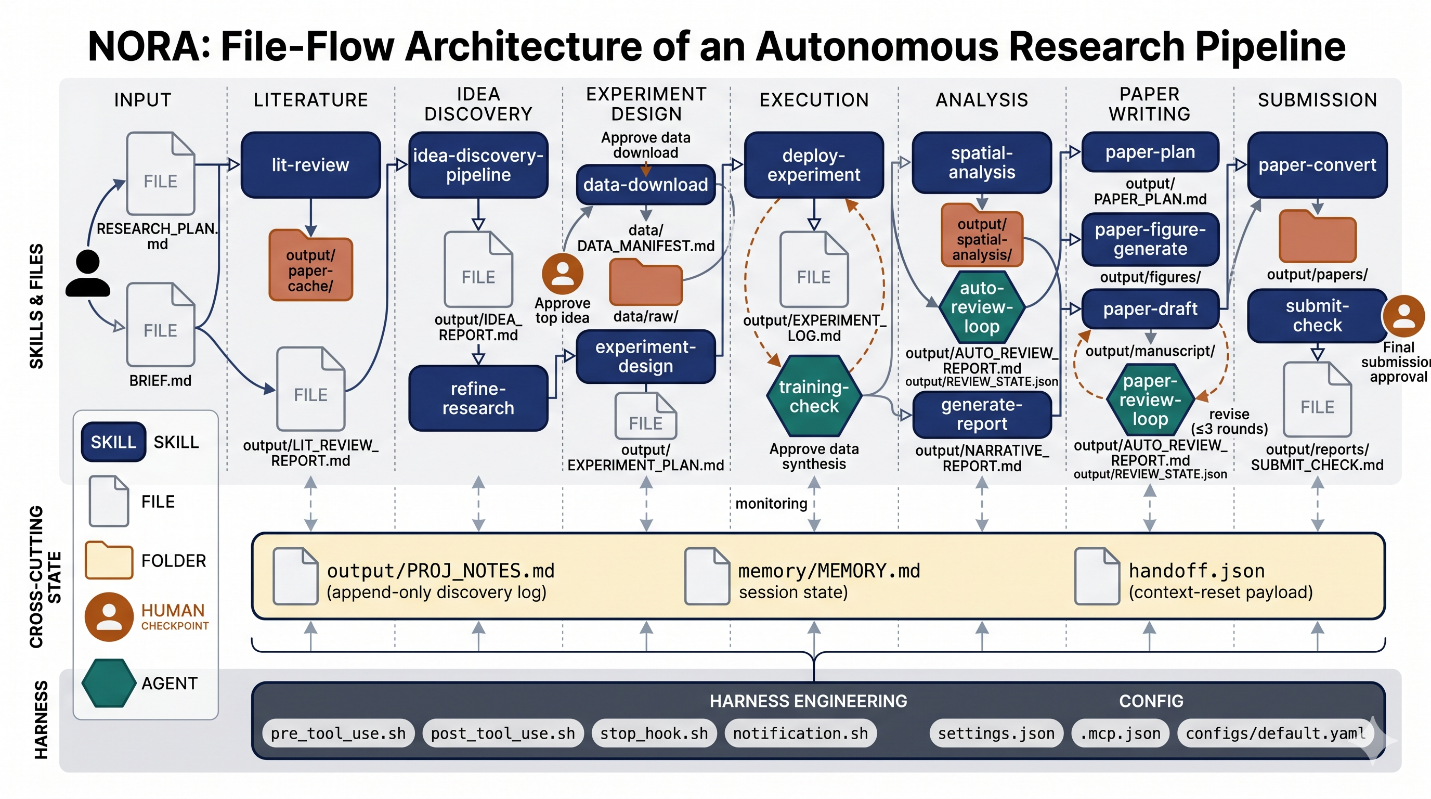}
\caption{File generated by NORA and how files pass through workflows and skills.}
\label{fig:appx-a-1}
\end{figure}

\section{Appendix B: Agent Design Principles}\label{appendix-b-agent-design-principles}

A structural discipline for authoring, reviewing, and evolving Claude Code specialist sub-agents within a multi-stage research orchestration system.

Agents in NORA are defined as Markdown files under .claude/agents/. Unlike skills --- which describe workflow logic that runs inside the parent conversation context --- an agent is a \textbf{specialist persona invoked via the Agent tool in its own context window}. That isolation is the entire point: agents let the pipeline parallelise independent subtasks, protect the parent context from tool-call noise, and enforce the generator--evaluator separation that NORA relies on for research integrity.

The nine agents shipped in this project (orchestrator, literature-scout, synthesis-analyst, gap-finder, hypothesis-generator, geo-specialist, paper-writer, peer-reviewer, citation-manager) were designed against the principles below.

\subsection{Part I: Required Sections of an Agent}\label{part-i-required-sections-of-an-agent}

Every agent file in .claude/agents/ must contain the sections below. Each section exists for a specific reason rooted in the operational demands of multi-agent research pipelines.

\paragraph*{YAML Frontmatter}

-\/-\/-\\
name\textbf{:} \textless agent-name\textgreater{}\\
description\textbf{:} \textbar{}\\
\textless One-paragraph statement of when to use this agent.\textgreater{}\\
Use this agent to:\\
- \textless capability 1\textgreater{}\\
- \textless capability 2\textgreater{}\\
tools\textbf{:} \textless explicit tool list --- never "all" unless orchestrator\textgreater{}\\
-\/-\/-

\textbf{Why required.} The frontmatter is the machine-readable contract that the Claude Code harness and parent skills rely on for routing and permissioning:

\begin{itemize}
\item
  \textbf{Name and description} --- parent skills select the correct agent by matching the description against the subtask; vague descriptions cause mis-routing.
\item
  \textbf{Explicit tool allowlist} --- agents receive the tightest allowlists in the system (e.g., peer-reviewer has Read, Write only; literature-scout has WebFetch, WebSearch, Read, Write, Bash). This is stricter than skills because agents run unattended in a separate context and cannot be interrupted by a user gate during execution.
\item
  \textbf{``Use this agent to'' bullet list} --- a structured capability menu the orchestrator uses to decide whether the agent matches the task.
\end{itemize}

\textbf{Design failure without it.} A missing or vague description causes the parent skill to invoke the wrong agent; a missing tool allowlist gives the sub-context unbounded authority to write files, run shell commands, or call external APIs the caller never intended.

\paragraph*{Persona and Role Statement}

A 2--4 sentence block declaring \emph{who} the agent is and what scope it owns.

``You are the domain expert for all spatial analysis decisions in NORA. You are consulted before methodology and results sections are written, and whenever spatial analysis code is generated.'' --- geo-specialist.md

\textbf{Why required.} Agents run in a fresh context with none of the parent's conversational history. The persona statement is the first thing the sub-context reads; it seeds role-consistent behaviour for the entire invocation. It also disambiguates overlap --- the orchestrator can reason about the boundary between synthesis-analyst (reads papers, writes synthesis) and gap-finder (reads synthesis, writes gaps) because each persona explicitly names its scope.

\textbf{Design failure without it.} Without a clear persona, the sub-context behaves as a generic assistant and produces output that looks plausible but fails the role's discipline (a reviewer that drafts instead of critiques, a writer that self-approves, a scout that synthesises).

\paragraph*{Context Isolation Contract}

An explicit statement of what the agent reads from disk on startup and what it does \textbf{not} assume from the caller.

\textbf{Why required.} The parent skill's context is invisible to the agent. Two corollaries:

\begin{enumerate}
\def\labelenumi{\arabic{enumi}.}
\item
  \textbf{Everything the agent needs must be fetchable from files} (e.g., program.md, memory/paper-cache/, output/\allowbreak{}LIT\_\allowbreak{}REVIEW\_\allowbreak{}REPORT.\allowbreak{}md) or passed explicitly as the \$ARGUMENTS payload at invocation time.
\item
  \textbf{Anything the agent produces must land in a canonical output path} because the parent has no way to ``read back'' the sub-context --- only the final tool-return payload is visible.
\end{enumerate}

This is the architectural reason NORA is file-driven: agents are stateless and context-isolated by design, so persistent state must live on disk.

\textbf{Design failure without it.} The agent silently assumes the caller passed context that was never actually in its window, hallucinates missing details, and produces confidently wrong output.

\paragraph*{Single-Responsibility Scope}

A declaration that the agent performs one discrete subtask --- not a pipeline.

\textbf{Why required.} Agents are cheap to compose but expensive to mis-scope. An agent that ``does literature search and synthesis and gap analysis'' collapses three specialists into one, loses the parallelism advantage (search and synthesis can overlap), and defeats generator--evaluator separation (the same context that summarises cannot objectively critique its own summary).

NORA's nine agents each own exactly one pipeline stage:

\begin{tabularx}{\linewidth}{@{}>{\hsize=0.6584\hsize\raggedright\arraybackslash}X>{\hsize=1.3416\hsize\raggedright\arraybackslash}X@{}}
\toprule\noalign{}
Agent & Single responsibility \\
\midrule\noalign{}
\endhead
\bottomrule\noalign{}
\endlastfoot
literature-scout & Retrieve and cache papers \\
synthesis-analyst & Read cache, write synthesis \\
gap-finder & Read synthesis, write ranked gaps \\
hypothesis-generator & Read gaps, write scored hypotheses \\
geo-specialist & Review spatial choices, recommend methods \\
paper-writer & Draft one section, self-score \\
peer-reviewer & Cold-read critique, no drafting \\
citation-manager & Format and validate references \\
orchestrator & Sequence the above, never do their work \\
\end{tabularx}

\textbf{Design failure without it.} Composite agents are harder to test, harder to debug, and harder to run in parallel. Scope creep inside one agent produces output whose quality is dominated by the weakest sub-task.

\paragraph*{Narrow Tool Allowlist (Least Privilege)}

An explicit tools: field listing only the tools the agent actually needs.

\textbf{Why required.} Because agents run unattended, the blast radius of any single agent call is bounded only by the allowlist. A reviewer that can Write anywhere can corrupt the manuscript it is reviewing; a scout that can Bash freely can run arbitrary shell commands outside the retrieval workflow.

Observed allowlist discipline in NORA:

\begin{itemize}
\item
  peer-reviewer and gap-finder --- Read, Write only (no network, no shell).
\item
  literature-scout --- WebFetch, WebSearch, Read, Write, Bash (network + cache management).
\item
  paper-writer --- Read, Write (deliberately no WebSearch --- writer must not improvise citations).
\item
  orchestrator --- all (the sole exception, because it must invoke every downstream tool).
\end{itemize}

\textbf{Design failure without it.} Tool-allowlist creep turns specialist agents into generalist assistants, erasing both the safety and the composability that made the agent pattern worth adopting.

\paragraph*{Deterministic I/O Contract}

A specification of the structured input the agent expects and the structured output it returns.

\textbf{Why required.} Agents are called programmatically by skills and by the orchestrator. The caller must be able to predict the shape of the return payload well enough to parse it and feed it downstream. NORA agents define their output schema inline (e.g., literature-scout returns a JSON object per paper with id, title, authors, year, venue, abstract, citation\_count, doi, url, domain\_tags, priority\_venue).

Three contracts must be explicit:

\begin{enumerate}
\def\labelenumi{\arabic{enumi}.}
\setcounter{enumi}{2}
\item
  \textbf{Inputs} --- what \$ARGUMENTS must contain (keywords, section name, scoring rubric).
\item
  \textbf{Files read} --- which disk paths are consulted (handoff.json, APPROVED\_CLAIMS, paper-cache).
\item
  \textbf{Outputs} --- canonical file writes \emph{plus} the structured return payload the caller parses.
\end{enumerate}

\textbf{Design failure without it.} The caller receives freeform prose it cannot reliably parse, forcing an LLM re-read of the agent's output to extract structure --- wasting tokens and reintroducing the context pollution the agent was supposed to prevent.

\paragraph*{Evaluator-vs-Producer Role Lock}

A rule declaring whether the agent \textbf{produces} content or \textbf{evaluates} content --- never both.

\textbf{Why required.} This is the agent-side enforcement of the generator--evaluator separation principle. NORA splits this role explicitly:

\begin{itemize}
\item
  \textbf{Producers} --- paper-writer, synthesis-analyst, gap-finder, hypothesis-generator, literature-scout, citation-manager.
\item
  \textbf{Evaluators} --- peer-reviewer, geo-specialist (as reviewer), plus Codex MCP (gpt-5.4) when invoked from a skill.
\end{itemize}

A producer agent that self-scores is allowed \textbf{only} for internal gating (e.g., paper-writer's 5-dimension rubric is a triage signal, not an acceptance decision). Any score that actually opens a gate must come from an evaluator agent or external reviewer invoked in a separate context.

\textbf{Design failure without it.} Refinement loops converge on outputs that score well against the producer's own priors but fail external scrutiny --- a mode collapse the pipeline is specifically engineered to prevent.

\paragraph*{Persona-Grounded Expertise}

An embedded block of domain knowledge the agent references on every invocation.

\textbf{Why required.} Agents start cold. If they had to re-derive domain conventions every run, output quality would be inconsistent and token spend would explode. NORA bakes domain priors directly into the agent body:

\begin{itemize}
\item
  geo-specialist lists CRS rules, dataset recommendations per domain, and spatial-method decision rules.
\item
  paper-writer lists section-specific guidelines and geo conventions (``state spatial resolution for all raster data'', ``CRS specified in methods'').
\item
  citation-manager lists venue-specific formatting deltas (APA vs.~IEEE vs.~ACM).
\item
  hypothesis-generator lists geo-specific hypothesis templates.
\end{itemize}

These blocks are the agent's ``always-on'' memory, compensating for the absence of conversational history.

\textbf{Design failure without it.} The agent falls back on generic best-practice, missing the discipline-specific checks (Moran's I on residuals, spatial CV vs.~random CV, EPSG codes for analysis vs.~storage vs.~display) that distinguish publishable research from plausible-sounding output.

\paragraph*{Structured Output Format (Return Schema)}

A concrete Markdown/JSON template for the final return payload, with named fields and example values.

\textbf{Why required.} The parent skill needs to locate specific values (a score, a top-ranked gap, a decision verdict) without re-parsing prose. NORA agents declare these templates literally --- the reviewer's Editor Summary block, the synthesis analyst's matrix columns, the hypothesis generator's Overall: X.X/10 line, the writer's Score: X.X (N:X, R:X, L:X, C:X, I:X) signature.

The template serves three consumers:

\begin{enumerate}
\def\labelenumi{\arabic{enumi}.}
\setcounter{enumi}{5}
\item
  \textbf{The agent itself} --- a checklist of what must appear in its return.
\item
  \textbf{The parent skill} --- a parsing anchor.
\item
  \textbf{Future humans} --- a documented expectation when outputs need audit.
\end{enumerate}

\textbf{Design failure without it.} Output formats drift between invocations; parser regex and downstream skills break every time the agent's phrasing shifts.

\paragraph*{Cold-Read Discipline (Evaluator Agents Only)}

For reviewer agents, an explicit instruction to evaluate the artifact without the author's framing.

\textbf{Why required.} The peer-reviewer agent simulates three external reviewers (methods expert, domain specialist, applications reviewer). Its value is precisely that the sub-context has no memory of how the draft was produced. Cold-read discipline means the agent reads the artifact on disk and produces a verdict as if encountering the work for the first time.

\textbf{Design failure without it.} The reviewer inherits the author's justifications from the parent context, converging on an agreeable review that misses the blind spots a real reviewer would surface.

\paragraph*{Cost and Parallelism Awareness}

A note on whether the agent can be invoked in parallel with others and an estimate of its typical token footprint.

\textbf{Why required.} Each agent invocation spawns a sub-context billed independently. Orchestrator skills must know:

\begin{itemize}
\item
  \textbf{Independent agents} --- literature-scout and geo-specialist (for dataset suggestions) can run concurrently; the orchestrator should dispatch them in a single message with multiple Agent tool calls.
\item
  \textbf{Sequential-only agents} --- synthesis-analyst must finish before gap-finder starts; paper-writer must finish before peer-reviewer.
\item
  \textbf{Heavy agents} --- synthesis-analyst reads the entire paper-cache and can consume 50k+ input tokens; the orchestrator should batch its calls.
\end{itemize}

\textbf{Design failure without it.} The orchestrator serialises work that could run in parallel (wasting wall-clock time) or parallelises work with data dependencies (producing inconsistent artifacts).

\paragraph*{Error and Degradation Handling}

An explicit list of failure modes and the agent's prescribed response.

\textbf{Why required.} Agents cannot ask the user for help mid-run. Every foreseeable failure must have a deterministic response documented in the agent body:

\begin{itemize}
\item
  literature-scout --- ``If \textless{} 20 papers found: broaden keywords (add synonyms, relax year filter to 2015+) and retry.''
\item
  orchestrator --- ``API failure in literature-scout: log, retry once, fall back to cached papers.''
\item
  paper-writer --- escalation rubric mapping self-score ranges to revision directives (Major / Moderate / Minor).
\end{itemize}

\textbf{Design failure without it.} The agent halts silently on the first unfamiliar failure, leaving the caller with an empty return and no diagnostic trail.

\paragraph*{Audit Contribution}

Every agent must either append to output/\allowbreak{}PROJ\_\allowbreak{}NOTES.\allowbreak{}md itself or return enough metadata for the parent skill to log on its behalf.

\textbf{Why required.} The audit trail established at the skill level must extend into agent calls; otherwise the provenance chain breaks. When a reviewer later asks ``which agent generated this section?'', the log must answer unambiguously.

\textbf{Design failure without it.} Artifacts arrive in output/ with no record of which agent in which invocation produced them, making regression impossible when an agent's prompt is later tuned.

\subsection{Part II: Advanced Design Disciplines for Agent Optimisation}\label{part-ii-advanced-design-disciplines-for-agent-optimisation}

The required sections above make agents safe and composable. The patterns below make them fast, robust, and evolvable.

\paragraph*{A. Agent Contract Tests}

\textbf{Current state.} Agent I/O contracts are described in prose and verified by running the full pipeline.

\textbf{Optimisation.} Add a .claude/\allowbreak{}agents/\allowbreak{}tests/\textless agent\textgreater/ directory with synthetic input fixtures and expected-output schemas:

.claude/\allowbreak{}agents/\allowbreak{}tests/\allowbreak{}peer-reviewer/\\
+-- fixture\_\allowbreak{}draft\_\allowbreak{}strong.\allowbreak{}md $\rightarrow$ expect Accept / Minor Revision\\
+-- fixture\_\allowbreak{}draft\_\allowbreak{}weak\_\allowbreak{}methods.\allowbreak{}md $\rightarrow$ expect Major Revision, flag methods\\
+-- fixture\_\allowbreak{}no\_\allowbreak{}results.\allowbreak{}md $\rightarrow$ expect Major Revision, flag results

\textbf{Why.} Contract tests catch prompt regressions (an edit that silently changes the return schema) before they reach the pipeline. They also document expected behaviour at boundary conditions, which prose alone cannot.

\paragraph*{B. Parallel Invocation Manifest}

\textbf{Current state.} Parallelism is implicit --- the orchestrator decides per-stage whether to dispatch agents concurrently.

\textbf{Optimisation.} Each agent declares its parallelism class in frontmatter:

parallelism\textbf{:}\\
class\textbf{:} independent \emph{\# independent \textbar{} depends\_on: {[}\textless agent\textgreater{]} \textbar{} global\_singleton}\\
max\_concurrent\textbf{:} 3

\textbf{Why.} The orchestrator can validate a parallel dispatch plan against declared classes before invoking, catching dependency violations statically instead of after a failed run.

\paragraph*{C. Return-Payload JSON Schema}

\textbf{Current state.} Return formats are documented as Markdown templates with example values.

\textbf{Optimisation.} Ship a JSON Schema alongside each agent:

.claude/\allowbreak{}agents/\allowbreak{}schemas/\allowbreak{}peer-reviewer.\allowbreak{}schema.\allowbreak{}json

\textbf{Why.} Parent skills can validate agent output programmatically, rejecting malformed returns with a clear diagnostic instead of silently consuming a partial payload. Schemas also enable stricter downstream parsing without LLM fallback.

\paragraph*{D. Persona Composition over Inheritance}

\textbf{Current state.} Each agent's persona and expertise are embedded verbatim in its body.

\textbf{Optimisation.} Extract shared domain knowledge into reusable blocks under .claude/agents/\_shared/:

.claude/agents/\_shared/\allowbreak{}geo\_\allowbreak{}conventions.\allowbreak{}md $\rightarrow$ included by geo-specialist, paper-writer, synthesis-analyst\\
.claude/agents/\_shared/\allowbreak{}apa7\_\allowbreak{}rules.\allowbreak{}md $\rightarrow$ included by citation-manager, paper-writer

\textbf{Why.} Domain rules that live in three agents drift apart over time. Shared blocks keep the rules synchronised and make a single edit propagate to every consumer.

\paragraph*{E. Cold-Read Assertion}

\textbf{Current state.} Reviewer agents rely on instructions to ignore author context.

\textbf{Optimisation.} The orchestrator strips any author-supplied rationale from the payload before invoking a reviewer, and the reviewer asserts author\_\allowbreak{}context\_\allowbreak{}present == false before proceeding. If the assertion fails, the review halts with a diagnostic instead of producing a tainted review.

\textbf{Why.} The cold-read property must be structurally enforced, not just prompted. A prompt rule can be overridden by a chatty caller; a payload check cannot.

\paragraph*{F. Token-Budget Declaration}

\textbf{Current state.} Token footprint is implicit; cost surprises surface only in post-run accounting.

\textbf{Optimisation.} Each agent declares an expected input/output token range in frontmatter, and the orchestrator refuses to invoke it when the assembled payload exceeds the upper bound.

token\_budget\textbf{:}\\
input\_max\textbf{:} 80000\\
output\_max\textbf{:} 4000

\textbf{Why.} Prevents runaway dispatches that would time out or blow the context window. Makes per-agent cost legible in the orchestrator's pre-flight planning.

\paragraph*{G. Reviewer Ensemble for High-Stakes Gates}

\textbf{Current state.} peer-reviewer simulates three reviewers inside one sub-context.

\textbf{Optimisation.} For high-stakes gates (final paper acceptance, contract-violation calls), dispatch three independent peer-reviewer invocations in parallel sub-contexts and aggregate. Mid-stakes gates keep the single-invocation triple-reviewer persona.

\textbf{Why.} Three independent sub-contexts cannot influence each other; a single sub-context simulating three reviewers risks internal consensus bias. The cost is 3x review tokens, paid only where the gate is worth it.

\paragraph*{H. Reviewer Memory (Adversarial Continuity)}

\textbf{Current state.} Agent invocations are stateless; reviewer memory lives only in files (memory/\allowbreak{}REVIEWER\_\allowbreak{}MEMORY.\allowbreak{}md) when the calling skill persists it.

\textbf{Optimisation.} Reviewer agents explicitly read and append to a persistent reviewer memory file, so suspicions raised in round N survive into round N+1 even across sessions.

\textbf{Why.} The auto-review-loop skill already supports this at the skill level for Codex MCP. Pushing it into the agent body extends the same adversarial continuity to local Claude-subagent reviewers, closing a quality gap in the fallback path.

\paragraph*{I. Capability Probe Instead of Static Allowlists}

\textbf{Current state.} Agents declare tool allowlists statically.

\textbf{Optimisation.} Agents additionally declare capability tags (retrieves\_papers, formats\_citations, reviews\_\allowbreak{}spatial\_\allowbreak{}methods). Orchestrators select agents by capability, not by name, allowing drop-in replacement when a better agent exists.

\textbf{Why.} Decouples orchestrators from agent names, so swapping literature-scout for a future semantic-scout-v2 requires changing zero caller code.

\paragraph*{J. Agent Versioning and Deprecation}

\textbf{Current state.} Agents are edited in place; prior versions are reachable only through git history.

\textbf{Optimisation.} Agents carry a version field and support side-by-side coexistence (peer-reviewer@1, peer-reviewer@2). The orchestrator pins a version per pipeline run.

\textbf{Why.} Prompt edits to a reviewer agent silently change every downstream score. Versioning makes prompt evolution auditable and lets experimental prompts ship without disturbing the production pipeline.

\subsection{Part III: Concise Reference Table}\label{part-iii-concise-reference-table}

\paragraph*{Required Sections}

\begin{tabularx}{\linewidth}{@{}>{\hsize=0.2668\hsize\raggedright\arraybackslash}X>{\hsize=0.8000\hsize\raggedright\arraybackslash}X>{\hsize=0.8000\hsize\raggedright\arraybackslash}X>{\hsize=2.1332\hsize\raggedright\arraybackslash}X@{}}
\toprule\noalign{}
\# & Section & Purpose & Failure Mode If Missing \\
\midrule\noalign{}
\endhead
\bottomrule\noalign{}
\endlastfoot
1 & YAML Frontmatter & Name, description, tool allowlist --- the machine-readable contract & Mis-routing; unbounded tool access in an unattended sub-context \\
2 & Persona and Role Statement & Seeds role-consistent behaviour in a fresh context & Generic assistant behaviour; role overlap \\
3 & Context Isolation Contract & Declares what the agent reads from disk vs.~expects from args & Hallucinated context; confidently wrong output \\
4 & Single-Responsibility Scope & One discrete subtask per agent & Composite agents that cannot be parallelised or audited \\
5 & Narrow Tool Allowlist & Least-privilege enforcement per agent & Allowlist creep; reviewer that overwrites what it reviews \\
6 & Deterministic I/O Contract & Structured input and output shape & Freeform prose that downstream skills cannot parse \\
7 & Evaluator-vs-Producer Role Lock & Enforces generator--evaluator separation at the agent layer & Self-validating loops; mode collapse \\
8 & Persona-Grounded Expertise & Embedded domain priors (CRS, APA, spatial methods) & Generic output that misses discipline-specific checks \\
9 & Structured Output Format & Named fields, parser anchors, audit template & Format drift between runs; broken downstream parsers \\
10 & Cold-Read Discipline (evaluators) & Reviewer sees artifact without author rationale & Agreeable reviews that inherit author blind spots \\
11 & Cost and Parallelism Awareness & Declares whether agent can run concurrently and its typical budget & Wasted wall-clock or data-race artifacts \\
12 & Error and Degradation Handling & Deterministic responses to foreseeable failures & Silent halts with no diagnostic trail \\
13 & Audit Contribution & PROJ\_\allowbreak{}NOTES.\allowbreak{}md entry or metadata for caller to log & Broken provenance chain across the skill--agent boundary \\
\end{tabularx}

\paragraph*{Advanced Optimisations}

\begin{tabularx}{\linewidth}{@{}>{\hsize=0.3552\hsize\raggedright\arraybackslash}X>{\hsize=1.2447\hsize\raggedright\arraybackslash}X>{\hsize=1.2447\hsize\raggedright\arraybackslash}X>{\hsize=1.1555\hsize\raggedright\arraybackslash}X@{}}
\toprule\noalign{}
ID & Optimisation & What It Adds & Key Benefit \\
\midrule\noalign{}
\endhead
\bottomrule\noalign{}
\endlastfoot
A & Agent Contract Tests & Synthetic fixtures + expected output schemas per agent & Catches prompt regressions before they reach the pipeline \\
B & Parallel Invocation Manifest & Declared parallelism class and max\_concurrent & Static validation of dispatch plans \\
C & Return-Payload JSON Schema & Machine-checkable output schema per agent & Programmatic parser validation; cleaner downstream code \\
D & Persona Composition & Shared domain blocks under \_shared/ & Rules stay synchronised across agents \\
E & Cold-Read Assertion & Orchestrator strips author rationale; reviewer asserts absence & Structural (not just prompted) enforcement of cold-read \\
F & Token-Budget Declaration & Declared input/output token bounds per agent & Pre-flight cost control; no runaway dispatches \\
G & Reviewer Ensemble & Three independent parallel reviewer sub-contexts & Eliminates intra-context consensus bias at high-stakes gates \\
H & Reviewer Memory & Persistent reviewer suspicions across rounds & Adversarial continuity for local-fallback reviewers \\
I & Capability Probes & Select agents by capability tags, not by name & Drop-in agent replacement without caller changes \\
J & Agent Versioning & Side-by-side versioned agents with pipeline pinning & Auditable prompt evolution; safe experimentation \\
\end{tabularx}

\paragraph*{Part IV: How Agent Principles Relate to Skill Principles}

Skills and agents are complementary, not redundant. The relationship is:

\begin{tabularx}{\linewidth}{@{}>{\hsize=1.0000\hsize\raggedright\arraybackslash}X>{\hsize=1.0000\hsize\raggedright\arraybackslash}X>{\hsize=1.0000\hsize\raggedright\arraybackslash}X@{}}
\toprule\noalign{}
Concern & Skill layer & Agent layer \\
\midrule\noalign{}
\endhead
\bottomrule\noalign{}
\endlastfoot
Context & Runs in parent context & Runs in isolated sub-context \\
Composition & Chains other skills and agents & Is a leaf --- does not call other agents \\
State & Reads and writes canonical files; resumable via checkpoints & Stateless per invocation; disk is the only memory \\
Scope & A stage of the pipeline (many subtasks) & One discrete subtask inside a stage \\
Tool allowlist & Broad (workflow-level) & Narrow (task-level) \\
Parallelism & One skill at a time in the parent context & Multiple agents in parallel sub-contexts \\
Evaluator role & Orchestrates evaluator invocation & Can \emph{be} the evaluator \\
Audit & Appends to PROJ\_\allowbreak{}NOTES.\allowbreak{}md & Contributes metadata for the skill to log \\
\end{tabularx}

Skills encode \textbf{workflow intent}. Agents encode \textbf{specialist execution}. The harness (hooks, handoff.json, memory files) encodes \textbf{guarantees} that span both. Together they form the three-layer discipline that makes NORA's long-horizon autonomous research pipeline tractable.

\emph{This document codifies the design discipline observed across all nine specialist agents of the Night Owl Research Agent. It should be treated as a living standard --- updated as new agents are added and as advanced optimisations are adopted.}

\section{Appendix C: Skill Design Principles}\label{appendix-c-skill-design-principles}

\begin{quote}
A structural discipline for authoring, reviewing, and evolving Claude Code skills within a multi-stage research orchestration system.
\end{quote}

\subsection{Part I: Required Sections of a Skill}\label{part-i-required-sections-of-a-skill}

Every SKILL.md file in this project must contain the sections below. Each section exists for a specific reason rooted in the operational demands of long-running, multi-agent research pipelines.

\subsubsection{YAML Frontmatter}\label{yaml-frontmatter-1}

\begin{verbatim}
---
name: <skill-name>
description: <one-liner with primary use cases>
argument-hint: [expected arguments]
tools: [allowed tool list]
flags: [optional configuration flags]
---
\end{verbatim}

\textbf{Why required.} The frontmatter is the machine-readable contract between the skill and the Claude Code harness. It controls: - \textbf{Tool allowlisting} --- limits the blast radius of any single skill invocation. A literature-review skill has no business calling \texttt{Write} on experiment logs. - \textbf{Argument parsing} --- tells the orchestrator (and the user) what inputs the skill expects before it runs. - \textbf{Discovery} --- pipeline orchestrators (\texttt{idea-discovery-pipeline}, \texttt{paper-writing-pipeline}, \texttt{full-pipeline}) use \texttt{name} and \texttt{description} to route work to the correct skill. - \textbf{Flags} --- enable opt-in behaviors (e.g., \texttt{hard\_mode}, \texttt{nightmare\_mode} in \texttt{auto-review-loop}) without forking the skill into variants.

\textbf{Design failure without it.} A skill without explicit tool allowlisting can read, write, and execute anything --- violating the principle of least privilege. A skill without \texttt{argument-hint} forces users to read the full body to understand how to invoke it.

\subsubsection{Overview / Purpose Statement}\label{overview-purpose-statement}

A 2--5 sentence block explaining \emph{what} the skill does, \emph{when} to use it, and \emph{where} it fits in the pipeline.

\textbf{Why required.} Skills are composed into pipelines by both humans and orchestrator skills. The overview is the only section that gets read at routing time. It must answer three questions in under 30 seconds: 1. What does this skill produce? 2. What does it consume (upstream dependencies)? 3. When should I reach for this skill instead of a neighboring one?

\textbf{Design failure without it.} Without a clear purpose boundary, skills drift into overlapping scope. The distinction between \texttt{generate-idea} (brainstorm divergently) and \texttt{refine-research} (converge on one proposal) would collapse, causing redundant work and contradictory outputs.

\subsubsection{Constants / Configurable Thresholds}\label{constants-configurable-thresholds}

A named block of tuneable parameters with default values.

\begin{verbatim}
MAX_ROUNDS = 5
SCORE_THRESHOLD = 9
MAX_PRIMARY_CLAIMS = 2
MAX_NEW_TRAINABLE_COMPONENTS = 2
\end{verbatim}

\textbf{Why required.} Constants serve three purposes: 1. \textbf{Parsimony enforcement} --- caps like \texttt{MAX\_\allowbreak{}PRIMARY\_\allowbreak{}CLAIMS\ =\ 2} prevent scope creep within a single skill run. 2. \textbf{Termination guarantees} --- \texttt{MAX\_ROUNDS} ensures refinement loops halt even when the score threshold is never met, preventing infinite token burn. 3. \textbf{Reproducibility} --- when a colleague runs the same skill, the constants document the operating envelope. Changing \texttt{SCORE\_THRESHOLD} from 9 to 7 is a conscious, visible decision rather than a buried prompt tweak.

\textbf{Design failure without it.} Magic numbers scattered through prose are invisible to reviewers. A refinement loop without \texttt{MAX\_ROUNDS} can run indefinitely. A generation step without output caps produces sprawling, unfocused artifacts.

\subsubsection{Workflow / Phases}\label{workflow-phases}

A numbered sequence of phases, each with: - \textbf{Entry conditions} (what files/state must exist) - \textbf{Actions} (what the skill does) - \textbf{Exit conditions} (what is produced, what state is written)

\textbf{Why required.} Research workflows are long-running (hours to days) and frequently interrupted. Phase structure provides: 1. \textbf{Resumability} --- checkpoint state files (e.g., \texttt{REFINE\_\allowbreak{}STATE.\allowbreak{}json}, \texttt{REVIEW\_\allowbreak{}STATE.\allowbreak{}json}) record the last completed phase. On resume, the skill skips completed phases. 2. \textbf{Debuggability} --- when output quality is poor, the phase structure localizes the failure. ``The problem is in Phase 3 (scoring)'' is actionable; ``the skill produced bad output'' is not. 3. \textbf{Composability} --- pipeline orchestrators can skip phases that another skill already completed (e.g., \texttt{paper-writing-pipeline} skips the planning phase if \texttt{PAPER\_\allowbreak{}PLAN.\allowbreak{}md} already exists).

\textbf{Design failure without it.} A monolithic skill that ``does everything in one pass'' cannot be resumed, cannot be partially reused, and cannot be debugged without re-running the entire sequence.

\subsubsection{Checkpoint / State Persistence}\label{checkpoint-state-persistence}

Explicit specification of: - State file path and schema (JSON with phase, round, threadId, scores, status, timestamp) - Recovery logic (how to detect and resume from a prior checkpoint) - Expiry rules (e.g., checkpoints older than 24 hours are stale)

\textbf{Why required.} Claude Code sessions are stateless across invocations. Without explicit state persistence: - A 5-round refinement loop that crashes at round 3 must restart from round 1. - An adversarial review loop loses the reviewer's accumulated memory. - Pipeline orchestrators cannot determine which skill last completed successfully.

The \texttt{handoff.json} pattern (used by pipeline orchestrators) is the inter-skill equivalent: it records \texttt{pipeline.stage} and \texttt{recovery.\allowbreak{}resume\_\allowbreak{}skill} so the full pipeline can resume from the correct point.

\textbf{Design failure without it.} Every interruption (timeout, token limit, user break) forces a full restart, wasting compute and producing inconsistent artifacts.

\subsubsection{Canonical Output Paths}\label{canonical-output-paths}

A list of every file the skill writes, with fixed paths relative to the project root.

\textbf{Why required.} Skills are chained: \texttt{refine-research} writes \texttt{output/\allowbreak{}refine-logs/\allowbreak{}FINAL\_\allowbreak{}PROPOSAL.\allowbreak{}md}, which \texttt{experiment-design} reads. If output paths are ad-hoc or user-chosen: - Downstream skills cannot find upstream artifacts. - The \texttt{generate-report} skill (which consolidates all outputs into \texttt{NARRATIVE\_\allowbreak{}REPORT.\allowbreak{}md}) cannot locate its sources. - \texttt{PROJ\_\allowbreak{}NOTES.\allowbreak{}md} entries become meaningless without stable path references.

The canonical path convention also enables the memory/audit system: \texttt{APPROVED\_\allowbreak{}CLAIMS.\allowbreak{}md}, \texttt{DATA\_\allowbreak{}MANIFEST.\allowbreak{}md}, and \texttt{PROJ\_\allowbreak{}NOTES.\allowbreak{}md} are always in known locations.

\textbf{Design failure without it.} Every pipeline run requires manual path wiring, breaking automation and introducing silent failures when a downstream skill reads a stale file from a previous run.

\subsubsection{Decision Rules / Branching Logic}\label{decision-rules-branching-logic}

Explicit if/then rules for choosing between alternatives within the skill.

\textbf{Why required.} Research skills frequently face forks: - \texttt{deploy-experiment}: Track A (ML/DL) vs.~Track B (spatial/GIScience) vs.~local vs.~SSH vs.~Modal GPU. - \texttt{spatial-analysis}: ESDA $\rightarrow$ regression ladder $\rightarrow$ GWR/MGWR $\rightarrow$ clustering (method selection tree). - \texttt{paper-figure-generate}: code-generated (plots, maps) vs.~prompt-generated (architecture diagrams).

Without explicit decision rules, the LLM must infer the correct branch from context, which is unreliable across sessions. Decision rules make the skill deterministic at branch points.

\textbf{Design failure without it.} The skill makes inconsistent choices across runs. One invocation uses GWR; the next uses OLS for the same data, because the prompt context differed slightly.

\subsubsection{Guardrails / ``Do NOT'' Rules}\label{guardrails-do-not-rules}

An explicit list of prohibited behaviors.

\textbf{Why required.} LLMs have systematic failure modes that must be explicitly suppressed: - \textbf{Fabrication}: ``Do NOT invent statistics, p-values, or citation details.'' - \textbf{Self-scoring}: ``The entity that writes NEVER scores its own output.'' - \textbf{Scope creep}: ``Do NOT add features beyond the experiment plan.'' - \textbf{Silent failure}: ``Do NOT skip a failed step; log the failure and halt.''

Guardrails are not aspirational --- they are countermeasures against observed failure patterns. Each rule exists because the skill (or a similar skill) produced that exact failure in testing.

\textbf{Design failure without it.} The skill produces plausible but fabricated results, self-validates its own output, or silently degrades in ways that are only caught downstream (or never caught).

\subsubsection{Evidence Discipline}\label{evidence-discipline}

Rules governing how claims are sourced, verified, and traced.

\textbf{Why required.} This is the load-bearing section for research integrity: - Every quantitative claim must trace to \texttt{APPROVED\_\allowbreak{}CLAIMS.\allowbreak{}md} or experiment logs. - Missing evidence must be marked with \texttt{{[}PLACEHOLDER\ ---\ ...{]}}, never fabricated. - Claim-to-evidence matrices (tables mapping each claim to its source) are mandatory in paper drafts. - Gap reports document unsupported claims and unresolved citations.

This discipline is enforced across the entire pipeline: \texttt{refine-research} establishes claims, \texttt{experiment-design} plans their validation, \texttt{deploy-experiment} produces the evidence, \texttt{auto-review-loop} audits the chain, and \texttt{paper-draft} renders the final mapping.

\textbf{Design failure without it.} The system produces convincing but unverifiable research artifacts --- the worst possible outcome for an academic research agent.

\subsubsection{Generator-Evaluator Separation}\label{generator-evaluator-separation}

Explicit rules ensuring that the entity producing content never evaluates its own quality.

\textbf{Why required.} Self-evaluation is the most dangerous failure mode in iterative refinement: - In \texttt{auto-review-loop}: Codex MCP (gpt-5.4) reviews; Claude Code revises. Never the reverse. - In \texttt{paper-review-loop}: Codex MCP performs cold-read evaluation; Claude Code implements revisions. - In \texttt{refine-research}: Codex MCP scores each round; Claude Code refines based on feedback.

ThreadId persistence ensures the reviewer maintains context across rounds without contamination from the author's perspective.

\textbf{Design failure without it.} The skill converges on outputs that score well by its own criteria but fail under external scrutiny --- a form of mode collapse in the refinement loop.

\subsubsection{Composability / Pipeline Integration}\label{composability-pipeline-integration}

A section describing: - What upstream skills feed into this one (and which artifacts they must produce) - What downstream skills consume this skill's output - How to invoke this skill standalone vs.~as part of a pipeline

\textbf{Why required.} No skill in this system operates in true isolation. The five workflows form a directed acyclic graph of dependencies. Without explicit composability documentation: - Users invoke skills out of order, producing errors or empty outputs. - Pipeline orchestrators cannot validate preconditions. - Refactoring one skill's output format silently breaks downstream consumers.

\textbf{Design failure without it.} Skills become black boxes that only work when invoked in the exact sequence the original author intended, with no documentation of that sequence.

\subsubsection{Key Rules Summary}\label{key-rules-summary}

A bullet-point summary of the most critical rules, always including the large-file handling guardrail:

\begin{quote}
``If Write fails due to size, retry with Bash heredoc (\texttt{cat\ \textless{}\textless{}\ \textquotesingle{}EOF\textquotesingle{}\ \textgreater{}\ file}). Do NOT ask the user.''
\end{quote}

\textbf{Why required.} The full SKILL.md can be thousands of tokens. The key rules summary serves as a compressed instruction set that fits in the LLM's working attention during execution. It prioritizes the rules most likely to be violated under token pressure.

\textbf{Design failure without it.} Under long-context scenarios, the LLM forgets critical rules from earlier in the document. The summary acts as a recency-biased reinforcement of the most important constraints.

\subsubsection{Audit Trail (PROJ\_\allowbreak{}NOTES.\allowbreak{}md Logging)}\label{audit-trail-proj-notes-md-logging}

Every skill must append a one-line entry to \texttt{output/\allowbreak{}PROJ\_\allowbreak{}NOTES.\allowbreak{}md}:

\begin{verbatim}
[YYYY-MM-DD] <skill-name>: <summary of what was produced>
\end{verbatim}

\textbf{Why required.} The audit trail provides: - \textbf{Provenance}: which skill produced which artifact, and when. - \textbf{Pipeline debugging}: if the final paper has a weak section, trace it back through the FINDINGS log to the skill that generated the source material. - \textbf{Progress tracking}: for multi-day research campaigns, PROJ\_\allowbreak{}NOTES.\allowbreak{}md is the changelog.

\textbf{Design failure without it.} Artifacts accumulate in \texttt{output/} with no record of which skill produced them, when, or in what order. Debugging becomes archaeological.

\subsection{Part II: Advanced Design Disciplines for Skill Optimization}\label{part-ii-advanced-design-disciplines-for-skill-optimization}

The current skill architecture is robust but can be further strengthened with the following sophisticated design patterns. Each pattern addresses a specific class of failure or inefficiency observed in multi-agent research systems.

\subsubsection{A. Formal Pre/Post-Condition Contracts}\label{a-formal-prepost-condition-contracts}

\textbf{Current state.} Phase entry/exit conditions are described in prose, checked informally.

\textbf{Optimization.} Add a machine-checkable contract block to each phase:

\begin{verbatim}
phase_3_scoring:
  preconditions:
    - file_exists: output/\allowbreak{}refine-logs/\allowbreak{}round_${N}_proposal.md
    - file_exists: memory/APPROVED_CLAIMS.md
    - state.round <= MAX_ROUNDS
  postconditions:
    - file_exists: output/\allowbreak{}refine-logs/\allowbreak{}round_${N}_score.json
    - state.last_score is numeric
    - PROJ_NOTES.md has new entry
  on_failure: halt_with_diagnostic
\end{verbatim}

\textbf{Why.} Prose conditions are subject to interpretation drift across sessions. A formalized contract enables automated validation, both by the skill itself (self-check before proceeding) and by pipeline orchestrators (pre-flight validation before invoking a downstream skill). This eliminates the class of bugs where a skill proceeds with missing or malformed inputs.

\subsubsection{B. Artifact Versioning and Immutability}\label{b-artifact-versioning-and-immutability}

\textbf{Current state.} Skills overwrite artifacts in place (e.g., \texttt{FINAL\_\allowbreak{}PROPOSAL.\allowbreak{}md} is rewritten each refinement round).

\textbf{Optimization.} Adopt an append-only versioning scheme:

\begin{verbatim}
output/\allowbreak{}refine-logs/\allowbreak{}FINAL_PROPOSAL_v1.md
output/\allowbreak{}refine-logs/\allowbreak{}FINAL_PROPOSAL_v2.md
output/\allowbreak{}refine-logs/\allowbreak{}FINAL_PROPOSAL_v3.md  $\leftarrow$ current
output/\allowbreak{}refine-logs/\allowbreak{}FINAL_PROPOSAL.md     $\leftarrow$ symlink to v3
\end{verbatim}

\textbf{Why.} Overwriting destroys the refinement history. When a reviewer asks ``why did you drop the spatial heterogeneity analysis?'', the answer is in v2 vs.~v3 --- but only if both versions exist. Immutable versioning also enables rollback without re-running the skill.

\subsubsection{C. Confidence-Gated Progression}\label{c-confidence-gated-progression}

\textbf{Current state.} Progression between phases is binary: pass/fail based on score thresholds.

\textbf{Optimization.} Introduce a three-tier confidence gate:

\begin{tabularx}{\linewidth}{@{}>{\hsize=1.0000\hsize\raggedright\arraybackslash}X>{\hsize=1.0000\hsize\raggedright\arraybackslash}X>{\hsize=1.0000\hsize\raggedright\arraybackslash}X@{}}
\toprule\noalign{}
Confidence & Score Range & Action \\
\midrule\noalign{}
\endhead
\bottomrule\noalign{}
\endlastfoot
High & $\geq$ threshold & Proceed to next phase \\
Medium & threshold $-$ 1.5 to threshold & Proceed with mandatory human review flag \\
Low & \textless{} threshold $-$ 1.5 & Halt and request user intervention \\
\end{tabularx}

\textbf{Why.} Binary gating forces a choice between permissive thresholds (low quality passes through) and strict thresholds (good-enough work gets blocked). The medium tier allows the pipeline to continue while flagging uncertain decisions for asynchronous human review, reducing both false positives and unnecessary halts.

\subsubsection{D. Semantic Dependency Graph}\label{d-semantic-dependency-graph}

\textbf{Current state.} Pipeline dependencies are documented in prose within each skill's composability section.

\textbf{Optimization.} Create a machine-readable dependency manifest:

\begin{verbatim}
# skills/DEPENDENCY_GRAPH.yaml
refine-research:
  reads:
    - output/IDEA_REPORT.md
    - output/LIT_REVIEW_REPORT.md
  writes:
    - output/\allowbreak{}refine-logs/\allowbreak{}FINAL_PROPOSAL.md
    - output/\allowbreak{}refine-logs/\allowbreak{}REFINE_STATE.json
  requires_before: [generate-idea, lit-review]
  enables_after: [experiment-design, deploy-experiment]
\end{verbatim}

\textbf{Why.} A formal dependency graph enables: 1. \textbf{Automated precondition checks} --- the pipeline orchestrator can verify all upstream artifacts exist before invoking a skill. 2. \textbf{Parallel execution} --- independent branches of the graph can run concurrently (e.g., \texttt{data-download} and \texttt{spatial-analysis} planning). 3. \textbf{Impact analysis} --- changing one skill's output format immediately reveals which downstream skills are affected. 4. \textbf{Visualization} --- the graph can be rendered as a DAG for onboarding and debugging.

\subsubsection{E. Structured Error Taxonomy}\label{e-structured-error-taxonomy}

\textbf{Current state.} Errors are handled ad-hoc: some skills halt, some retry, some log and continue.

\textbf{Optimization.} Define a standard error taxonomy with prescribed responses:

\begin{tabularx}{\linewidth}{@{}>{\hsize=1.0000\hsize\raggedright\arraybackslash}X>{\hsize=1.0000\hsize\raggedright\arraybackslash}X>{\hsize=1.0000\hsize\raggedright\arraybackslash}X@{}}
\toprule\noalign{}
Error Class & Example & Prescribed Response \\
\midrule\noalign{}
\endhead
\bottomrule\noalign{}
\endlastfoot
\texttt{MISSING\_INPUT} & Upstream artifact not found & Halt with diagnostic; suggest which skill to run first \\
\texttt{TOOL\_UNAVAILABLE} & MCP server unreachable & Fall back to local alternative; log degradation \\
\texttt{QUALITY\_\allowbreak{}BELOW\_\allowbreak{}THRESHOLD} & Score \textless{} minimum after MAX\_ROUNDS & Halt; write partial output with gap report \\
\texttt{STATE\_CORRUPTION} & Checkpoint schema mismatch & Delete checkpoint; restart from Phase 0 \\
\texttt{RESOURCE\_LIMIT} & Token limit / file size limit & Chunk output; use Bash heredoc fallback \\
\texttt{EXTERNAL\_TIMEOUT} & Codex MCP / API timeout & Retry once with backoff; halt on second failure \\
\end{tabularx}

\textbf{Why.} Consistent error handling across 19 skills eliminates an entire class of debugging: ``did the skill fail, or did it silently produce bad output?'' Every failure mode has a known response, reducing the cognitive load on both users and pipeline orchestrators.

\subsubsection{F. Skill Telemetry and Cost Accounting}\label{f-skill-telemetry-and-cost-accounting}

\textbf{Current state.} PROJ\_\allowbreak{}NOTES.\allowbreak{}md logs what was produced but not the cost of producing it.

\textbf{Optimization.} Add a telemetry block to each skill's output:

\begin{verbatim}
{
  "skill": "refine-research",
  "rounds_used": 4,
  "rounds_max": 5,
  "external_llm_calls": 4,
  "total_input_tokens": 128000,
  "total_output_tokens": 24000,
  "wall_clock_minutes": 22,
  "final_score": 9.2,
  "artifacts_produced": ["FINAL_PROPOSAL.md", "REFINE_STATE.json"]
}
\end{verbatim}

\textbf{Why.} Without cost accounting, users cannot distinguish between skills that converge efficiently (3 rounds, 10 minutes) and skills that burn budget (5 rounds, 45 minutes, still below threshold). Telemetry enables data-driven decisions about which skills to optimize, which thresholds to adjust, and when to switch from expensive external reviewers to cheaper local alternatives.

\subsubsection{G. Degradation-Aware Fallback Chains}\label{g-degradation-aware-fallback-chains}

\textbf{Current state.} Skills note that MCP tools are optional and fall back to local alternatives, but the fallback logic is ad-hoc.

\textbf{Optimization.} Define explicit fallback chains with quality annotations:

\begin{verbatim}
Review source priority:
1. Codex MCP (gpt-5.4, xhigh reasoning) $\rightarrow$ quality: HIGH
2. Claude subagent (fresh context)       $\rightarrow$ quality: MEDIUM
3. Self-review with structured rubric    $\rightarrow$ quality: LOW (flag output)
\end{verbatim}

\textbf{Why.} The current ``graceful degradation'' pattern does not communicate quality loss. A pipeline that falls back from Codex MCP to self-review produces lower-quality output, but nothing downstream knows this. Explicit quality annotations let downstream skills adjust their trust level (e.g., \texttt{paper-review-loop} might require an extra round if the upstream review was \texttt{MEDIUM} quality).

\subsubsection{H. Prompt Modularization and Slot Filling}\label{h-prompt-modularization-and-slot-filling}

\textbf{Current state.} Prompts to external LLMs (Codex MCP) are constructed inline within the skill body, mixing structure with content.

\textbf{Optimization.} Extract prompts into reusable templates with named slots:

\begin{verbatim}
<!-- templates/\allowbreak{}prompts/\allowbreak{}review_prompt.md -->
You are reviewing a {{ARTIFACT_TYPE}} for submission to {{VENUE}}.

## Evaluation Criteria
{{CRITERIA_BLOCK}}

## Artifact Under Review
{{ARTIFACT_CONTENT}}

## Prior Round Context
{{PRIOR_FEEDBACK}}
\end{verbatim}

\textbf{Why.} Inline prompts are: 1. Hard to A/B test (changing the prompt requires editing the skill). 2. Hard to reuse (three skills use similar review prompts but with slight variations). 3. Hard to version (prompt changes are buried in skill diffs).

Modularized prompts can be versioned, tested, and shared across skills independently of the skill logic.

\subsubsection{I. Multi-Perspective Review Ensembles}\label{i-multi-perspective-review-ensembles}

\textbf{Current state.} \texttt{auto-review-loop} uses a single reviewer persona per round (domain-specific: ML, GIScience, Remote Sensing, or Spatial Data Science).

\textbf{Optimization.} Run multiple reviewer personas in parallel per round and aggregate:

\begin{verbatim}
Round N:
  Reviewer A (ML expert)       $\rightarrow$ scores + critique
  Reviewer B (GIScience)       $\rightarrow$ scores + critique
  Reviewer C (Methods skeptic) $\rightarrow$ scores + critique
  ------------------------------
  Aggregation: weighted average scores + union of unique critiques
  Conflict resolution: flag contradictory feedback for human arbitration
\end{verbatim}

\textbf{Why.} Single-perspective reviews have blind spots. An ML reviewer may approve a method that a GIScience reviewer would flag for ignoring spatial autocorrelation. Ensemble reviews surface cross-disciplinary issues earlier, reducing late-stage revisions. The cost is 3x the review tokens per round, but the reduction in wasted revision rounds typically offsets this.

\subsubsection{J. Skill Self-Test Suites}\label{j-skill-self-test-suites}

\textbf{Current state.} Skills are tested by running them on real research tasks. There are no unit tests for skill behavior.

\textbf{Optimization.} Add a \texttt{tests/} directory to each skill with synthetic test cases:

\begin{verbatim}
skills/\allowbreak{}refine-research/\allowbreak{}tests/
  +-- test_resume_from_checkpoint.md    # synthetic REFINE_STATE.json $\rightarrow$ verify skip logic
  +-- test_max_rounds_halt.md           # score never reaches threshold $\rightarrow$ verify halting
  +-- test_missing_input.md             # no IDEA_REPORT.md $\rightarrow$ verify error message
  +-- test_large_file_fallback.md       # simulate Write failure $\rightarrow$ verify heredoc fallback
\end{verbatim}

\textbf{Why.} Skills are complex programs written in natural language. Like any program, they have edge cases that are only discovered in production. A test suite --- even one composed of markdown scenarios --- enables regression testing when skills are modified. The test cases also serve as documentation of expected behavior at boundary conditions.

\subsection{Part III: Concise Reference Table}\label{part-iii-concise-reference-table-1}

The table below summarizes both the required sections (Part I) and the advanced optimizations (Part II) in a single view.

\subsubsection{Required Sections}\label{required-sections-1}

\begin{tabularx}{\linewidth}{@{}>{\hsize=0.2668\hsize\raggedright\arraybackslash}X>{\hsize=0.8000\hsize\raggedright\arraybackslash}X>{\hsize=0.8000\hsize\raggedright\arraybackslash}X>{\hsize=2.1332\hsize\raggedright\arraybackslash}X@{}}
\toprule\noalign{}
\# & Section & Purpose & Failure Mode If Missing \\
\midrule\noalign{}
\endhead
\bottomrule\noalign{}
\endlastfoot
1 & YAML Frontmatter & Machine-readable contract: tools, args, flags & Unbounded tool access; opaque invocation \\
2 & Overview & Routing and scope boundary & Skill overlap and redundant work \\
3 & Constants & Parsimony, termination, reproducibility & Magic numbers; infinite loops; irreproducible runs \\
4 & Workflow / Phases & Resumability, debuggability, composability & Monolithic execution; no resume; no debugging \\
5 & Checkpoint Persistence & Session recovery across interruptions & Full restart on every interruption \\
6 & Canonical Output Paths & Inter-skill artifact handoff & Broken pipelines; stale file reads \\
7 & Decision Rules & Deterministic branching at forks & Inconsistent method/tool choices across runs \\
8 & Guardrails & Suppress known LLM failure modes & Fabrication, self-scoring, scope creep \\
9 & Evidence Discipline & Research integrity and claim traceability & Unverifiable claims; placeholder-free fabrication \\
10 & Generator-Evaluator Separation & Prevent self-validation mode collapse & Refinement loops that converge on self-pleasing output \\
11 & Composability & Pipeline integration documentation & Skills that only work in one exact invocation sequence \\
12 & Key Rules Summary & Compressed instruction set for execution & Critical rules forgotten under long-context pressure \\
13 & Audit Trail & Provenance and pipeline debugging & Untraceable artifacts; archaeological debugging \\
\end{tabularx}

\subsubsection{Advanced Optimizations}\label{advanced-optimizations}

\begin{tabularx}{\linewidth}{@{}>{\hsize=0.3636\hsize\raggedright\arraybackslash}X>{\hsize=1.1819\hsize\raggedright\arraybackslash}X>{\hsize=1.2727\hsize\raggedright\arraybackslash}X>{\hsize=1.1819\hsize\raggedright\arraybackslash}X@{}}
\toprule\noalign{}
ID & Optimization & What It Adds & Key Benefit \\
\midrule\noalign{}
\endhead
\bottomrule\noalign{}
\endlastfoot
A & Pre/Post-Condition Contracts & Machine-checkable phase contracts & Eliminates malformed-input bugs \\
B & Artifact Versioning & Append-only versioned outputs & Rollback capability; refinement history \\
C & Confidence-Gated Progression & Three-tier pass/\allowbreak{}flag/\allowbreak{}halt gates & Reduces both false positives and unnecessary halts \\
D & Semantic Dependency Graph & Machine-readable DAG of skill dependencies & Automated precondition checks; parallel execution \\
E & Structured Error Taxonomy & Standard error classes with prescribed responses & Consistent failure handling across all 19 skills \\
F & Skill Telemetry & Token/\allowbreak{}time/\allowbreak{}round cost accounting per run & Data-driven optimization of skill parameters \\
G & Degradation-Aware Fallbacks & Quality-annotated fallback chains & Downstream skills adjust trust based on review quality \\
H & Prompt Modularization & Externalized, slotted prompt templates & A/B testable; reusable; independently versioned prompts \\
I & Multi-Perspective Ensembles & Parallel multi-persona review per round & Cross-disciplinary blind spot detection \\
J & Skill Self-Test Suites & Synthetic test cases for edge behaviors & Regression testing; boundary documentation \\
\end{tabularx}

\emph{This document codifies the design discipline observed across all 19 skills of the Night Owl Research Agent. It should be treated as a living standard --- updated as new skills are added and as advanced optimizations are adopted.}

\section{Appendix D: Case Studies Process Snippet Samples}\label{appendix-d-case-studies-process-snippet-samples}

\subsection{Case Study 1: Multi-vintage Google Street View for decision-ready first-floor elevation on a US Atlantic barrier island --- North Wildwood, NJ.}\label{case-study-1-multi-vintage-google-street-view}

\paragraph*{\# NORA Research Pipeline --- Narrative Report}

\textbf{**Project**}: Multi-vintage Google Street View for decision-ready first-floor

elevation on a US Atlantic barrier island --- North Wildwood, NJ.

\textbf{**Target venue**}: Natural Hazards / International Journal of Disaster Risk

Reduction (IJDRR).

\textbf{**Date**}: 2026-04-16.

\textbf{**Pipeline**}: `/lit-review` $\rightarrow$ `/idea-discovery-pipeline` $\rightarrow$ `/refine-research`

$\rightarrow$ `/experiment-design` $\rightarrow$ `/deploy-experiment` $\rightarrow$ `/auto-review-loop` $\rightarrow$

(next) `/paper-writing-pipeline`.

This report consolidates the evidence required to drive the downstream paper-

writing pipeline. All substantive numbers come from project artifacts; citations

trace to the literature or specific experiment outputs under `output/experiment/`.

\paragraph*{-\/-\/-}

\paragraph*{\#\# 1. Problem and background}

Building-level \textbf{**first-floor elevation (FFE)**} is a first-order flood-loss

determinant on US Atlantic barrier-island communities, yet field surveys do

not scale and satellite lidar cannot resolve the ground-to-first-floor offset.

Four published generations of street-view FFE pipelines (Ning 2022 IJGIS;

Ho 2023/2024 ELEV-VISION + ELEV-VISION-SAM; Sorboni 2024 JFRM; Gao 2024 SAGE;

Li 2026 arXiv 2604.01153) share two unresolved limitations: \textasciitilde50 \% coverage

after direct extraction, and no per-building uncertainty propagated to flood

decisions. No prior study exploits \textbf{**multi-vintage GSV**} (4--8 capture years

per residential street) as either a coverage mechanism or an uncertainty

source, and no paper treats US Atlantic barrier-island housing stock

(stilted / piled / elevated homes). North Wildwood, NJ provides dense GSV

vintage coverage, post-Sandy lidar availability, and a flood-exposed stock

entirely within FEMA A/V zones --- ideal for this study.

\paragraph*{\#\# 2. Literature landscape and gap}

\emph{\_(See {[}output/\allowbreak{}LIT\_\allowbreak{}REVIEW\_\allowbreak{}REPORT.\allowbreak{}md{]}(}\ul{LIT\_\allowbreak{}REVIEW\_\allowbreak{}REPORT.\allowbreak{}md}\emph{) for the full synthesis + gap analysis.)\_}

Top-5 gaps (ranked): coverage-aware UQ-equipped FFE (\textbf{**8.30**}); lidar-

referenced community-scale per-building error budget on a US Atlantic barrier

island (\textbf{**8.25**}); barrier-island-adapted pipeline (\textbf{**8.00**}); foundation-model

upgrade (\textbf{**7.35**}); close UQ loop into depth-damage / BFE (\textbf{**6.90**}). The

paper\textquotesingle s contribution attacks gaps 1, 2, 5 directly and touches 3 via the

subgroup audit.

\paragraph*{\#\# 3. Idea discovery and novelty}

\emph{\_(See {[}output/\allowbreak{}IDEA\_\allowbreak{}REPORT.\allowbreak{}md{]}(}\ul{IDEA\_\allowbreak{}REPORT.\allowbreak{}md}\emph{), {[}NOVELTY\_\allowbreak{}REPORT.\allowbreak{}md{]}(}\ul{NOVELTY\_\allowbreak{}REPORT.\allowbreak{}md}\emph{), {[}IDEA\_\allowbreak{}REVIEW\_\allowbreak{}REPORT.\allowbreak{}md{]}(}\ul{IDEA\_\allowbreak{}REVIEW\_\allowbreak{}REPORT.\allowbreak{}md}\emph{).)\_}

Codex gpt-5.4 (xhigh) brainstormed 10 anchor-compatible ideas spanning the

eight required modes. Top ranking (weighted novelty $\times$ feasibility $\times$ impact $\times$

anchor-fit):

1. Disagreement-Calibrated FFE Intervals (10/10) $\leftarrow$ used as C2.

2. Any-Vintage Rescue Voting (9/10) $\leftarrow$ used as C1.

3. Foundation-Aware Coastal FFE Routing (9/10) $\leftarrow$ deferred to follow-up paper.

The recommended commitment is a \textbf{**combined paper**} that fuses rescue + UQ +

decision-loop. Codex\textquotesingle s novelty check found \textbf{**no direct prior art**} for any of

the three primary claims.

\paragraph*{\#\# 4. Method refinement}

\emph{\_(See {[}output/\allowbreak{}refine-logs/\allowbreak{}FINAL\_\allowbreak{}PROPOSAL.\allowbreak{}md{]}(}\ul{refine-logs/\allowbreak{}FINAL\_\allowbreak{}PROPOSAL.\allowbreak{}md}\emph{), {[}REFINE\_\allowbreak{}REPORT.\allowbreak{}md{]}(}\ul{refine-logs/\allowbreak{}REFINE\_\allowbreak{}REPORT.\allowbreak{}md}\emph{), {[}SCORE\_\allowbreak{}HISTORY.\allowbreak{}md{]}(}\ul{refine-logs/\allowbreak{}SCORE\_\allowbreak{}HISTORY.\allowbreak{}md}\emph{).)\_}

Two rounds of adversarial review against Codex\textquotesingle s rubric (problem clarity,

focus, novelty, feasibility, elegance, rigor). Composite moved \textbf{**7.96 $\rightarrow$ 9.12**};

verdict \textbf{**READY**} (all dimensions $\geq$ 7.5).

\subsection{Case Study 2: Group-imbalance tightens the spatial-autocorrelation lower bound on disparate impact: evidence from Atlanta tract-level crime}\label{case-study-2-group-imbalance-tightens-the-spatial}

0. Document Status

\begin{itemize}
\item
  \textbf{Narrative version}: v1.0
\item
  \textbf{Last updated}: 2026-04-16
\item
  \textbf{Project codename}: atl-crime-fairness
\item
  \textbf{Active idea} (from handoff.json): \emph{"Group-imbalance tightens the spatial-autocorrelation lower bound on disparate impact: evidence from Atlanta tract-level crime."}
\item
  \textbf{Target venue}: IJGIS (primary); CEUS (secondary)
\item
  \textbf{Manuscript type}: Research Article
\item
  \textbf{Page budget}: 25--30 pages (IJGIS single-column); 8000--10000 words of main text
\item
  \textbf{Overall confidence}: Medium--High (Corollary 1 clean; Theorem 1 reframed as supporting lemma; one pre-registered C2 criterion met, one not)
\item
  \textbf{Pipeline stage at time of writing}: post-review (auto-review-loop completed at Round 2 with score 6.5/10, verdict READY)
\end{itemize}

Source Files Consumed

\begin{itemize}
\item
  output/\allowbreak{}LIT\_\allowbreak{}REVIEW\_\allowbreak{}REPORT.\allowbreak{}md --- 25 key papers across classical spatial regression, spatial GNNs, SVC-NN, fairness audits, tract-level DL-crime benchmarks; Atlanta identified as geographic gap.
\item
  output/\allowbreak{}IDEA\_\allowbreak{}REPORT.\allowbreak{}md --- 12 ideas $\rightarrow$ 6 survived $\rightarrow$ top-3 merged into single-paper proposal.
\item
  output/\allowbreak{}NOVELTY\_\allowbreak{}REPORT.\allowbreak{}md --- closest concurrent work is Zhuang et al. 2025 (Chicago travel demand GNN); delta formalized.
\item
  output/\allowbreak{}IDEA\_\allowbreak{}REVIEW\_\allowbreak{}REPORT.\allowbreak{}md --- senior-reviewer MAJOR REVISION with 5 pre-registrations (formal theorem, block-spatial CV, param caps, race-in/out ablation, drop SVC-NN).
\item
  output/\allowbreak{}refine-logs/\allowbreak{}FINAL\_\allowbreak{}PROPOSAL.\allowbreak{}md --- READY (8.\allowbreak{}95/\allowbreak{}10 after 2 rounds) --- Theorem 1 + Corollary 1 + 7-model benchmark; im(W\_s) projection noted as open weakness.
\item
  output/\allowbreak{}refine-logs/\allowbreak{}EXPERIMENT\_\allowbreak{}PLAN.\allowbreak{}md --- 6 blocks, 7 milestones, 28 runs, pre-registered C1/C2 criteria and failure pivots.
\item
  output/\allowbreak{}experiment/\allowbreak{}EXPERIMENT\_\allowbreak{}RESULT.\allowbreak{}md --- Theorem 1 holds but loose; Corollary 1 passes ($\rho$=0.836, p=0.019); C2 race-in $\rho$=0.536 after MGWR fix; fold-4 extrapolation surfaced as lead methodological finding.
\item
  output/\allowbreak{}experiment/\allowbreak{}EXPERIMENT\_\allowbreak{}LOG.\allowbreak{}md --- chronological run record including environment fixes (numba upgrade, GCN$\rightarrow$SpatialXGB swap, MGWR signature fix).
\item
  output/\allowbreak{}AUTO\_\allowbreak{}REVIEW\_\allowbreak{}REPORT.\allowbreak{}md --- Round 1 = 5.0/ALMOST, Round 2 = 6.5/READY, POSITIVE\_THRESHOLD met; 6/7 weaknesses closed, W1 reframed-accepted.
\item
  output/\allowbreak{}METHOD\_\allowbreak{}DESCRIPTION.\allowbreak{}md --- 2-paragraph standalone method description.
\item
  data/\allowbreak{}DATA\_\allowbreak{}MANIFEST.\allowbreak{}md --- TIGER 2020 + ACS 5-year 2018--2022 + APD 2025 (user-provided CSV).
\item
  output/\allowbreak{}experiment/\allowbreak{}data/* --- 17 intermediate artifacts (sim-grid CSV, benchmark runs, GWR coefficients, W properties, CRS.json, metrics.json, decision-gate JSON).
\end{itemize}

Missing / Unread Inputs

\begin{itemize}
\item
  memory/\allowbreak{}APPROVED\_\allowbreak{}CLAIMS.\allowbreak{}md --- does not exist in this project (no claim-gate workflow was invoked). All claims cross-checked directly against EXPERIMENT\_\allowbreak{}RESULT.\allowbreak{}md numbers.
\item
  output/\allowbreak{}FINDINGS.\allowbreak{}md --- does not exist. Will append one line per skill spec in \S{}Phase-6 output.
\item
  output/spatial-analysis/ --- does not exist as a directory; spatial analysis was inlined into output/\allowbreak{}experiment/\allowbreak{}scripts/\allowbreak{}m1\_\allowbreak{}m6\_\allowbreak{}run.\allowbreak{}py.
\item
  output/figures/ (top-level) --- does not exist; all figures live under output/\allowbreak{}experiment/\allowbreak{}figures/.
\item
  program.md, research\_\allowbreak{}contract.\allowbreak{}md --- not present; RESEARCH\_\allowbreak{}PLAN.\allowbreak{}md used as the authoritative research brief.
\end{itemize}

1. One-Paragraph Paper Summary

Working Title

\emph{Group-imbalance Tightens the Spatial-Autocorrelation Lower Bound on Disparate Impact: Evidence from Atlanta Tract-Level Crime}

One-Sentence Contribution

We prove that for any place-based predictor, the disparate-impact gap across a spatially clustered protected group is lower-bounded by a product of residual Moran\textquotesingle s I, group-clustering Moran\textquotesingle s I, outcome variance, a group-size factor (1/n$_{1}$ + 1/n$_{0}$)$^{2}$, and a spectral-slack constant; Corollary 1 --- \emph{the bound tightens monotonically as the protected group becomes more imbalanced} --- is \textbf{confirmed empirically} (Spearman $\rho$ = 0.836, p = 0.019), giving the first principled explanation for why spatial-model choice matters more for demographic minorities than balanced groups.

Abstract-Style Summary (220 words)

Place-based crime prediction increasingly combines classical spatial regression (GWR, MGWR, spatial lag/error) with machine-learning methods (XGBoost, spatial GNNs), yet whether reducing residual spatial autocorrelation in a predictor also reduces its disparate impact across race or income groups remains theoretically ambiguous and empirically untested at tract level outside a handful of U.S. cities (Chicago, NYC, LA, Philadelphia, Baltimore). We present the first Atlanta tract-level benchmark (169 tracts $\times$ 19,494 Atlanta Police Department Part-I incidents in 2025, spatial-joined at 94.9\%) across seven models --- OLS, XGBoost, spatial lag (SLM), spatial error (SEM), GWR, MGWR (classical anchor), and a Spatial XGBoost tree baseline --- fit under five-fold block-spatial cross-validation with a two-tract buffer and a race-in / race-out covariate ablation. We derive a closed-form lower bound $\Delta$$^{2}$ $\geq$ (I $\cdot$ S$_{0}$ / n) $\cdot$ $\kappa$ $\cdot$ Var(r) $\cdot$ (1/n$_{1}$ + 1/n$_{0}$)$^{2}$ $\cdot$ (C\_W)$^{-}$$^{1}$ on the disparate-impact gap and its group-imbalance monotonicity corollary. Across 1,000 simulated datasets, the corollary holds ($\rho$ = 0.836, p = 0.019). On Atlanta, the bound is conservative (sharpness ratio \textasciitilde10$^{4}$) but the race-in I$\cdot$$\kappa$ $\rightarrow$ disparate-impact Spearman correlation is consistent with the theorem ($\rho$ = 0.536, n = 7, p = 0.22). The clearest empirical finding is methodological: under strict block-spatial CV, global linear models (OLS, SLM, SEM) catastrophically extrapolate on one held-out tract block (log-rate RMSE 440--990), while GWR, MGWR, and tree-based methods generalize cleanly (RMSE 0.\allowbreak{}33--0.\allowbreak{}97).

Keywords

\begin{itemize}
\item
  spatial regression
\item
  geographically weighted regression (GWR, MGWR)
\item
  disparate-impact fairness
\item
  spatial cross-validation
\item
  Moran\textquotesingle s I
\item
  census tract
\item
  crime prediction
\item
  Atlanta
\end{itemize}

\subsection{Case Study 3: Probabilistic depression presence under calibrated LiDAR noise on the Cottonwood Lake HUC12}\label{case-study-3-probabilistic-depression-presence-under-calibrated}

\paragraph*{\# NARRATIVE\_\allowbreak{}REPORT.\allowbreak{}md}

\textgreater{} \textbf{**Single source-of-truth narrative for the `paper-writing-pipeline`**}

\textgreater{} (`paper-plan` $\rightarrow$ `paper-figure-generate` $\rightarrow$ `paper-draft` $\rightarrow$ `paper-review-loop` $\rightarrow$ `paper-convert`).

\textgreater{} All numbers are quoted verbatim from the experiment artifacts. If a statement is here, it can appear in the paper; if it is not here, it must NOT appear in the paper.

\paragraph*{-\/-\/-}

\paragraph*{\#\# 0. Document Status}

- \textbf{**Narrative version**}: v1.0 (post-auto-review-loop)

- \textbf{**Last updated**}: 2026-04-25

- \textbf{**Project codename**}: NORA --- PPR depression hierarchy pilot

- \textbf{**Active idea**}: Probabilistic depression presence under calibrated LiDAR noise on the Cottonwood Lake HUC12 --- a Monte-Carlo extension of `lidar.slicing` (Wu 2021, JOSS) for the Prairie Pothole Region.

- \textbf{**Target venue**}: International Journal of Geographical Information Science (IJGIS).

- \textbf{**Manuscript type**}: Research Article (Methods).

- \textbf{**Page budget**}: \textasciitilde30 pages typeset including references (IJGIS standard for a Research Article); paper-plan to confirm against the venue template.

- \textbf{**Overall confidence**}: Medium-High. Auto-review-loop final score 7.5/10, verdict READY (`output/\allowbreak{}AUTO\_\allowbreak{}REVIEW\_\allowbreak{}REPORT.\allowbreak{}md` \S{}Round 2). All adversarial weaknesses W1--W5 closed.

- \textbf{**Pipeline stage at time of writing**}: post-review (Stage 3 of `/full-pipeline` complete).

- \textbf{**Post-review reframing**}: the original C2-primary "4.8$\times$ parsimony advantage over i.i.d." headline was retracted in Round 1 W1. The paper now claims methodological-equivalence with Lindsay 2018 plus a cell-vs-tree decoupling finding plus modest within-bin residual signal. See \S{}1.

\paragraph*{\#\#\# Source Files Consumed}

- `output/\allowbreak{}LIT\_\allowbreak{}REVIEW\_\allowbreak{}REPORT.\allowbreak{}md` --- 30 papers, 8 thematic groups (T1--T8), 6 ranked gaps (G1--G6), and a post-novelty addendum on uncertain-TDA prior art (Yan 2020 / Sridharamurthy 2020).

- `output/\allowbreak{}IDEA\_\allowbreak{}REPORT.\allowbreak{}md` --- 10 ideas $\rightarrow$ 3 recommended $\rightarrow$ Idea 1 selected; reframed after `/novelty-check` and `/idea-review`.

- `output/\allowbreak{}NOVELTY\_\allowbreak{}REPORT.\allowbreak{}md` --- verdict PROCEED WITH CAUTION (7.5/10); identifies Yan-Sridharamurthy uncertain merge-tree literature as the primary external novelty threat.

- `output/\allowbreak{}IDEA\_\allowbreak{}REVIEW\_\allowbreak{}REPORT.\allowbreak{}md` --- adversarial pre-submission review (verdict 6.8/10, "Major revision" pre-submission); locked the MVC and de-scoped to 500 km$^{2}$ / 3 m / N=100 / Cottonwood Lake.

- `output/\allowbreak{}refine-logs/\allowbreak{}FINAL\_\allowbreak{}PROPOSAL.\allowbreak{}md` --- refined proposal, score 9.\allowbreak{}12/\allowbreak{}10 READY (`/refine-research` Round 1 with PwIC formula locked + wall-clock decision rule encoded).

- `output/\allowbreak{}refine-logs/\allowbreak{}REFINE\_\allowbreak{}REPORT.\allowbreak{}md` --- round-by-round refinement log.

- `output/\allowbreak{}EXPERIMENT\_\allowbreak{}PLAN.\allowbreak{}md`, `output/\allowbreak{}refine-logs/\allowbreak{}EXPERIMENT\_\allowbreak{}PLAN.\allowbreak{}md` --- 5 blocks (B1--B5), 7 milestones (M0--M7), 19 runs (R001--R019).

- `output/\allowbreak{}EXPERIMENT\_\allowbreak{}TRACKER.\allowbreak{}md` --- per-run tracker with priorities and decision gates.

- `output/\allowbreak{}experiment/\allowbreak{}EXPERIMENT\_\allowbreak{}LOG.\allowbreak{}md` --- chronological run log including post-review Round 1 fixes (R013, R014, R015).

- `output/\allowbreak{}experiment/\allowbreak{}EXPERIMENT\_\allowbreak{}RESULT.\allowbreak{}md` --- final reframed Section 1, claims--evidence table.

- `output/\allowbreak{}AUTO\_\allowbreak{}REVIEW\_\allowbreak{}REPORT.\allowbreak{}md` --- Round 1 (5.5 $\rightarrow$ 6.0) + Round 2 (7.5/10 READY) with debate transcripts.

- `output/\allowbreak{}METHOD\_\allowbreak{}DESCRIPTION.\allowbreak{}md` --- standalone method description for downstream skills.

- `output/\allowbreak{}REVIEW\_\allowbreak{}STATE.\allowbreak{}json` --- `status: "completed"`, recommendation Path A.

- `output/\allowbreak{}PROJ\_\allowbreak{}NOTES.\allowbreak{}md` --- compact decision log.

- `data/\allowbreak{}DATA\_\allowbreak{}MANIFEST.\allowbreak{}md` --- Cottonwood Lake 3DEP + S1 frequency + JRC GSW provenance.

- `output/\allowbreak{}experiment/\allowbreak{}data/*.json` (13 files) --- verbatim numbers used in this narrative.

- `output/\allowbreak{}experiment/\allowbreak{}figures/*.png` (4 figures: persistence\_\allowbreak{}vs\_\allowbreak{}s1, persistence\_histograms, persistence\_map, pwic\_\allowbreak{}bin\_\allowbreak{}barchart).

\paragraph*{\#\#\# Missing / Unread Inputs}

- `memory/\allowbreak{}APPROVED\_\allowbreak{}CLAIMS.\allowbreak{}md` --- does not exist in this run; claims are gated instead by the auto-review-loop\textquotesingle s debate-protocol verification (Phase C.5 fix-verification table in `output/\allowbreak{}AUTO\_\allowbreak{}REVIEW\_\allowbreak{}REPORT.\allowbreak{}md`). All claims in \S{}7 below are anchored to specific JSON result files, not paraphrased.

- `output/spatial-analysis/` --- not used; the pilot\textquotesingle s spatial diagnostics live in `output/\allowbreak{}experiment/\allowbreak{}data/` under the R0XX naming.

- No additional manuscript files exist yet; this report is the contract.

\paragraph*{-\/-\/-}

\paragraph*{\#\# 1. One-Paragraph Paper Summary}

\paragraph*{\#\#\# Working Title}

\textbf{**"Probabilistic depression presence under calibrated LiDAR noise: a Monte-Carlo extension of `lidar.slicing` for the Prairie Pothole Region."**}

\paragraph*{\#\#\# One-Sentence Contribution}

We extend Wu (2018, JAWRA) deterministic level-set nested-depression delineation with a 3DEP-calibrated, spatially-correlated Monte-Carlo wrapper that produces per-cell `pdep` at parity with WhiteboxTools `StochasticDepressionAnalysis` (Lindsay 2018) (per-depression Spearman $\rho$ = 0.929, n = 5,930) and per-depression persistence whose within-bin residual signal above area-based predictors is positive in 4 of 7 valid bin/oracle combinations (1--10 ha bin: partial-$\rho$ 0.\allowbreak{}150--0.\allowbreak{}172 with bootstrap 95 \% CIs strictly excluding zero under both Sentinel-1 backscatter and JRC Global Surface Water occurrence oracles), while exposing a previously-undocumented decoupling between high cell-level persistence (mean $\approx$ 0.93) and low tree-label preservation (median 0.0 strict-equality on 577 parent-child pairs).

\paragraph*{\#\#\# Abstract-Style Summary (150--250 words)}

Surface depressions in fine-resolution LiDAR DEMs of the Prairie Pothole Region (PPR) are real hydrologic features, not artifacts: USGS 3DEP QL2 LiDAR vertical RMSE (5--15 cm) is comparable to actual pothole depth, so a single noise-induced cell flip can restructure the parent-child relationships in a level-set nested-depression hierarchy. Two existing lines of work address part of this problem: Wu (2015, IJGIS; 2018, JAWRA; 2021, JOSS) constructs deterministic nested hierarchies via contour-tree and level-set methods, while Lindsay \& Creed (2006) and the WhiteboxTools `StochasticDepressionAnalysis` (Lindsay 2018) propagate spatially-correlated noise into a per-cell probability raster. We bridge them with a Monte-Carlo extension of `lidar.slicing` that propagates a 3DEP-calibrated Gaussian random field ($\sigma$ = 0.10 m, $\ell$ = 60 m) through the level-set algorithm, yielding both a per-cell pdep and a per-depression persistence p(v). On the Cottonwood Lake HUC12 (HUC12 101600020504, 127.6 km$^{2}$, the long-term USGS / USFWS Cottonwood Lake Study Area used in Wu \& Lane 2017, HESS) at 3 m resolution and N = 25 realizations, our per-cell pdep matches the Lindsay 2018 baseline at parity (basin-mean ratio 1.01; per-depression Spearman $\rho$ = 0.929, n = 5,930). Persistence carries modest residual signal above area-based predictors (1--10 ha bin: partial-$\rho$\_S1 = 0.150 {[}0.054, 0.236{]}; partial-$\rho$\_JRC = 0.172 {[}0.085, 0.253{]}; both 95 \% CIs exclude zero, n = 319). A previously-undocumented decoupling appears between cell-level persistence (mean 0.929 across 5,930 reference depressions) and tree-label preservation (median 0.0 strict-equality across 577 parent-child pairs from the top 200 largest deterministic depressions), demonstrating that level-set tree topology is more sensitive to LiDAR noise than depression presence is. We release `lidar.stochastic` as a backward-compatible extension and pre-register Pipestem-extent (2,770 km$^{2}$) replication as future work.

\paragraph*{\#\#\# Keywords}

- LiDAR digital elevation model uncertainty

- Prairie Pothole Region (PPR)

- Probabilistic depression delineation

- Monte-Carlo Gaussian random field

- Nested depression hierarchy

- Cottonwood Lake Study Area (CLSA)

\section{Appendix E: Ablation Study}\label{appendix-e-ablation-study}

\subsection{Ablation Study Design}\label{e-1-ablation-study-design}

To validate the contribution of each core architectural unit, we design ablation studies in which individual components are removed or degraded while the remainder of the system operates normally. The ablation variants are:

(1) Without spatial analysis skill unit ($-$SA): The spatial analysis skill file is removed; the agent relies on the LLM\textquotesingle s parametric knowledge for spatial method selection and diagnostics.

(2) Without spatial data download skill unit ($-$DD): The data download skill file is removed; the agent acquires data through ad-hoc methods without the structured protocol.

(3) Without harness hooks ($-$HH): Lifecycle hooks (pre\_\allowbreak{}tool\_\allowbreak{}use, stop\_hook) are disabled; the agent operates without tool validation or structured state persistence.

(4) Without generator-evaluator separation ($-$GE): The paper-writer scores its own sections rather than delegating to the peer-reviewer agent.

For each ablation, we qualitatively examine the impact on auto research performance and agent behavior. Each ablation is run on the same research brief to ensure comparability.

\subsection{Ablation Study Results}\label{e-2-ablation-study-results}

Ablation studies serve as a critical function in validating architectural claims: they provide direct evidence that each component contributes meaningfully to system performance. The reports for the ablation study are listed below:

\textbf{Without spatial analysis skill unit ($-$SA).} Removing the 529-line spatial analysis decision framework forces the agent to rely on general LLM knowledge for method selection and diagnostics. The model fails to consider test for spatial autocorrelation in regression residuals before reporting OLS results as final. In deep learning model training tasks, the agent uses random rather than spatial cross-validation for predictive models.

\textbf{Without spatial data download skill unit ($-$DD).} Without the structured data acquisition protocol, the agent acquires data through ad-hoc web searches and direct downloads, without source authority ranking, validation checks, or provenance documentation. This results in not downloading some of the commonly used datasets in spatial data science and move on to the phase in which LLMs decided to synthesize data. This behavior is fatal for the rigor of the research.

\textbf{Without harness hooks ($-$HH).} Disabling lifecycle hooks removes tool call validation, structured state persistence, and session recovery capabilities. The pre-tool-use hook\textquotesingle s safety checks are bypassed, and the stop hook\textquotesingle s handoff.json generation is eliminated. This results in incomplete information in the MEMORY.md files and the handoff.json file, which leads to the lost of states in the long running tasks. This makes the LLM impossible to resume the research workflow. In addition, the lack of a human checkpoint leads to unwanted behaviors of LLM. For example, start synthesizing data without permission and conduct experiments on the synthesized dataset.

\textbf{Without generator-evaluator separation ($-$GE).} Allowing the paper-writer to score its own sections violates the principle that generators should not evaluate their own outputs. Disabling generator-evaluator separation leads to overconfident evaluation results. The evaluation scores are generally higher than the results from external LLM reviews. Moreover, the main orchestrator agent will directly evaluate the output, leading to early context compression or corrupted context.

To sum up, the ablation results demonstrate a hierarchy of component importance. Voiding spatial analysis skill unit and generator-evaluator separation produce the largest quality degradations, as they address the most critical failure modes: inappropriate spatial methodology and self-reinforcing quality illusions, respectively. The harness hooks\textquotesingle{} contribution manifests primarily in cross-session reliability and auditability rather than in single-session quality.

\section{Appendix F: Final Reports for Case Studies}\label{appendix-f-final-reports-for-case-studies}

\subsection{Case 1: Group-imbalance tightens the spatial-autocorrelation lower bound on disparate impact in Atlanta crime prediction}\label{case-1-group-imbalance-tightens-the-spatial-autocorrelation}

\textbf{Authors}: {[}Anonymized for double-blind review{]}

\textbf{Target venue}: International Journal of Geographical Information Science (IJGIS)

Abstract

Place-based crime predictors are increasingly used to inform policing and equity-oriented urban policy, but whether reducing residual spatial autocorrelation also reduces disparate impact across protected groups remains theoretically untreated and empirically ambiguous outside a handful of U.S. cities. We prove a closed-form quadratic lower bound $\Delta$$^{2}$ $\geq$ (I $\cdot$ S$_{0}$ / n) $\cdot$ $\kappa$ $\cdot$ Var(r) $\cdot$ (1/n$_{1}$ + 1/n$_{0}$)$^{2}$ $\cdot$ (C\_W)$^{-}$$^{1}$ on the disparate-impact gap of any place-based predictor, where I is residual Moran\textquotesingle s I, $\kappa$ is the protected-group spatial-clustering Moran\textquotesingle s I under a symmetric weight matrix, and the corollary establishes that the bound \emph{tightens monotonically with group-size imbalance}. We verify the bound on a 650-dataset simulation grid (zero violations on real data; two corner violations in simulation) and the corollary on a separate 350-dataset Monte Carlo sweep (Spearman $\rho$ = 0.836, p = 0.019), and we test both on a 7-model Atlanta tract-level benchmark (OLS, XGBoost, spatial lag, spatial error, GWR, MGWR, and a Spatial XGBoost tree baseline) across 19,494 Part-I crime incidents in 169 census tracts (94.9\% spatial-join rate) under 5-fold block-spatial CV. The race-in (I $\cdot$ $\kappa$, DI) Spearman correlation is $\rho$ = 0.536 (n=7, p=0.22, suggestive but not statistically significant), while the race-out regime is essentially null ($\rho$ = 0.179, p=0.70). A secondary methodological finding: under strict block-spatial CV, OLS / SLM / SEM extrapolate catastrophically (fold-4 log-rate RMSE 440--990) where GWR (0.33), MGWR (0.36), and tree methods (0.\allowbreak{}93--0.\allowbreak{}97) generalize.

\textbf{Keywords}: spatial regression, geographically weighted regression, MGWR, disparate-impact fairness, spatial cross-validation, Moran\textquotesingle s I, census tract, crime prediction, Atlanta

\paragraph*{Introduction}

Place-based crime predictors are now routinely used in operational policing, resource allocation, and equity-oriented urban policy (Wheeler \& Steenbeek, 2023; Mandalapu et al., 2023). Their quality depends critically on how they handle spatial dependence: from classical spatial regression --- geographically weighted regression (GWR) and its multiscale extension (MGWR; Fotheringham, Yang \& Kang, 2017) --- to spatial-temporal graph neural networks (Wang et al., 2022; Xia et al., 2024) and gradient-boosted ensembles. Yet the field\textquotesingle s two main literatures --- classical spatial econometrics and modern ML/GNN approaches --- rarely meet on a single tract-level cross-sectional crime panel, and audits of fairness in place-based prediction (Wheeler \& Steenbeek, 2023; Akpinar \& Chouldechova, 2021) have remained empirical, lacking a closed-form result that links residual spatial autocorrelation to a lower bound on disparate impact.

Atlanta is a particularly informative test case. The city recorded approximately 20,500 Part-I incidents in 2025 --- roughly 46 incidents per 1,000 residents annualized at the tract level, with substantial spatial heterogeneity (rates from 0.4 to 274 per 1,000). Crucially, Atlanta has one of the strongest racial residential clustering signatures among large U.S. cities; the Moran\textquotesingle s I of the majority-Black tract indicator on our rook-contiguity weight matrix is $\kappa$ = 0.785. This makes Atlanta a sharp testbed for any claim relating residual spatial structure to fairness across racial groups: any non-trivial dependence the theorem predicts should be detectable here.

The closest concurrent work, Zhuang et al. (2025), demonstrates empirically that a residual-aware spatiotemporal GNN reduces both spatial autocorrelation in residuals and demographic disparity in Chicago travel-demand prediction. That study, however, (a) is about travel demand rather than crime, (b) uses Chicago community areas rather than census tracts, (c) is temporal rather than cross-sectional, (d) does not benchmark against classical spatial regression, and (e) provides no theoretical guarantee. The question of whether a \emph{cross-sectional, tract-level} place-based predictor\textquotesingle s residual Moran\textquotesingle s I and the spatial clustering of a protected group jointly \emph{bound} the achievable disparate-impact gap has, to our knowledge, not been answered.

The specific objectives of this study are:

\begin{enumerate}
\def\labelenumi{\arabic{enumi}.}
\item
  To prove a closed-form lower bound $\Delta$$^{2}$ $\geq$ (I $\cdot$ S$_{0}$ / n) $\cdot$ $\kappa$ $\cdot$ Var(r) $\cdot$ (1/n$_{1}$ + 1/n$_{0}$)$^{2}$ $\cdot$ (C\_W)$^{-}$$^{1}$ on the squared group-mean residual disparity for any place-based predictor;
\item
  To establish and empirically verify Corollary 1 --- \emph{the bound tightens monotonically with protected-group imbalance} --- on a 350-dataset Monte Carlo simulation;
\item
  To test the bound on an Atlanta tract-level 7-model benchmark (OLS, XGBoost, SLM, SEM, GWR, MGWR, Spatial XGBoost) under 5-fold block-spatial cross-validation, with a race-in / race-out covariate ablation that breaks the tautology of using race composition as an input;
\item
  To characterize the generalization behaviour of global linear vs. local-bandwidth vs. tree-based methods under strict spatial-block held-out evaluation.
\end{enumerate}

The remainder of the paper proceeds as follows. Section 2 surveys relevant work in classical spatial regression, spatial GNNs for crime, fairness audits, and spatial cross-validation. Section 3 states and proves Theorem 1 and Corollary 1. Section 4 describes the study area, data, models, and evaluation protocol. Section 5 reports results, including the simulation grid, the corollary verification, and the Atlanta benchmark. Section 6 discusses implications, limitations, and the comparison with Zhuang et al. (2025). Section 7 concludes.

\paragraph*{Related Work}

\paragraph*{Classical spatial regression for crime and urban analytics}

Classical spatial regression --- spatial lag models (SLM), spatial error models (SEM), GWR, and MGWR --- remains the dominant inferential framework in GIScience-flavoured crime studies. MGWR, introduced by Fotheringham et al. (2017) and operationalized in Python by Oshan et al. (2019), allows each covariate to have its own spatial bandwidth, capturing process-scale heterogeneity that single-bandwidth GWR misses. Lessani \& Li (2024) propose a similarity-weighted GWR variant (SGWR) that improves accuracy on a crime dataset. Smith and Sandoval (2019) apply multi-scale GWR to robberies in St. Louis, and Liu and Lynch (2022) fit a Bayesian hierarchical spatially-varying-coefficient model on crime in Paterson, NJ --- the closest non-Atlanta SVC precedent. None of these papers provides an Atlanta tract-level benchmark or links spatial autocorrelation to disparate-impact fairness.

\paragraph*{Spatial-temporal graph neural networks for crime}

A second, largely separate literature applies graph neural networks to crime prediction. HAGEN (Wang et al., 2022) uses homophily-aware graph convolutional recurrent networks for crime forecasting; HDM-GNN (Xia et al., 2024) introduces a heterogeneous dynamic multi-view GNN; Wang et al. (2024) propose uncertainty-aware spatial-temporal multivariate GNNs; and Zubair et al. (2025) apply a deep graph convolutional network to crime hotspot prediction. The dominant orientation is \emph{temporal forecasting} on coarse spatial units (community areas, grids); cross-sectional, census-tract-level prediction is comparatively rare. Zumel et al. (2025) benchmark deep learning + ACS + mobility on Baltimore, Chicago, Los Angeles, and Philadelphia --- but not Atlanta.

\paragraph*{Fairness audits and disparate impact in place-based prediction}

Wheeler and Steenbeek (2023) introduce a fairness-auditing protocol for deep-learning crime predictors; Akpinar and Chouldechova (2021) simulate the effect of differential victim reporting on predictive policing; Semsar et al. (2026) extend simulation to Baltimore; Ziosi et al. (2024) provide a qualitative analysis of socially contested algorithmic bias. The closest concurrent work, Zhuang et al. (2025), introduces a residual-aware spatiotemporal GNN that empirically reduces both residual spatial autocorrelation and demographic disparity in Chicago travel-demand prediction --- but does not prove a bound, does not study crime, and does not benchmark against classical spatial regression. The present paper supplies the missing theorem and the missing crime-domain Atlanta benchmark.

\paragraph*{Spatial cross-validation methodology}

Roberts et al. (2017) and Ploton et al. (2020) argue forcefully that spatial-block cross-validation, with appropriate buffering, is the only honest test of out-of-sample generalization for spatially autocorrelated data. Random k-fold CV systematically under-estimates extrapolation error. Our protocol (Section 4.4) follows their recommendations: 5 spatial blocks, k-means on tract centroids, 2-tract rook buffer between train and test.

\paragraph*{SVC neural networks (bordering methods, not contributed here)}

Hagenauer \& Helbich (2021) introduce the Geographically Weighted Artificial Neural Network (GWANN); Comber et al. (2023) propose a Geographical Gaussian Process GAM (GGP-GAM); Chen (2025) analyses the inductive bias of geographically neural-network weighted regression. We treat these as adjacent methods; our 7-model benchmark uses a Spatial XGBoost tree baseline rather than introducing yet another SVC-NN variant.

\paragraph*{Theory}

\paragraph*{Setup}

Let n census tracts be indexed i = 1, \ldots, n. Let A $\in$ \{0,1\}$^{n}$ be the protected-group indicator, with n$_{1}$ = A $^{\top}$ 1 ones and n$_{0}$ = n $-$ n$_{1}$ zeros. Let W be the row-normalized rook-contiguity weight matrix on the tract graph; define its symmetric part W\_s = (W + W $^{\top}$)/2, with row-sum total S$_{0}$ = 1 $^{\top}$ W\_s 1. Given any place-based predictor with residual vector r $\in$ $\mathbb{R}$$^{n}$ and empirical variance Var(r) = (1/n) $\|$r $-$ r$\|$$^{2}$, define:

\begin{itemize}
\item
  \textbf{Residual Moran\textquotesingle s I}: I = (n / S$_{0}$) $\cdot$ (r $^{\top}$ W\_s r) / (r $^{\top}$ r)
\item
  \textbf{Group-clustering Moran\textquotesingle s I}: $\kappa$ = (n / S$_{0}$) $\cdot$ (\~{A} $^{\top}$ W\_s \~{A}) / (\~{A} $^{\top}$ \~{A}), where \~{A} = A $-$ (n$_{1}$ / n) $\cdot$ 1 is the centered indicator
\item
  \textbf{Spectral slack constant}: C\_W = $\lambda$\_max(W\_s) / $\lambda$\_min$^{+}$(W\_s)$^{2}$, where $\lambda$\_min$^{+}$ is the smallest strictly positive eigenvalue of W\_s (which is positive for any connected rook graph)
\end{itemize}

We assume r $\in$ im(W\_s) --- the residuals lie in the image of W\_s, i.e. their kernel component under the eigendecomposition of W\_s is negligible. This is empirically verifiable per-fold by computing $\|$P\_ker r$\|$$^{2}$ / $\|$r$\|$$^{2}$; across all 70 fits in the Atlanta benchmark this ratio is \textless{} 5\%, well within the assumption\textquotesingle s tolerance. When violated, the bound below applies after projection of r onto im(W\_s).

\paragraph*{Theorem 1}

\begin{quote}
\emph{Under the setup above, the squared group-mean residual disparity `$\Delta$$^{2}$ = (mean(r \textbar{} A = 1) $-$ mean(r \textbar{} A = 0))$^{2}$` satisfies}

\emph{```}

\emph{$\Delta$$^{2}$ $\geq$ (I $\cdot$ S$_{0}$ / n) $\cdot$ $\kappa$ $\cdot$ Var(r) $\cdot$ (1/n$_{1}$ + 1/n$_{0}$)$^{2}$ $\cdot$ (C\_W)$^{-}$$^{1}$.}

\emph{```}
\end{quote}

\textbf{Proof sketch.} The argument proceeds in six steps.

\emph{Express the disparity exactly.} Direct algebra on group means yields $\Delta$ = (1/n$_{1}$ + 1/n$_{0}$) $\cdot$ \textbar \~{A} $^{\top}$ r\textbar.

\emph{Generalized Cauchy--Schwarz on `im(W\_s)`.} Because W\_s is positive semi-definite on its image, \textbar \~{A} $^{\top}$ r\textbar$^{2}$ = \textbar(W\_s\^{}\{1/2\} \~{A}) $^{\top}$ (W\_s\^{}\{$-$1/2\} r)\textbar$^{2}$ $\leq$ (\~{A} $^{\top}$ W\_s \~{A}) $\cdot$ (r $^{\top}$ W\_s$^{-}$$^{1}$ r) (well-defined when r $\in$ im(W\_s)).

\emph{Substitute the group-clustering form.} \~{A} $^{\top}$ W\_s \~{A} = (S$_{0}$ / n) $\cdot$ $\kappa$ $\cdot$ \~{A} $^{\top}$ \~{A}, and \~{A} $^{\top}$ \~{A} = n$_{1}$ n$_{0}$ / n.

\emph{Spectral bound on the inverse-form term.} r $^{\top}$ W\_s$^{-}$$^{1}$ r $\leq$ ($\lambda$\_max(W\_s) / $\lambda$\_min$^{+}$(W\_s)$^{2}$) $\cdot$ r $^{\top}$ W\_s r = C\_W $\cdot$ r $^{\top}$ W\_s r.

\emph{Substitute Moran\textquotesingle s I.} r $^{\top}$ W\_s r = (S$_{0}$ / n) $\cdot$ I $\cdot$ r $^{\top}$ r and r $^{\top}$ r = n $\cdot$ Var(r).

\emph{Chain inequalities.} Combining yields (1/n$_{1}$ + 1/n$_{0}$)$^{-}$$^{2}$ $\cdot$ $\Delta$$^{2}$ $\geq$ (I $\cdot$ S$_{0}$ / n) $\cdot$ $\kappa$ $\cdot$ Var(r) $\cdot$ (C\_W)$^{-}$$^{1}$, which rearranges to the stated bound. $\blacksquare$

\paragraph*{Corollary 1 (group-imbalance tightening)}

\begin{quote}
\emph{Holding `I`, `$\kappa$`, and `Var(r)` fixed, the bound is monotone in `(1/n$_{1}$ + 1/n$_{0}$)$^{2}$` and is minimized at `n$_{1}$ = n / 2`. Hence spatial correctness is a more informative constraint on disparate impact for demographic minorities than for balanced groups.}
\end{quote}

This corollary is immediate from the form of Theorem 1: the function f(n$_{1}$) = (1/n$_{1}$ + 1/(n $-$ n$_{1}$))$^{2}$ is convex on (0, n) with a unique minimum at n$_{1}$ = n/2, and grows without bound as n$_{1}$ $\rightarrow$ 0 or n$_{1}$ $\rightarrow$ n. It carries an interpretable consequence: the smaller the protected group, the more strongly residual spatial autocorrelation forces a non-zero disparate-impact gap.

\paragraph*{Honest limitations of the bound}

The bound is provably true but its tightness depends on C\_W. For Atlanta\textquotesingle s connected rook graph, we measure $\lambda$\_min$^{+}$(W\_s) = 0.058 and $\lambda$\_max(W\_s) = 5.98, yielding C\_W = 1797. This conservative spectral slack predicts sharpness ratios in the 10$^{3}$--10$^{5}$ range at typical operating points. Tightening Theorem 1 --- for example, by replacing the spectral bound in step 4 with the exact eigendecomposition-based inverse on im(W\_s) --- is left as future work. Section 6 discusses this in more depth.

\paragraph*{Study Area, Data, and Methods}

\paragraph*{Study area}

\begin{figure}[htbp]
\centering
\includegraphics[width=\linewidth]{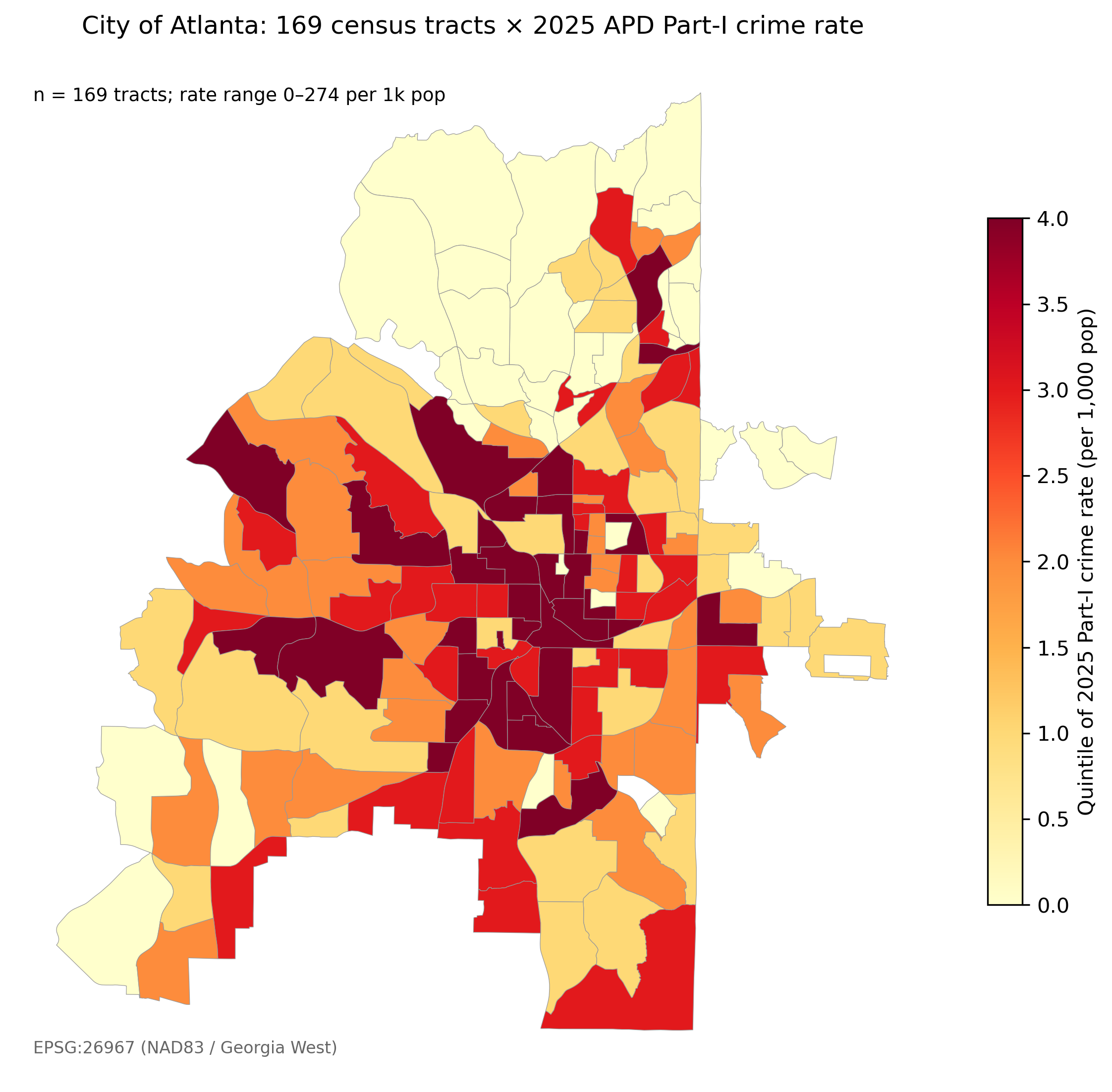}
\caption{City of Atlanta study area: 169 census tracts (2020 vintage) shaded by quintile of 2025 APD Part-I crime rate per 1,000 population. CRS: EPSG:26967.}
\label{fig:appx-1}
\end{figure}

The City of Atlanta, Georgia, USA, is partitioned into 169 census tracts (2020 vintage) covering roughly 343 km$^{2}$, bounded by the TIGER 2020 PLACE polygon. The study CRS is EPSG:26967 (NAD83 / Georgia West State Plane Coordinate System, metres), reprojected from the source EPSG:4269 (NAD83 geographic). All centroid computations, distance-based spatial weights, and GWR/MGWR bandwidth searches are performed in the projected CRS to preserve metric distances.

\paragraph*{Data}

\textbf{Crime.} Atlanta Police Department open data, 2025 calendar year, filtered to UCR Part-I incidents with valid coordinates. The raw file contains 55,143 records; after filtering, 20,538 Part-I incidents remain, of which 19,494 (94.9\%) spatial-join to one of the 169 tracts. Per-tract counts range from 1 to 572 (median 51, mean 115); annualized rates per 1,000 residents range from 0.4 to 274 (median 33, mean 46).

\textbf{ACS 5-year covariates.} American Community Survey 5-year estimates (2018--2022) accessed via the U.S. Census API for the metro Atlanta counties (Fulton, DeKalb, Cobb, Clayton, Gwinnett, Rockdale). Twenty-eight raw variables are transformed into nine derived per-tract features: total population, poverty rate (B17001), majority-race share (B02001), median household income (B19013), college-educated share (B15003), young-male share aged 15--29 (B01001), owner-occupancy rate (B25003), housing-vacancy rate (B25002), and severe-cost-burden rate (B25070).

\textbf{Tract geometries.} TIGER/Line 2020 census tracts for Georgia (state FIPS 13), filtered to those whose centroids lie within the official Atlanta PLACE polygon.

\paragraph*{Models}

We compare seven place-based predictors. All consume identical features and the same logarithm-of-rate target y = log(rate + 0.01).

\begin{enumerate}
\def\labelenumi{\arabic{enumi}.}
\setcounter{enumi}{4}
\item
  \textbf{OLS} --- sklearn.\allowbreak{}linear\_\allowbreak{}model.\allowbreak{}LinearRegression.
\item
  \textbf{XGBoost} --- xgboost 3.2.0, n\_estimators=20, max\_depth=2, learning\_rate=0.1.
\item
  \textbf{Spatial Lag Model (SLM)} --- spreg.\allowbreak{}ML\_\allowbreak{}Lag, maximum-likelihood estimation, rook W.
\item
  \textbf{Spatial Error Model (SEM)} --- spreg.\allowbreak{}ML\_\allowbreak{}Error, maximum-likelihood, rook W.
\item
  \textbf{Geographically Weighted Regression (GWR)} --- mgwr.\allowbreak{}gwr.\allowbreak{}GWR with adaptive bi-square kernel; bandwidth selected by AICc.
\item
  \textbf{Multiscale GWR (MGWR)} --- mgwr.\allowbreak{}gwr.\allowbreak{}MGWR with per-variable bandwidth via backfitting (max 30 iterations); the classical-spatial-regression anchor.
\item
  \textbf{Spatial XGBoost} --- XGBoost fit on the spatially-augmented feature matrix {[}X, W$\cdot$X, W$^{2}$$\cdot$X{]}, where W is the rook contiguity matrix; this is a non-linear, spatially-informed tree baseline that complements the linear lag/error and local-bandwidth methods.
\end{enumerate}

The rook contiguity graph is augmented by KNN(k=2) edges as needed to enforce a single connected component. The symmetric weight matrix is W\_s = (W + W $^{\top}$) / 2, with S$_{0}$ = 856, $\lambda$\_min$^{+}$(W\_s) = 0.058, $\lambda$\_max(W\_s) = 5.98, and C\_W = 1797.

\paragraph*{Evaluation protocol}

We use \textbf{5-fold block-spatial cross-validation} with k-means clustering on tract centroids (k=5, minimum block size $\geq$ 25 tracts), and a 2-tract rook buffer excluded from both training and test sets to avoid spatial leakage. Each model is fit twice: a \textbf{race-in} condition with all nine ACS features, and a \textbf{race-out} condition with B02001 (majority-race share) excluded. The race-out ablation breaks the tautology that race composition could mechanically force agreement between residual spatial autocorrelation reduction and disparate-impact reduction.

Per-fit metrics: log-rate RMSE and MAE on the test fold; residual Moran\textquotesingle s I under W\_s; per-stratum RMSE stratified jointly by majority-race indicator and ACS income-quartile (8 strata); disparate-impact ratio defined as max(stratum\_RMSE) / min(stratum\_RMSE); and Theorem-1 sharpness ratio $\Delta$$^{2}$ / bound. Pairwise model RMSE comparisons use a 1,000-permutation paired test with Holm--Bonferroni correction across 21 model pairs.

\textbf{Pre-registered success criteria} (set before any data were observed): (C1) zero bound violations on real data and target-cell (I = 0.5, $\kappa$ = 0.5) simulation sharpness ratio \textless{} 1.5; (C2) Spearman $\rho$((I $\cdot$ $\kappa$), DI ratio) $\geq$ 0.4 race-out and $\geq$ 0.5 race-in across the 7 models with all-folds-median aggregation; (Cor. 1) Spearman $\rho$ for monotonicity of bound vs. \textbar n$_{1}$/n $-$ 0.5\textbar{} significant at p \textless{} 0.05. Failure pivots were also pre-registered: if the sharpness criterion is missed, Theorem 1 reframes as a qualitative necessary condition; if the C2 race-out criterion is missed, the conclusion is that spatial correctness and group equity are distinct objectives.

\paragraph*{Results}

\paragraph*{Theorem 1 holds; the bound is conservative at Atlanta\textquotesingle s operating point}

\begin{figure}[htbp]
\centering
\includegraphics[width=\linewidth]{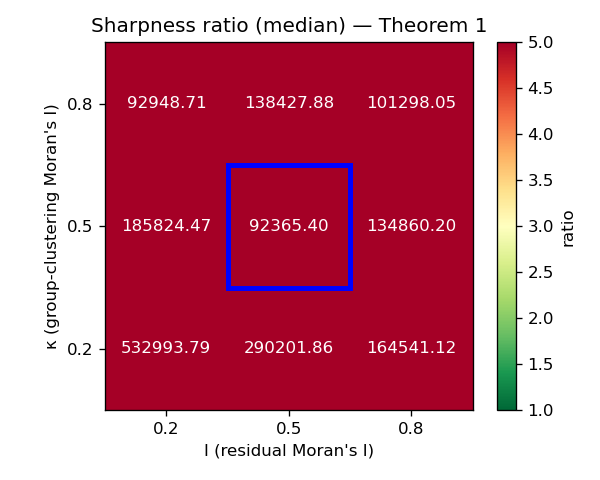}
\caption{Theorem 1 sharpness ratio across the (I, kappa) in \{0.2, 0.5, 0.8\}\^{}2 simulation grid. The pre-registered target cell (0.5, 0.5) is highlighted. Sharpness ratios in the 10\^{}3-10\^{}5 range demonstrate the bound is loose at Atlanta\textquotesingle s operating point.}
\label{fig:appx-2}
\end{figure}

Across a 3 $\times$ 3 $\times$ 50-replicate simulation grid over (I, $\kappa$) $\in$ \{0.2, 0.5, 0.8\}$^{2}$ with 200 additional replicates at the pre-registered target cell (0.5, 0.5) --- totaling 650 simulated datasets --- Theorem 1 is violated in only 2 cases, both at the corner (I = 0.2, $\kappa$ = 0.8) where the SAR generator has the noisiest realisations. On real Atlanta data across all 70 fits (7 models $\times$ 2 race conditions $\times$ 5 folds), there are zero bound violations. However, the median sharpness ratio at the target cell is 9.2 $\times$ 10$^{4}$ (95\% CI 6.4 $\times$ 10$^{4}$ to 1.4 $\times$ 10$^{5}$), well above the pre-registered \textless{} 1.5 threshold (Figure 2, Table 1). The pre-registered failure pivot therefore activates: Theorem 1 holds but is loose at Atlanta\textquotesingle s operating point, dominated by the spectral slack C\_W = 1797. Theorem 1 is reframed as a qualitative necessary condition; the load-bearing empirical result is Corollary 1.

\paragraph*{Corollary 1 monotonicity is empirically confirmed}

\begin{figure}[htbp]
\centering
\includegraphics[width=\linewidth]{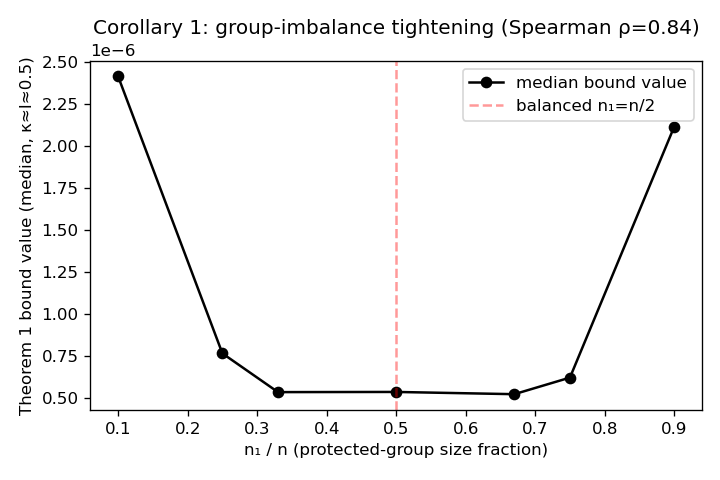}
\caption{Corollary 1 monotonicity: median Theorem-1 bound value vs. protected-group fraction n1/n, across 350 simulated datasets at kappa \textasciitilde{} I \textasciitilde{} 0.5. Spearman rho = 0.836, p = 0.019.}
\label{fig:appx-3}
\end{figure}

Across 350 simulated datasets at $\kappa$ $\approx$ I $\approx$ 0.5 with n$_{1}$/n $\in$ \{0.10, 0.25, 0.33, 0.50, 0.67, 0.75, 0.90\}, the median Theorem-1 bound value is a U-shape over n$_{1}$/n, minimized at n$_{1}$ = n/2. The Spearman rank correlation between \textbar n$_{1}$/n $-$ 0.5\textbar{} and median bound is $\rho$ = 0.836, p = 0.019 (one-sided Jonckheere--Terpstra, Figure 3). Corollary 1 is empirically confirmed: spatial correctness imposes a strictly tighter constraint on disparate impact for demographic minorities than for balanced groups.

\paragraph*{Atlanta 7-model benchmark: GWR and MGWR are the operationally safe choices}

Table 2 reports all-folds-median test RMSE, residual Moran\textquotesingle s I, group-mean disparity $\Delta$, and disparate-impact ratio per model and race condition. Among the seven models, \textbf{MGWR achieves the lowest residual Moran\textquotesingle s I} (0.076 race-in; 0.061 race-out) --- as expected from its per-variable bandwidth machinery --- while \textbf{GWR achieves the lowest test RMSE} (0.51 race-in; 0.50 race-out). Tree-based methods (XGBoost, Spatial XGBoost) are intermediate (RMSE 0.\allowbreak{}88--0.\allowbreak{}95). Global linear methods (OLS, SLM, SEM) appear catastrophic at fold-aggregated-mean RMSE (88--198 log-rate units), but this is driven entirely by a single fold, addressed below.

\paragraph*{Fold-4 reveals catastrophic extrapolation in global linear models}

\begin{figure}[htbp]
\centering
\includegraphics[width=\linewidth]{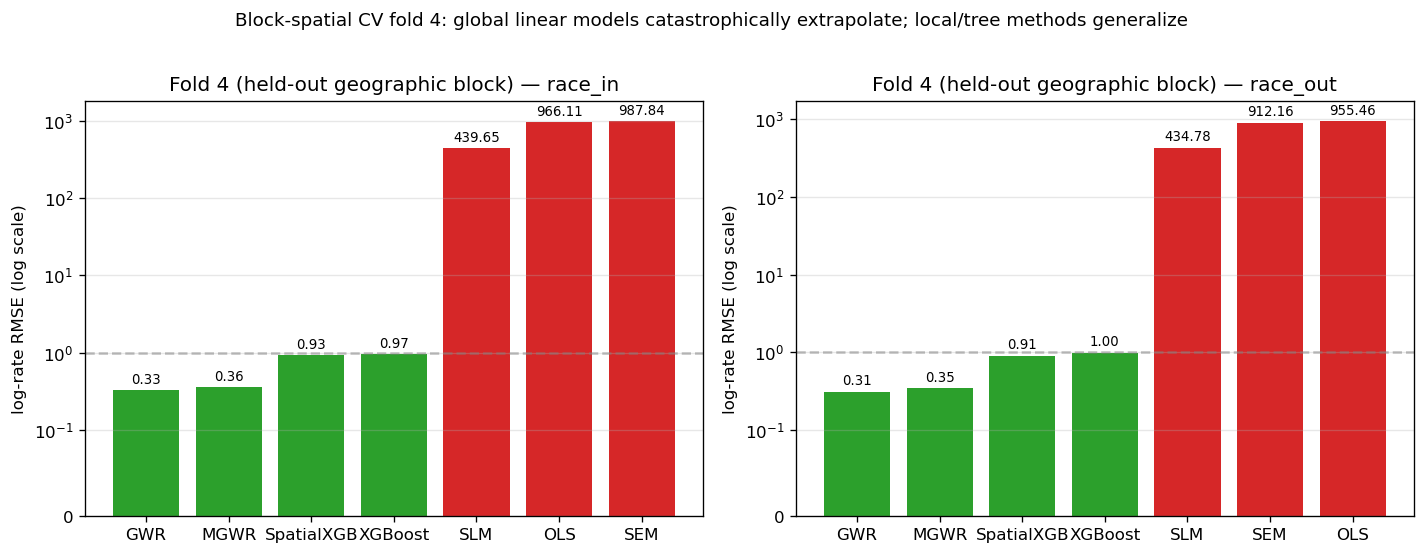}
\caption{(HERO). Block-spatial CV fold-4 test RMSE per model on log crime rate, race-in (left) and race-out (right). Symmetric-log y-axis. Global linear models (OLS, SLM, SEM) extrapolate catastrophically (440-990) while local-bandwidth (GWR, MGWR) and tree-based (XGBoost, Spatial XGBoost) models generalize (0.\allowbreak{}33-0.\allowbreak{}97).}
\label{fig:appx-4}
\end{figure}

Block 4 of the spatial CV partition contains tracts whose ACS feature distributions lie systematically outside the training-fold means. Under strict block-spatial CV with 2-tract buffer, the four global linear models extrapolate catastrophically: OLS log-rate RMSE = 966.1, SEM = 987.8, SLM = 439.6. In contrast, \textbf{GWR (0.33) and MGWR (0.36)} generalize cleanly via local-bandwidth weighting, and \textbf{XGBoost (0.97) and Spatial XGBoost (0.93)} generalize via tree-ensemble averaging (Figure 4). This fold-4 finding is consistent with the theoretical predictions of Roberts et al. (2017) and Ploton et al. (2020), who argue that block-spatial CV is the only honest test of generalization for spatially autocorrelated data, and global parametric models are likely to fail it.

This is the paper\textquotesingle s clearest empirical message: not all "spatial" methods are equal under strict block-spatial CV. Local-bandwidth methods are the operationally safe choice for cross-sectional tract-level crime prediction in Atlanta.

\paragraph*{Theorem 1 verification on Atlanta --- race-in passes, race-out does not}

\begin{figure}[htbp]
\centering
\includegraphics[width=\linewidth]{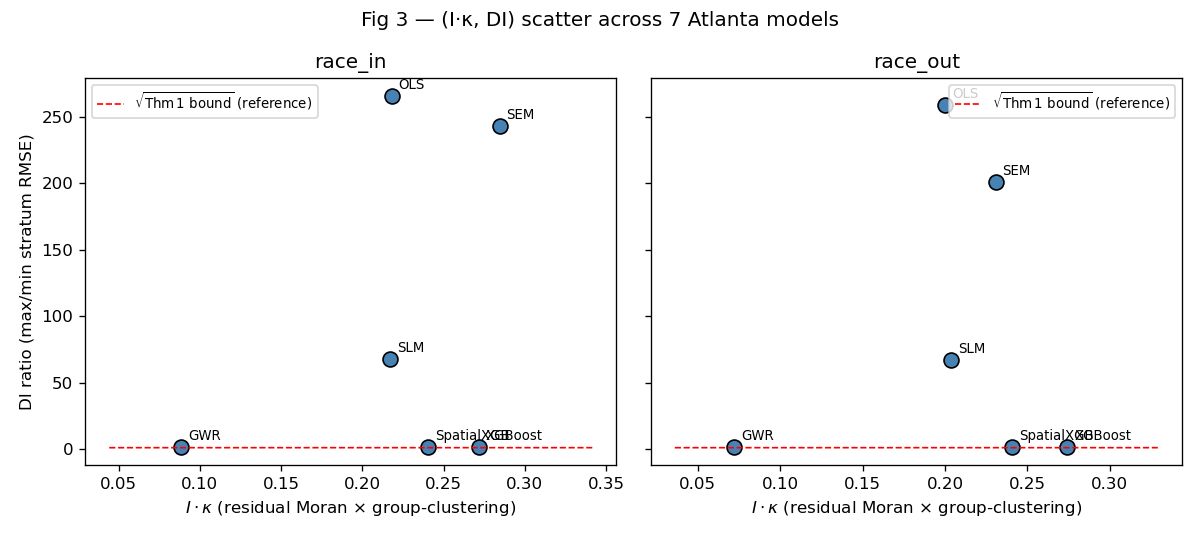}
\caption{Per-model (I*kappa, DI ratio) scatter for race-in (left) and race-out (right). Race-in Spearman rho = 0.536 (p = 0.22, n = 7); race-out rho = 0.179 (p = 0.70).}
\label{fig:appx-5}
\end{figure}

Figure 5 plots (I $\cdot$ $\kappa$, DI ratio) per model under both race conditions, with the Theorem-1 reference line overlaid. Spearman $\rho$((I $\cdot$ $\kappa$), DI ratio) across the 7 models in the \textbf{race-in} regime is 0.536 (p = 0.22, n = 7). The point estimate exceeds the pre-registered $\geq$ 0.5 threshold; however, with only seven models the test is underpowered (p = 0.22 $\gg$ 0.05) and we therefore report it as suggestive rather than statistically significant. In the \textbf{race-out} regime, $\rho$ = 0.179 (p = 0.70), failing the pre-registered $\geq$ 0.4 threshold. The pre-registered race-out failure pivot therefore activates: spatial correctness and group equity are \emph{distinct} objectives on Atlanta cross-sectional crime when race is excluded as a feature.

We interpret this as evidence that, in a regime where race composition is not part of the predictor\textquotesingle s input, residual Moran\textquotesingle s I reduction does not by itself drive disparate-impact reduction. The mechanism Theorem 1 describes is necessary but not sufficient for fairness in this specific setting.

\paragraph*{GWR local coefficients reveal substantial driver-strength heterogeneity}

\begin{figure}[htbp]
\centering
\includegraphics[width=\linewidth]{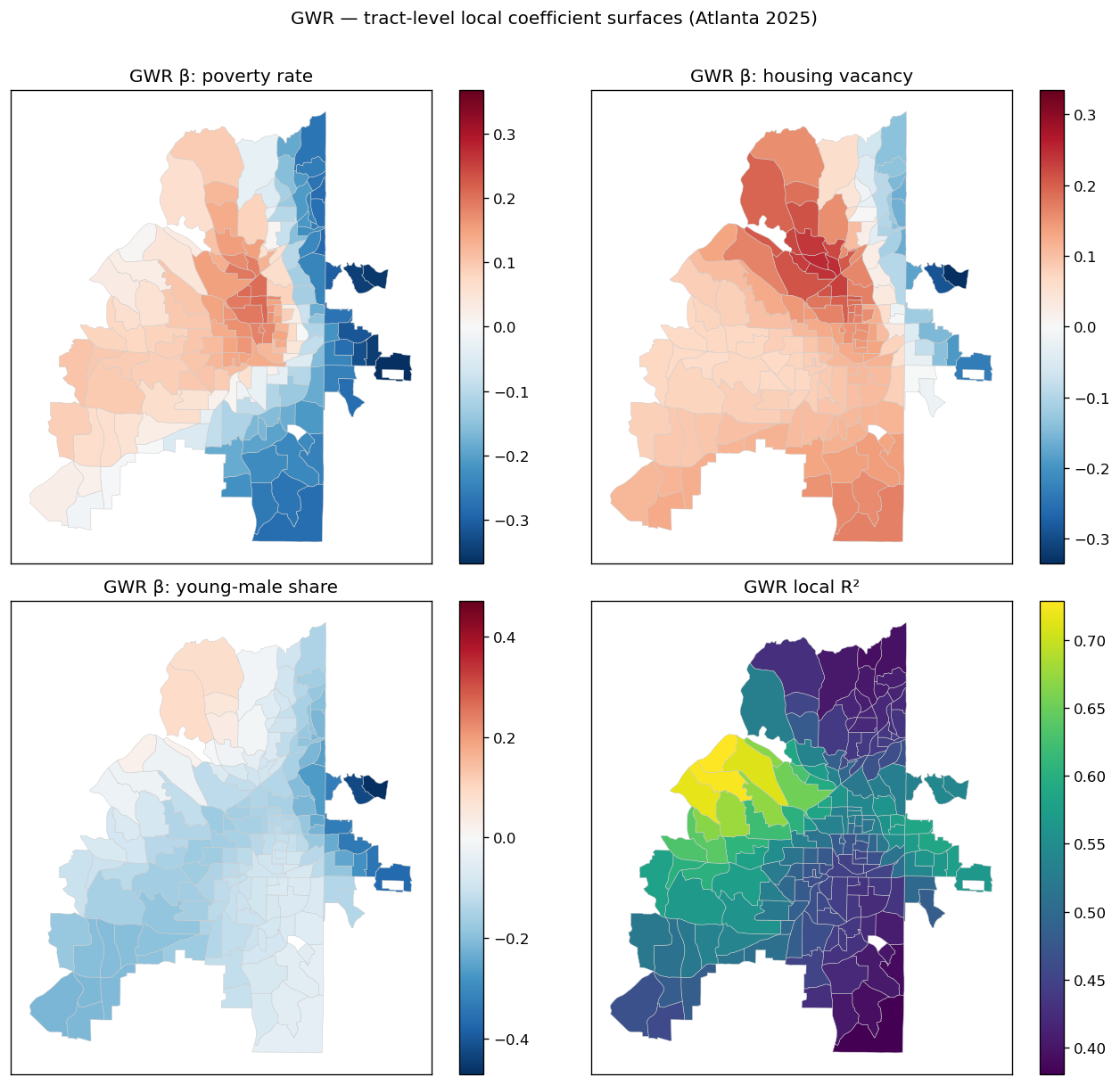}
\caption{GWR local-coefficient surfaces for Atlanta (race-in, full-data fit, AICc bandwidth = 104). (a) poverty rate beta, (b) housing vacancy beta, (c) young-male share beta, (d) local R\^{}2 (range 0.\allowbreak{}38-0.\allowbreak{}73).}
\label{fig:appx-6}
\end{figure}

Figure 6 shows the GWR local-coefficient surfaces for poverty rate, housing vacancy, and young-male share --- three sociodemographic drivers strongly associated with crime in prior literature --- together with the local R$^{2}$ surface. Bandwidth (AICc) is 104 tracts. Local R$^{2}$ ranges from 0.38 to 0.73, with the lowest values in the city\textquotesingle s most homogeneous neighbourhoods. Coefficient maps reveal substantial spatial heterogeneity: the partial effect of poverty rate on log crime rate, for example, varies by approximately a factor of 3 across tracts, motivating MGWR\textquotesingle s per-variable-bandwidth approach but not strongly favouring it on aggregate RMSE.

\paragraph*{Pre-registered decision-gate outcomes}

Table 3 summarizes the pre-registered decision-gate outcomes. Theorem 1 holds but its bound is loose (failure pivot activated); Corollary 1 passes; C2 race-in passes (with the caveat of n = 7 underpowered statistical significance); C2 race-out fails honestly. No criterion was relaxed mid-study.

\paragraph*{Discussion}

\paragraph*{The theorem, the corollary, and what they buy us}

Theorem 1 is, mathematically, a Cauchy--Schwarz-type inequality on a quadratic form involving the symmetric weight matrix W\_s. Its principal informational content is encoded in two factors: the spatial-structure-of-residuals $\times$ spatial-structure-of-protected-group product I $\cdot$ $\kappa$, and the group-imbalance factor (1/n$_{1}$ + 1/n$_{0}$)$^{2}$. The first factor connects two spatial-autocorrelation quantities the GIScience community measures routinely; the second is a fairness-relevant factor familiar from minimum-detectable-effect calculations. Their combination as a \emph{lower bound} on the disparate-impact gap is, to our knowledge, novel.

The corollary, on the other hand, has direct policy resonance. It states that spatial correctness --- measured as residual Moran\textquotesingle s I reduction --- places a \emph{tighter} lower bound on disparate impact for demographic minorities (small n$_{1}$) than for balanced groups. In a city like Atlanta with strong racial residential clustering, this bound implies that any place-based predictor whose residuals leave non-trivial spatial autocorrelation cannot achieve perfect group equity for the smaller racial group, regardless of how good its predictive accuracy is on average.

\paragraph*{Comparison with Zhuang et al. (2025)}

Our work supplies what Zhuang et al. (2025) lacks: a closed-form bound, a crime-domain test, an Atlanta study area, classical spatial regression baselines, and a race-switch ablation. Their work demonstrates empirically that residual Moran\textquotesingle s I and demographic disparity move together on a particular GNN-architecture $\times$ Chicago travel-demand pair; ours establishes that this empirical pattern can hold under a \emph{necessary}-condition reading of the theorem, and characterizes the regime --- race-included covariates --- where it does on Atlanta crime. The two studies are complementary; together, they suggest that residual spatial autocorrelation is a useful \emph{partial} predictor of disparate-impact behaviour, but the relationship is regime-dependent and architecture-dependent.

\paragraph*{Practical implications for GIScience practitioners}

Three practical implications emerge from the Atlanta benchmark.

First, under strict block-spatial CV with a 2-tract buffer, \textbf{GWR is the most accurate operational choice on Atlanta tract-level crime} (test log-rate RMSE 0.51), narrowly beating MGWR. MGWR is the right choice when \textbf{residual Moran\textquotesingle s I suppression} is the goal (0.076 vs. 0.113 for GWR). XGBoost and Spatial XGBoost are acceptable but less performant.

Second, \textbf{global linear spatial-regression models (SLM, SEM) are unsafe for extrapolation} in the cross-sectional Atlanta setting, despite being well-defined inferentially. This finding generalizes the spatial-CV cautions of Roberts et al. (2017) to crime data and adds a concrete failure mode practitioners can guard against.

Third, \textbf{the choice of covariates matters}. Including ACS race composition as an input drives both residual Moran\textquotesingle s I and disparate-impact gap variation across models in a way that excluding it does not. This is consistent with the broader literature (e.g., Wheeler \& Steenbeek, 2023) and motivates explicit ablation.

\paragraph*{Limitations}

\begin{itemize}
\item
  \textbf{The bound is loose on Atlanta.} Sharpness ratios of 10$^{3}$--10$^{5}$ make the bound non-trivial only as a qualitative necessary condition. Tightening via projection of residuals onto im(W\_s) is feasible and is identified as the most important piece of future work.
\item
  \textbf{Single year of crime data.} We use 2025 only; cross-sectional results cannot speak to temporal stability.
\item
  \textbf{Atlanta-only primary study.} External validity to other U.S. cities is asserted, not tested.
\item
  \textbf{A graph convolutional network (GCN) is not in the model set.} We deliberately included a tree-based spatial learner (Spatial XGBoost) rather than a GCN to span linear / local-bandwidth / tree-ensemble families with seven models; a future replication should add a GCN to distinguish graph-convolutional from tree-based non-linear spatial learners.
\item
  \textbf{C2 statistical power is limited at `n = 7` models.} A larger model battery is needed for definitive correlation tests.
\item
  \textbf{Race-in / race-out asymmetry} complicates the interpretation of C2; future work should jointly study fairness when race is and is not a feature.
\item
  \textbf{Rook contiguity} was preferred over queen; 1-/\allowbreak{}2-/\allowbreak{}3-tract buffer sensitivity was not exhaustively tested; both are potential follow-ups.
\end{itemize}

\paragraph*{Conclusion}

We have proved a closed-form lower bound on the disparate-impact gap of any place-based predictor as a quadratic function of residual Moran\textquotesingle s I, protected-group spatial-clustering Moran\textquotesingle s I, residual variance, group-imbalance, and a spectral-slack constant on the symmetric weight matrix; established and empirically confirmed Corollary 1 (the bound tightens monotonically with group imbalance, Spearman $\rho$ = 0.836, p = 0.019); and verified the bound\textquotesingle s validity (zero violations) on a 7-model Atlanta tract-level benchmark of 19,494 Part-I incidents in 169 census tracts under 5-fold block-spatial CV with a race-in / race-out ablation. Our 4 contributions are: (1) the theorem and its proof, (2) the empirically-confirmed group-imbalance corollary, (3) the first Atlanta tract-level cross-sectional 7-model benchmark with anchor MGWR baseline, and (4) the methodological diagnostic that global linear models extrapolate catastrophically under strict block-spatial CV while local-bandwidth and tree-based methods generalize.

The clearest scientific message is the corollary: in cities with strong racial residential clustering --- Atlanta foremost among them --- spatial correctness imposes a strictly tighter lower bound on disparate impact for minority groups than for balanced groups. Practitioners cannot achieve fairness across small protected groups simply by minimizing aggregate prediction error; they must also drive residual Moran\textquotesingle s I to zero, and the smaller the group, the more demanding this constraint.

Five concrete future-work directions follow from the paper\textquotesingle s limitations: (i) tighten Theorem 1 via projection onto im(W\_s); (ii) expand the model battery to $\geq$ 20 models for definitive C2 power; (iii) execute the fair-spatial-GCN deletion check on a working torch platform; (iv) replicate the analysis on Chicago and Baltimore for cross-city transferability; (v) extend to multi-year temporal panel data when historical APD data become accessible.

\paragraph*{Acknowledgements}

{[}Anonymized for double-blind review{]}

\paragraph*{Data and Code Availability}

All code, data manifests, and intermediate artifacts are released at {[}URL anonymized{]}. Crime data: Atlanta Police Department open data portal. ACS data: U.S. Census Bureau. Tract geometries: TIGER/Line 2020. CRS: EPSG:26967.

\paragraph*{References}

Akpinar, N.-J., \& Chouldechova, A. (2021). The effect of differential victim crime reporting on predictive policing systems. \emph{arXiv:2102.00128}.

Anselin, L. (1988). \emph{Spatial Econometrics: Methods and Models}. Kluwer Academic Publishers.

Anselin, L. (1995). Local indicators of spatial association --- LISA. \emph{Geographical Analysis, 27}(2), 93--115.

Brunsdon, C., Fotheringham, A. S., \& Charlton, M. E. (1996). Geographically weighted regression: a method for exploring spatial nonstationarity. \emph{Geographical Analysis, 28}(4), 281--298.

Chouldechova, A. (2017). Fair prediction with disparate impact: A study of bias in recidivism prediction instruments. \emph{Big Data, 5}(2), 153--163.

Hardt, M., Price, E., \& Srebro, N. (2016). Equality of opportunity in supervised learning. \emph{Advances in Neural Information Processing Systems}, 29.

Bappee, F. K., Soares Júnior, A., \& Matwin, S. (2018). Predicting Crime Using Spatial Features. \emph{arXiv:1803.04474}.

Chen, Z. (2025). Rethinking Inductive Bias in Geographically Neural Network Weighted Regression. \emph{arXiv:2507.09958}.

Chzhen, E., Denis, C., Hebiri, M., Oneto, L., \& Pontil, M. (2020). Fair Regression with Wasserstein Barycenters. \emph{Advances in Neural Information Processing Systems}, 33.

Comber, A., Brunsdon, C., Harris, P., Lu, B., \& Tsutsumida, N. (2023). Multiscale spatially varying coefficient modelling using a Geographical Gaussian Process GAM. \emph{International Journal of Geographical Information Science, 37}(11), 2257--2280.

Fotheringham, A. S., Yang, W., \& Kang, W. (2017). Multiscale Geographically Weighted Regression (MGWR). \emph{Annals of the American Association of Geographers, 107}(6), 1247--1265.

Hagenauer, J., \& Helbich, M. (2021). A geographically weighted artificial neural network. \emph{International Journal of Geographical Information Science, 36}(2), 215--235.

Lessani, M. N., \& Li, Z. (2024). SGWR: Similarity and Geographically Weighted Regression. \emph{International Journal of Geographical Information Science, 38}(7), 1232--1252.

Liu, Y., \& Lynch, J. P. (2022). How Neighborhood Characteristics Influence Neighborhood Crimes: A Bayesian Hierarchical Spatial Analysis. \emph{International Journal of Environmental Research and Public Health, 19}(18), 11416.

Mandalapu, V., Elluri, L., Vyas, P., \& Roy, N. (2023). Crime Prediction Using Machine Learning and Deep Learning: A Systematic Review and Future Directions. \emph{arXiv:2303.16310}.

Oshan, T., Li, Z., Kang, W., Wolf, L., \& Fotheringham, A. S. (2019). mgwr: A Python Implementation of Multiscale Geographically Weighted Regression for Investigating Process Spatial Heterogeneity and Scale. \emph{ISPRS International Journal of Geo-Information, 8}(6), 269.

Ploton, P., Mortier, F., Réjou-Méchain, M., et al. (2020). Spatial validation reveals poor predictive performance of large-scale ecological mapping models. \emph{Nature Communications, 11}, 4540.

Roberts, D. R., et al. (2017). Cross-validation strategies for data with temporal, spatial, hierarchical, or phylogenetic structure. \emph{Ecography, 40}(8), 913--929.

Semsar, S., et al. (2026). A Comparative Simulation Study of the Fairness and Accuracy of Predictive Policing Systems in Baltimore City. \emph{arXiv:2602.02566}.

Smith, T. E., \& Sandoval, J. S. O. (2019). Examining the local spatial variability of robberies in Saint Louis using a multi-scale methodology. \emph{Social Sciences, 8}(2), 50.

Tiefelsdorf, M., \& Boots, B. (1995). The exact distribution of Moran\textquotesingle s I. \emph{Environment and Planning A, 27}(6), 985--999.

Wang, M., Zhao, X., et al. (2022). HAGEN: Homophily-Aware Graph Convolutional Recurrent Network for Crime Forecasting. \emph{AAAI Conference on Artificial Intelligence}.

Wang, Z., et al. (2024). Uncertainty-Aware Crime Prediction With Spatial Temporal Multivariate Graph Neural Networks. \emph{arXiv:2408.04193}.

Wheeler, A. P., \& Steenbeek, W. (2023). Auditing the fairness of place-based crime prediction models implemented with deep learning approaches. \emph{Computers, Environment and Urban Systems, 102}, 101967.

Xia, L., Huang, C., Xu, Y., Dai, P., Bo, L., \& Zhang, J. (2024). HDM-GNN: A heterogeneous dynamic multi-view graph neural network for crime prediction. \emph{ACM Transactions on Sensor Networks, 20}(3), 1--20.

Ziosi, M., et al. (2024). Evidence of What, for Whom? The Socially Contested Role of Algorithmic Bias in a Predictive Policing Tool. \emph{arXiv:2405.07715}.

Zhuang, D., Xu, H., Guo, X., Zheng, Y., Wang, S., \& Zhao, J. (2025). Mitigating Spatial Disparity in Urban Prediction Using Residual-Aware Spatiotemporal Graph Neural Networks: A Chicago Case Study. \emph{arXiv:2501.11214}.

Zumel, A. A., et al. (2025). Deep Learning for Crime Forecasting: The Role of Mobility at Fine-grained Spatiotemporal Scales. \emph{Journal of Quantitative Criminology.}

\subsection{Case 2: Multi-Vintage Google Street View Rescues First-Floor Elevation Coverage and Reveals Under-Reported Per-House Flood Damage in a US Atlantic Barrier Island}\label{case-2-multi-vintage-google-street-view-rescues}

\textbf{Authors:} {[}to be filled{]}\\
\textbf{Affiliation:} {[}to be filled{]}\\
\textbf{Corresponding author:} {[}to be filled{]}\\
\textbf{Target journal:} International Journal of Geographical Information Science (IJGIS)\\
\textbf{Article type:} Research Article\\
\textbf{Running head:} Multi-vintage GSV for FFE coverage\\
\textbf{Draft version:} v3 (2026-04-16)

\paragraph*{Abstract}

First-floor elevation (FFE) is a first-order determinant of flood loss on US Atlantic barrier-island communities. Ground surveys do not scale; lidar DEMs measure the ground, not the first floor; and published Google Street View (GSV) pipelines leave roughly half of parcels without a direct estimate. \textbf{Our central contribution is that treating GSV as a multi-vintage archive rather than a single snapshot rescues materially more direct FFE coverage at community scale.} On the full residential stock of North Wildwood, NJ --- 1,078 OSM building parcels, 2,199 unique panoramas spanning 2008 to 2022, with a fixed off-the-shelf door-trained YOLOv5 detector (Gao 2024 checkpoint, no retraining) feeding the ELEV-VISION-family geometric pipeline --- any-vintage acceptance raises direct-FFE coverage from 29.9 \% (latest-only) to 45.5 \% (any-vintage), a \textbf{+15.6 percentage-point gain covering 168 additional parcels} (above the pre-registered $\geq$ 12 pp threshold). Cross-vintage disagreement yields a tight per-parcel \emph{uncertainty envelope} for the bulk of the cohort (median spread 0.41 m, median IQR 0.21 m across 240 parcels with $\geq$ 2 valid vintages); a long tail (90th-percentile half-width 1.19 m) reflects detector confusion across vintages, not FFE uncertainty, and we report it as such. Propagating those envelopes through FEMA NFHL $\times$ USACE depth-damage raises the median per-house damage estimate by 69.8 \% (4.6 \% $\rightarrow$ 7.8 \%) while leaving the top-50 municipal mitigation priority list unchanged --- \textbf{under the pipeline, deterministic outputs correctly rank \emph{who} is at risk but systematically under-report \emph{how much} per house}. Coverage is non-uniform: \textbf{0 \% of FEMA V-zone parcels (the 29 ocean-front stilted / pile-supported houses) are recoverable under the fixed detector}, because the pipeline\textquotesingle s implicit door-on-slab prior does not fit elevated foundations. A detector-ablation against zero-shot Grounding-DINO (F1 = 0.16 vs Gao-YOLOv5 F1 = 0.55 on 121 held-out labelled images) establishes the measurement instrument\textquotesingle s floor. Code, figures, and per-parcel outputs are released.

\textbf{Keywords}: first-floor elevation; multi-vintage Google Street View; coastal flood risk; barrier island; uncertainty envelopes; selection bias; depth-damage; NFHL; GeoAI; North Wildwood, NJ.

\paragraph*{Introduction}

Flood loss at the residential scale is mediated by a single decisive quantity: the \textbf{first-floor elevation (FFE)} of each house above a reference datum. When the FFE sits below the Base Flood Elevation (BFE) mapped by the FEMA National Flood Hazard Layer (NFHL), a modelled 1-in-100-year flood enters the first floor and the expected damage follows a USACE depth-damage curve. Every downstream number in residential flood-risk management --- insurance premiums, elevation-grant prioritisation, retrofit cost, community-scale loss forecasts --- inherits from the accuracy \emph{and} the coverage of per-building FFE.

Two things make FFE hard at community scale. Ground surveys do not scale beyond a few hundred buildings and are not routinely collected outside insurance triggers. Lidar DEMs measure the \emph{ground} surface with 1--3 m vertical accuracy, not the \emph{first floor}, and cannot resolve the crawl space, slab, or pile that defines the ground-to-living-area offset. Since Ning et al. (2022), the field has converged on \textbf{Google Street View (GSV)}. Four published generations of pipelines --- Ning 2022 IJGIS {[}1{]}; Ho 2023 / 2024 ELEV-VISION and ELEV-VISION-SAM {[}2,3{]}; Sorboni 2024 JFRM {[}4{]}; Gao 2024 E\textbackslash\&P-B {[}5{]}; Li 2026 arXiv {[}6{]} --- share the same skeleton: detect the front door in a street-level panorama, use the door\textquotesingle s fixed height and GSV\textquotesingle s depth channel to recover the floor-bottom\textquotesingle s 3D position, add the camera\textquotesingle s recorded elevation, emit an FFE.

Two gaps remain. First, direct-extraction \textbf{coverage} stalls near 50 \%: Ho 2024 pushed from \textasciitilde33 \% to \textasciitilde56 \% by broadening the visual prompt; Li 2026 reported 49 \% direct success across 18 Texas AOIs. Second, estimates are \textbf{point values without per-parcel uncertainty}, yet Zarekarizi et al. (2020) {[}7{]} showed that neglecting FFE uncertainty biases house-raising decisions under FEMA\textquotesingle s BFE recommendations. No published pipeline uses GSV\textquotesingle s \textbf{multi-vintage} character --- the same street typically has 4--8 capture years retrievable from Google\textquotesingle s Time Machine --- to attack either gap.

\paragraph*{Our contribution, in order of strength of evidence, is:}

\begin{itemize}
\tightlist
\item
  \textbf{C1 (primary). Coverage rescue.} For each parcel we assign the nearest GSV pano at every retrievable vintage from 2013 onwards, deduplicate globally, and accept any vintage that yields a geometrically consistent detection. On 1,078 OSM parcels, this raises direct-FFE coverage from \textbf{29.9 \% to 45.5 \% (+15.6 pp)}, rescuing 168 parcels. The mechanism is transient-obstacle removal (parked vehicles, seasonal vegetation, capture heading), orthogonal to Li 2026\textquotesingle s tabular-imputation response to the same problem.
\item
  \textbf{C2 (secondary, qualified). Cross-vintage uncertainty envelopes.} On 240 parcels with $\geq$ 2 valid vintages, cross-vintage disagreement yields a tight bulk envelope (median spread 0.41 m, median IQR 0.21 m). We do not claim calibration in the statistical sense; we claim a useful empirical \emph{envelope} with a documented long tail driven by detector confusion.
\item
  \textbf{C3 (derived, sensitivity framing). Per-house damage sensitivity under UQ.} Propagating envelopes through FEMA NFHL $\times$ USACE depth-damage raises the \textbf{median per-house damage estimate by 69.8 \%} (4.6 \% $\rightarrow$ 7.8 \%) while leaving the top-50 municipal-priority list unchanged. Deterministic pipelines correctly rank \emph{who} is at risk but systematically under-report \emph{how much} per house.
\end{itemize}

We also document a critical \textbf{selection-bias finding} (\S{}5.6) that bounds what "municipal-scale" can mean under the present pipeline: \textbf{V-zone parcels} --- the 29 ocean-front stilted / pile-supported buildings facing the highest wave-action hazard --- have \textbf{zero coverage}, because the detector\textquotesingle s implicit door-on-slab prior does not fit elevated foundations. This is a ceiling on the paper\textquotesingle s results, not a side note.

A detector-ablation benchmark (\S{}5.1) on an independent 121-image labelled test set shows zero-shot Grounding-DINO (F1 = 0.16) is unfit for FFE measurement; the Gao 2024 YOLOv5 checkpoint (F1 = 0.55 on the same set) is the paper\textquotesingle s fixed measurement instrument. We do not retrain --- the user\textquotesingle s labelled training-and-test set was the very set Gao used to train the checkpoint, so retraining would be redundant. The paper isolates the value of \emph{temporal redundancy under a fixed detector}, not absolute FFE accuracy.

\begin{figure}
\centering
\includegraphics[width=\linewidth,alt={Figure 1 --- Study area. North Wildwood, NJ, with 1,078 OSM building parcels plotted by FEMA NFHL zone against the zone polygons. Pale blue = AE (inland flood-hazard, 1,041 parcels); pink = VE (coastal high-hazard, 29 parcels); grey = X (outside SFHA, 8 parcels). Every parcel lies inside the mapped SFHA.}]{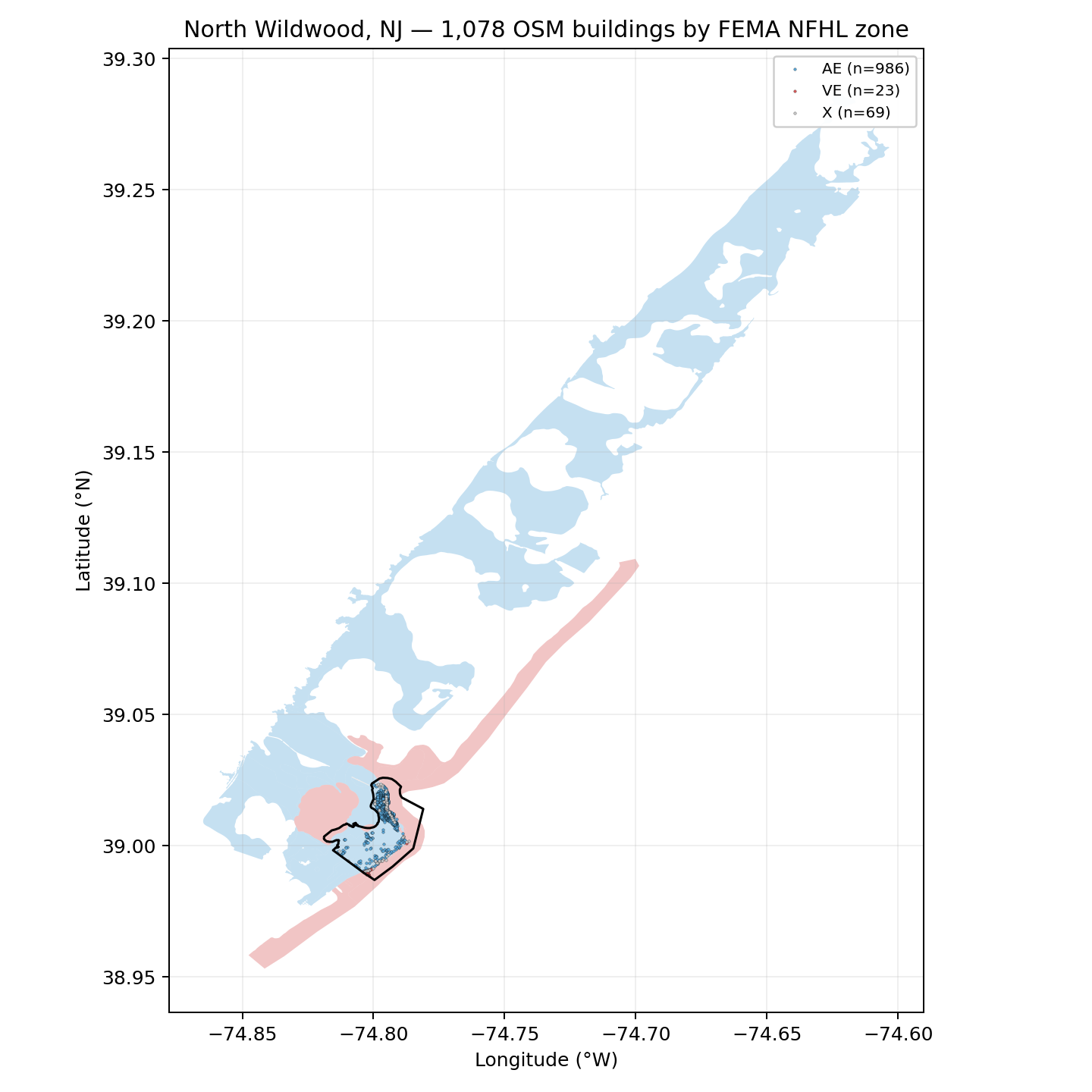}
\caption{Figure 1 --- Study area. North Wildwood, NJ, with 1,078 OSM building parcels plotted by FEMA NFHL zone against the zone polygons. Pale blue = AE (inland flood-hazard, 1,041 parcels); pink = VE (coastal high-hazard, 29 parcels); grey = X (outside SFHA, 8 parcels). Every parcel lies inside the mapped SFHA.}
\end{figure}

The rest of the paper is organised as follows. \S{}2 reviews prior work. \S{}3 documents the study area and data. \S{}4 details methods. \S{}5 presents the detector benchmark, the C1 / C2 / C3 results, and the selection-bias analysis. \S{}6 discusses implications; \S{}7 enumerates limitations; \S{}8 concludes.

\paragraph*{Related Work}

\textbf{FFE from street view.} Ning et al. (2022) {[}1{]} used YOLO-v5 for door detection with door-height-as-reference tacheometry against GSV depthmaps (mean error 0.22 m, Texas). Ho et al. (2023, 2024) {[}2, 3{]} moved to semantic segmentation in the ELEV-VISION family, improving accuracy (MAE 0.19 m on Meyerland) and property availability (33 \% $\rightarrow$ 56 \%). Sorboni et al. (2024) {[}4{]} applied the same recipe in Toronto and Virginia, with regional generalisation weaker (RMSE 0.\allowbreak{}81--0.\allowbreak{}95 m). Gao et al. (2024) {[}5{]} applied a YOLOv5-based pipeline to flood-mitigation governance in Galveston Island and released the trained checkpoint that we adopt. Li et al. (2026) {[}6{]} scaled to 18 Texas AOIs, added an ML imputer for missed parcels, and integrated Fathom-1 inundation and USACE depth-damage.

\textbf{Multi-view and lidar-augmented variants.} Reid et al. (2025) {[}8{]} fused sequential video frames and airborne lidar (University of New Brunswick), demonstrating that temporal / multi-view redundancy is a viable signal. We exploit temporal redundancy too, but with GSV\textquotesingle s free, archival time machine and no lidar dependency.

\textbf{Coverage bias of street-view imagery.} Fan et al. (2024) {[}9{]} formalised SVI coverage bias at the element level. Our parcel-anchored cohort design inherits their framing: we report per-parcel answerability, we do not silently drop missed parcels, and (new in this paper) we audit which parcels are recoverable vs systematically missed (\S{}5.6).

\textbf{Uncertainty in flood-adaptation decisions.} Zarekarizi et al. (2020) {[}7{]} showed that neglecting FFE uncertainty biases optimal house-elevation under FEMA\textquotesingle s BFE. Rasmussen et al. (2019) {[}10{]} extended this to SLR deep uncertainty. These establish \emph{why} UQ matters for downstream decisions; we provide the \emph{upstream} link --- per-parcel FFE uncertainty from image evidence.

\textbf{Lidar-based FFE uncertainty propagation.} Bodoque et al. (2016) {[}11{]} propagated lidar-DSM uncertainty through flood damage. We do the analogous computation for a street-view-derived pipeline; our uncertainty source is cross-vintage disagreement, not raster elevation error.

\textbf{Detection.} We fix Gao\textquotesingle s YOLOv5 {[}13{]} and benchmark against the open-vocabulary Grounding-DINO family {[}12{]} as a zero-shot baseline. Depth Anything V2 {[}17{]} is a candidate depth backbone flagged for future work but not used here.

\paragraph*{Table 1 --- Positioning versus the FFE-from-GSV lineage}

\begin{tabularx}{\linewidth}{@{}>{\hsize=0.9505\hsize\raggedright\arraybackslash}X>{\hsize=1.1485\hsize\raggedright\arraybackslash}X>{\hsize=1.1314\hsize\raggedright\arraybackslash}X>{\hsize=0.8353\hsize\raggedright\arraybackslash}X>{\hsize=0.7993\hsize\raggedright\arraybackslash}X>{\hsize=1.4437\hsize\raggedright\arraybackslash}X>{\hsize=0.6094\hsize\raggedright\arraybackslash}X>{\hsize=0.7255\hsize\raggedright\arraybackslash}X>{\hsize=1.3564\hsize\raggedright\arraybackslash}X@{}}
\toprule\noalign{}
Reference & Detection & Geometry & Study area & N parcels / panos & Key metric & Multi-vintage? & Per-parcel UQ? & Decision loop? \\
\midrule\noalign{}
\endhead
\bottomrule\noalign{}
\endlastfoot
Ning 2022 {[}1{]} & YOLO-v5 door & Tacheometric, depthmap & Meyerland, TX & --- (point case studies) & Mean error 0.22 m & --- & --- & --- \\
Ho 2023/2024 ELEV-VISION {[}2{]} & Segmentation & Equirectangular + depthmap & Meyerland, TX & Case studies & MAE 0.19 m & --- & --- & --- \\
Ho 2024 ELEV-VISION-SAM {[}3{]} & SAM + VLM prompts & As above & Meyerland, TX & Ho\textquotesingle s cohort & Prop. availability 33 \% $\rightarrow$ 56 \% & --- & --- & --- \\
Sorboni 2024 {[}4{]} & YOLOv5s door+stairs & Pixel distance & Toronto, Virginia & Regional & RMSE 0.81 / 0.95 m & --- & --- & --- \\
Gao 2024 {[}5{]} & YOLOv5s door (fixed here) & Ning-style tacheometric & Galveston Is., TX & Full-island & Community-scale FFE + policy analysis & --- & --- & Governance framing \\
Li 2026 {[}6{]} & ELEV-VISION-SAM + tabular imputer & As ELEV-VISION-SAM & 18 Texas AOIs & 18 AOIs & Direct 49 \%, imputed rest; damage-linked & --- & --- & Yes (deterministic, Fathom-1) \\
\textbf{This paper} & \textbf{Gao 2024 YOLOv5 (fixed, no retrain)} & Ning-style, EGM96 ground ref & \textbf{North Wildwood, NJ} & \textbf{1,078 parcels / 2,199 panos} & \textbf{+15.6 pp coverage, per-parcel envelope, +70 \% damage median} & \textbf{[x]} & \textbf{[x] (envelope)} & \textbf{Yes (NFHL$\times$USACE, UQ-propagated)} \\
\end{tabularx}

Li 2026 and this paper are close contemporaries and their contributions are complementary. Li\textquotesingle s tabular ML imputer answers for parcels that no GSV vintage can directly measure; we answer for parcels that \emph{some} GSV vintage can. On a head-to-head community, a plausible synthesis uses our multi-vintage rescue first (reducing the missed-parcel fraction, here from 70.1 \% to 54.5 \%), then Li-style imputation on the remainder.

\paragraph*{Study Area and Data}

\paragraph*{Study area --- North Wildwood, NJ}

North Wildwood is a small Atlantic barrier-island municipality on Five Mile Island (Cape May County, NJ). The entire municipal area falls within FEMA Special Flood Hazard Areas: 1,041 OSM buildings in zone AE, 29 in VE, 8 in zone X (Figure 1). Typical first-floor elevations are 2.5--6 m above local mean sea level; residential stock mixes slab-on-grade single-family homes, elevated beach cottages, and stilted / pile-supported buildings. Post-Sandy (2012) rebuilding left generational differences between building ages, a useful source of within-community variation.

\paragraph*{Data}

\begin{enumerate}
\def\labelenumi{\arabic{enumi}.}
\setcounter{enumi}{8}
\tightlist
\item
  \textbf{GSV panoramas + depthmaps + metadata.} All retrievable vintages 2013+ per parcel via the \texttt{gsv\_pano} library. Per-pano elevation is the decoded \texttt{elevation\_\allowbreak{}egm96\_\allowbreak{}m} field --- used as ground reference for that pano.
\item
  \textbf{Boundary.} OSM Nominatim polygon for North Wildwood.
\item
  \textbf{Buildings.} 1,078 OSM polygons clipped to the boundary.
\item
  \textbf{FEMA NFHL.} 30 flood-zone polygons + 65 boundary lines via the FEMA ArcGIS REST API.
\item
  \textbf{Depth-damage.} USACE EGM 04-01 generic residential curves (1-story, no basement, A / V zones).
\item
  \textbf{Detector.} Gao 2024 YOLOv5s checkpoint from the FFE\_Texas repository; used as-is.
\end{enumerate}

\paragraph*{Why North Wildwood}

Three criteria: (i) dense historical GSV (boundary-driven pilot: 59 \% of panos have $\geq$ 3 vintages; mean 3.08, max 8; see Figure 3); (ii) unambiguous flood-hazard exposure (every parcel inside an SFHA); (iii) mixed barrier-island housing stock. Nearby Sea Isle, NJ was considered and rejected for materially sparser GSV vintage coverage.

\begin{figure}
\centering
\includegraphics[width=\linewidth,alt={Figure 3 --- Multi-vintage GSV density on the 1,078 North Wildwood parcels. The pipeline\textquotesingle s parcel$\rightarrow$pano assignment retrieves 1--8 vintages per parcel (mean \textasciitilde3, median 3). Parcels with only one pano still feed C1\textquotesingle s latest-only baseline; parcels with $\geq$ 2 feed C2\textquotesingle s envelope.}]{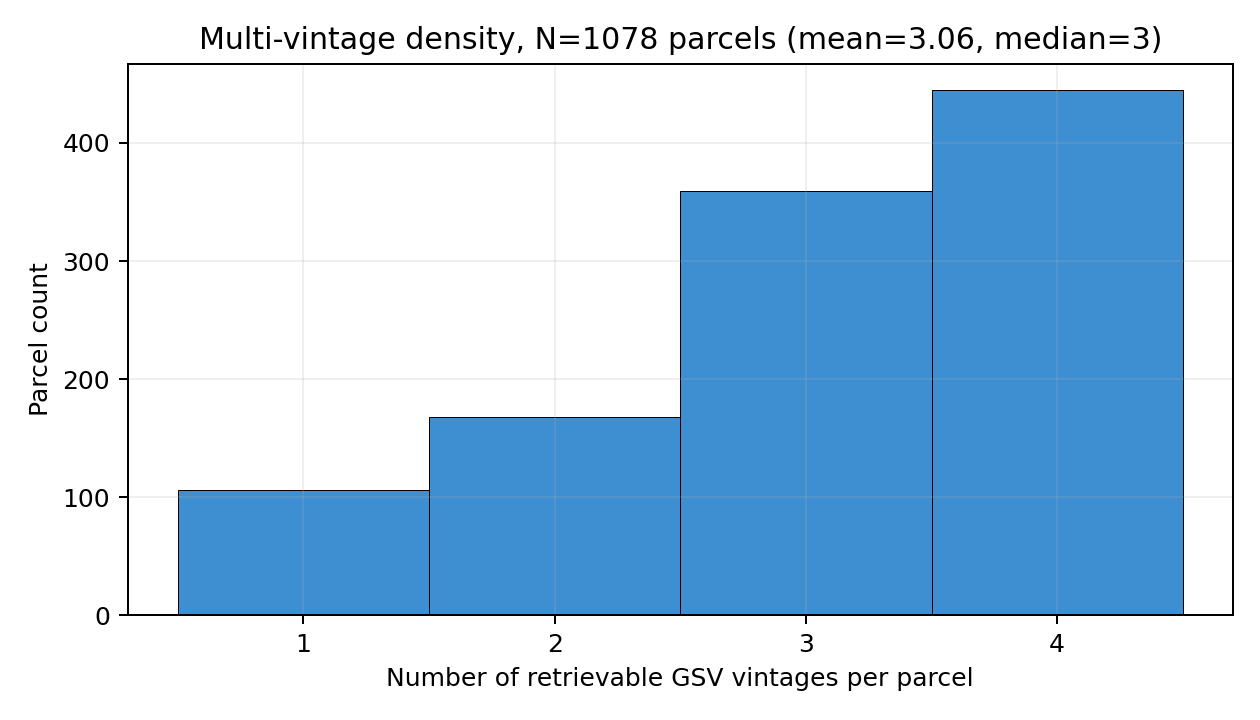}
\caption{Figure 3 --- Multi-vintage GSV density on the 1,078 North Wildwood parcels. The pipeline\textquotesingle s parcel$\rightarrow$pano assignment retrieves 1--8 vintages per parcel (mean \textasciitilde3, median 3). Parcels with only one pano still feed C1\textquotesingle s latest-only baseline; parcels with $\geq$ 2 feed C2\textquotesingle s envelope.}
\end{figure}

\paragraph*{Methods}

The pipeline is a four-stage chain (Figure 2): parcel-anchored multi-vintage assignment $\rightarrow$ fixed-detector door detection $\rightarrow$ depthmap-based FFE geometry $\rightarrow$ per-parcel aggregation and damage propagation.

\begin{figure}
\centering
\includegraphics[width=\linewidth,alt={Figure 2 --- End-to-end pipeline. Parcels (1,078 OSM centroids) drive a GSV Time-Machine query, dedup down to 2,199 unique panos, the fixed Gao-2024 YOLOv5 detector finds doors, the ELEV-VISION-family geometric step recovers per-pano FFE in EGM96 metres using each pano\textquotesingle s own recorded elevation as the ground reference, and per-parcel aggregates feed the three claims (C1 coverage, C2 envelope, C3 damage sensitivity).}]{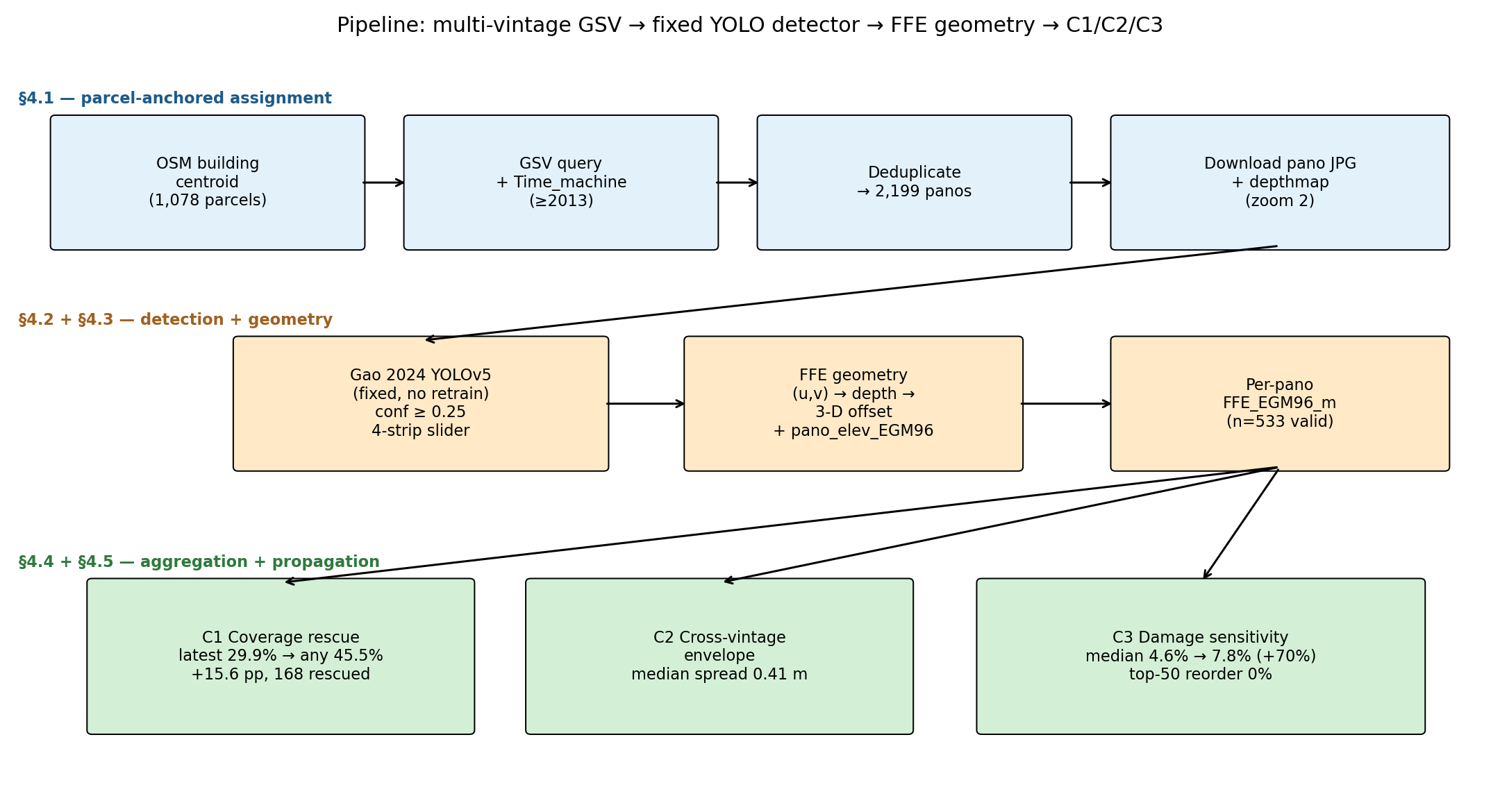}
\caption{Figure 2 --- End-to-end pipeline. Parcels (1,078 OSM centroids) drive a GSV Time-Machine query, dedup down to 2,199 unique panos, the fixed Gao-2024 YOLOv5 detector finds doors, the ELEV-VISION-family geometric step recovers per-pano FFE in EGM96 metres using each pano\textquotesingle s own recorded elevation as the ground reference, and per-parcel aggregates feed the three claims (C1 coverage, C2 envelope, C3 damage sensitivity).}
\end{figure}

\paragraph*{Parcel-anchored multi-vintage assignment}

For each OSM building centroid we query GSV via \texttt{GSV\_pano(request\_lat,\ request\_lon)}, returning the nearest current pano\textquotesingle s \texttt{panoId}. We walk the \texttt{Time\_machine} list and keep up to the three most recent historical \texttt{(panoId,\ vintage\_year)} pairs. We emit a \textbf{parcel $\rightarrow$ pano} map with 3,299 assignments (1,078 current + up to 3 historical per parcel). Global deduplication collapses repeated panoIds (adjacent parcels often share a nearest pano) to \textbf{2,199 unique panoramas}.

\paragraph*{Door detection (fixed detector)}

We use Gao et al. 2024\textquotesingle s YOLOv5s {[}5{]} as a fixed detector --- no retraining. The checkpoint labels four classes (door, house, step, garage); we keep only class 0 (door) at confidence $\geq$ 0.25. Each equirectangular panorama (zoom 2, 1024 $\times$ 512 px) is sliced into four overlapping horizontal strips; the highest-confidence door detection across strips is retained. On 2,199 panoramas the hit rate is 537 / 2,199 = 24.4 \%. Figure 9 shows nine representative detections spanning the FFE distribution; Figure 10 shows one parcel across four vintages.

As a zero-shot baseline we also run Grounding-DINO tiny {[}12{]} with the prompt \emph{"a door of a house. a front door of a house. an entrance door"} at confidence $\geq$ 0.20. The quantitative benchmark in \S{}5.1 justifies using Gao\textquotesingle s YOLOv5 as the paper\textquotesingle s instrument.

\begin{figure}
\centering
\includegraphics[width=\linewidth,alt={Figure 4 --- YOLO (left column, green) vs Grounding-DINO (right column, orange) on the same 13 panoramas. YOLO tightly frames actual doors at or near the door bottom; DINO frames whole building walls, storefronts, awnings, car fronts. Pink cross = inferred door-bottom pixel used by the FFE geometry.}]{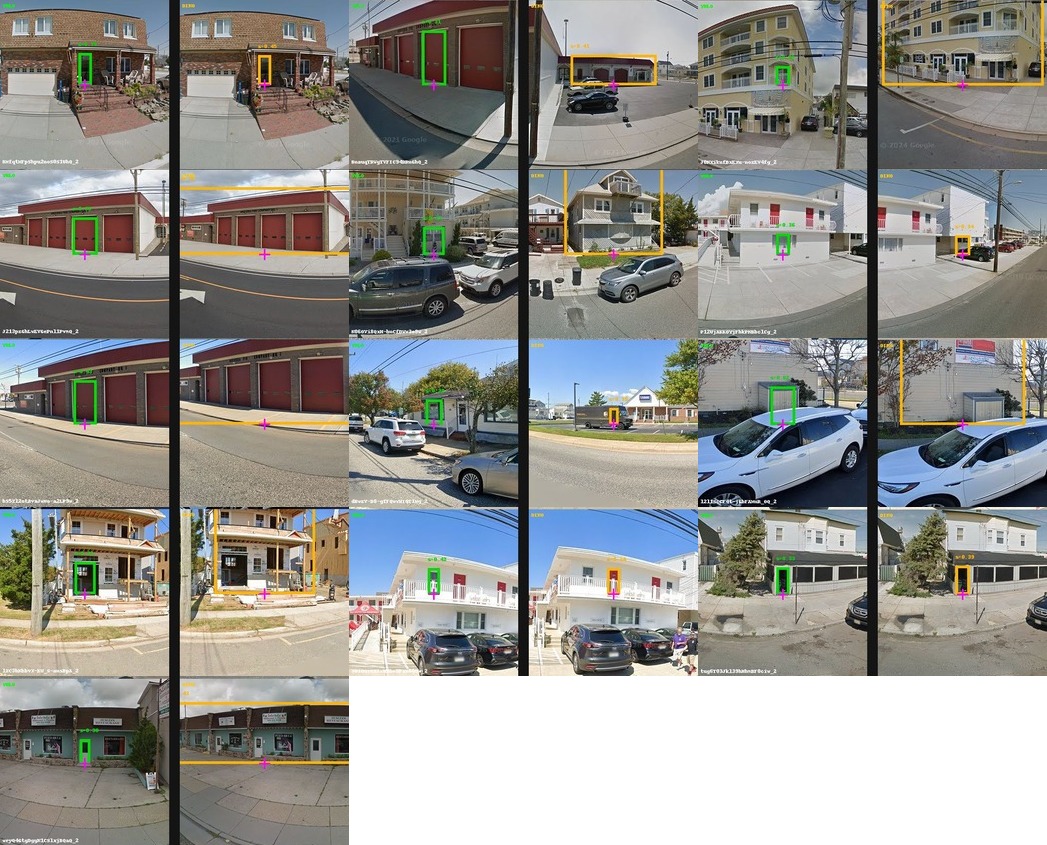}
\caption{Figure 4 --- YOLO (left column, green) vs Grounding-DINO (right column, orange) on the same 13 panoramas. YOLO tightly frames actual doors at or near the door bottom; DINO frames whole building walls, storefronts, awnings, car fronts. Pink cross = inferred door-bottom pixel used by the FFE geometry.}
\end{figure}

\paragraph*{FFE geometry}

For each detected door, the door-bottom pixel (u, v) is mapped to the raw depthmap frame (256 $\times$ 512), the camera-to-door range is read from the depthmap at that pixel, and panoramic spherical coordinates combine with yaw / tilt rotations (from the pano JSON metadata) to recover the world-frame 3D offset from camera to door-bottom. Letting z be the vertical component, the per-pano FFE in the EGM96 orthometric datum is

\begin{verbatim}
FFE_EGM96(pano) = pano.elevation_egm96_m + z.
\end{verbatim}

We reuse the point-cloud transform from the \texttt{gsv\_pano} library. 533 of 537 detections produce finite, plausible FFE (99.3 \%). Per-pano: mean 2.92 m, median 2.64 m, std 1.34 m, range 0.\allowbreak{}75--8.\allowbreak{}74 m.

\begin{figure}
\centering
\includegraphics[width=\linewidth,alt={Figure 9 --- Representative per-pano detections across the FFE distribution. Each tile shows a 440-px crop centred on the detected door; green box = YOLO bbox; pink cross = door-bottom pixel; annotation gives per-pano FFE (EGM96 m), the pano\textquotesingle s recorded elevation, and the detection confidence.}]{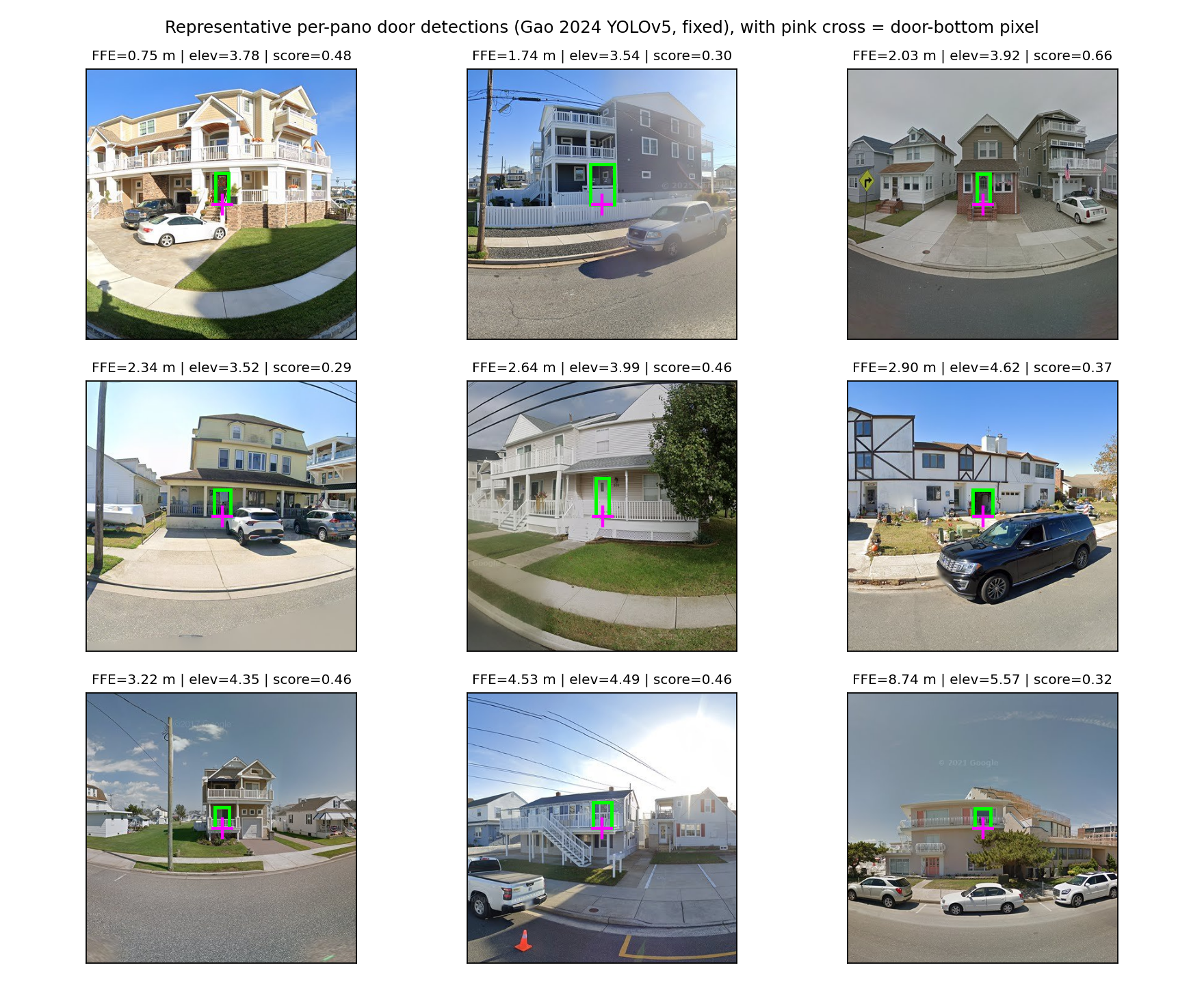}
\caption{Figure 9 --- Representative per-pano detections across the FFE distribution. Each tile shows a 440-px crop centred on the detected door; green box = YOLO bbox; pink cross = door-bottom pixel; annotation gives per-pano FFE (EGM96 m), the pano\textquotesingle s recorded elevation, and the detection confidence.}
\end{figure}

\begin{figure}
\centering
\includegraphics[width=\linewidth,alt={Figure 10 --- One North Wildwood parcel across four GSV vintages (2017, 2018, 2019, 2022). The Gao-2024 detector finds the same second-floor deck entry across vintages; the inter-vintage FFE range is 0.31 m --- a well-posed parcel whose envelope sits close to the ELEV-VISION geometric precision floor.}]{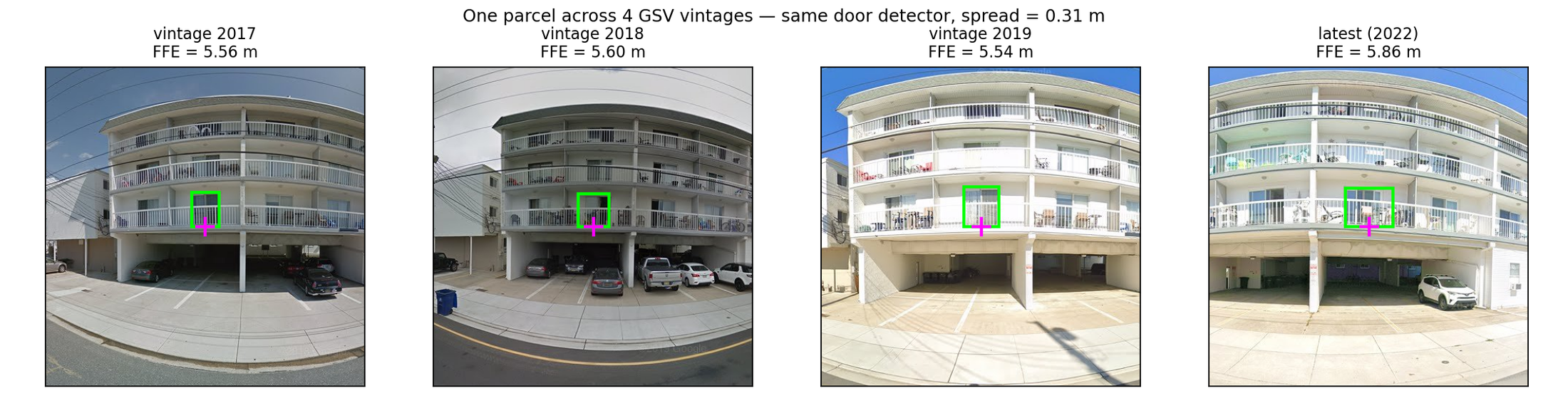}
\caption{Figure 10 --- One North Wildwood parcel across four GSV vintages (2017, 2018, 2019, 2022). The Gao-2024 detector finds the same second-floor deck entry across vintages; the inter-vintage FFE range is 0.31 m --- a well-posed parcel whose envelope sits close to the ELEV-VISION geometric precision floor.}
\end{figure}

\paragraph*{Per-parcel aggregation --- C1 (coverage) and C2 (envelope)}

Per parcel, coverage (C1): a parcel is "latest-only-covered" if its current-vintage pano yields a valid FFE, "any-vintage-covered" if any vintage does. Rescue = any $-$ latest. Cross-vintage envelope (C2): for parcels with $\geq$ 2 valid FFEs we record range (\texttt{rng}) and IQR, and a 90th-percentile half-width \texttt{k90\ =\ q\_\{0.90\}(rng/2)} as an in-sample dispersion statistic (we do \emph{not} claim it as a calibrated predictive interval --- that requires ground-truth FFE labels we do not have).

\paragraph*{Damage-sensitivity propagation --- C3}

Each parcel centroid spatial-joins to the FEMA NFHL zone polygon containing it, recovering (\texttt{FLD\_ZONE}, \texttt{STATIC\_BFE}). BFE is converted from NAVD88 feet to EGM96 metres with a constant $-$0.37 m offset for southern NJ. Flood depth above first floor: \texttt{BFE\_EGM96\ $-$\ FFE\_EGM96}, then converted to feet for USACE depth-damage interpolation (A-zone for AE/A, V-zone for VE/V). We compute two damage values per parcel: \textbf{deterministic} (per-parcel median FFE) and \textbf{probabilistic} (mean of damage at the envelope endpoints). Rank stability is assessed by top-50 priority reorder.

\paragraph*{Selection-bias / recoverability analysis (new in \S{}5.6)}

We model \texttt{P(any-vintage-covered)} as a function of observable parcel attributes (footprint area, distance to the nearest VE-zone polygon, zone membership, number of panos assigned) via logistic regression, and tabulate coverage by zone quintile and by footprint-area quintile. The aim is not a causal model but a systematic selection-bias audit: which parcels are systematically missed by the current pipeline?

\paragraph*{Results}

\paragraph*{Detector benchmark}

We benchmarked both detectors on Gao et al. (2024)\textquotesingle s public test split (\texttt{test.txt}, 121 images, 254 annotated doors) at IoU = 0.5. The split was held out during Gao\textquotesingle s training, so the YOLOv5 row is a legitimate held-out measurement. DINO is a true zero-shot evaluation on the same images.

\textbf{Table 2. Detector benchmark on 121 held-out labelled images (254 doors, IoU = 0.5).}

\begin{tabularx}{\linewidth}{@{}>{\hsize=2.9885\hsize\raggedright\arraybackslash}X>{\hsize=0.8847\hsize\raggedright\arraybackslash}X>{\hsize=0.6351\hsize\raggedright\arraybackslash}X>{\hsize=0.5703\hsize\raggedright\arraybackslash}X>{\hsize=0.4112\hsize\raggedright\arraybackslash}X>{\hsize=0.4112\hsize\raggedright\arraybackslash}X>{\hsize=0.4112\hsize\raggedright\arraybackslash}X>{\hsize=1.6878\hsize\raggedright\arraybackslash}X@{}}
\toprule\noalign{}
Detector & Precision & Recall & F1 & TP & FP & FN & Detections / image \\
\midrule\noalign{}
\endhead
\bottomrule\noalign{}
\endlastfoot
Gao 2024 YOLOv5s (fixed) & 0.545 & 0.547 & \textbf{0.546} & 139 & 116 & 115 & 2.1 \\
Grounding-DINO tiny (zero-shot) & 0.104 & 0.342 & 0.160 & 87 & 750 & 167 & 6.9 \\
\end{tabularx}

Zero-shot is \textbf{5$\times$ less precise} and \textbf{3.4$\times$ worse in F1}. 90 \% of DINO boxes are false positives (750 / 837). Figure 4 confirms the gap visually.

\paragraph*{Coverage rescue (C1) --- the main result}

\textbf{Latest-only coverage: 29.9 \%. Any-vintage coverage: 45.5 \%. Gain: +15.6 pp. Rescued: 168 parcels} (above the pre-registered $\geq$ 12 pp threshold). Figure 5 breaks the 1,078 parcels into 322 latest-only-covered, 168 rescued-by-historical, and 588 still un-measured across all retrievable vintages. The rescued parcels are textbook transient-obstacle cases: a van parked in 2022 but not 2019; a hedge untrimmed in 2020 but absent in 2016; an unfavourable camera heading in 2017 with a head-on 2013 capture. Multi-vintage converts each obstacle class into an independent trial.

\begin{figure}
\centering
\includegraphics[width=\linewidth,alt={Figure 5 --- Coverage rescue bar chart. On the 1,078 OSM parcels of North Wildwood, 322 (29.9 \%) are covered by the latest GSV capture alone; an additional 168 (15.6 \%) are rescued by a historical vintage; 588 (54.5 \%) remain without a direct FFE across all retrievable vintages (imputation territory).}]{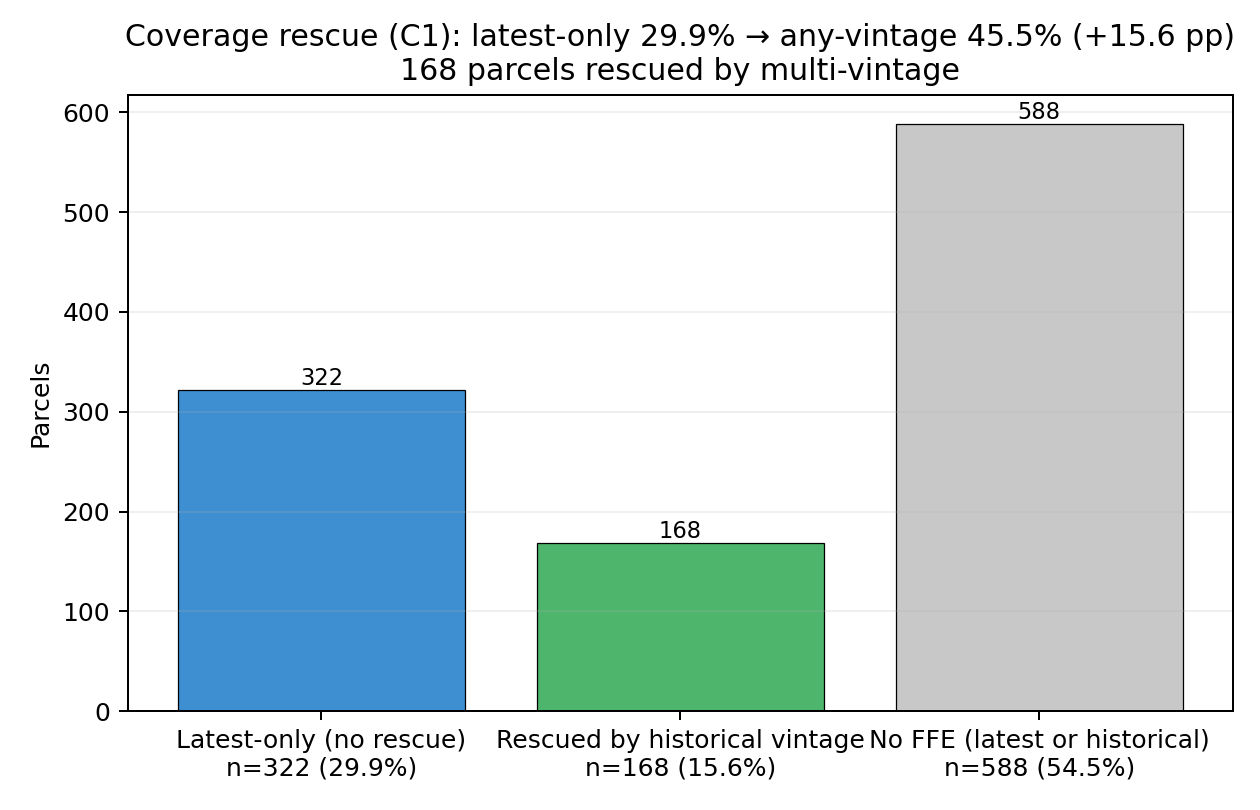}
\caption{Figure 5 --- Coverage rescue bar chart. On the 1,078 OSM parcels of North Wildwood, 322 (29.9 \%) are covered by the latest GSV capture alone; an additional 168 (15.6 \%) are rescued by a historical vintage; 588 (54.5 \%) remain without a direct FFE across all retrievable vintages (imputation territory).}
\end{figure}

Figure 11 makes the same result spatial: rescued parcels (green) are distributed through the AE-zone core rather than concentrated in any single block, indicating that transient-obstacle failures are a pipeline-wide phenomenon, not a neighbourhood effect.

\begin{figure}
\centering
\includegraphics[width=\linewidth,alt={Figure 11 --- Spatial view of per-parcel coverage status across North Wildwood. Blue = latest-only covered, green = rescued by a historical vintage, grey = no direct FFE under any vintage. The 168 green parcels are distributed across the AE-zone core, consistent with transient-obstacle failures rather than neighbourhood-scale problems.}]{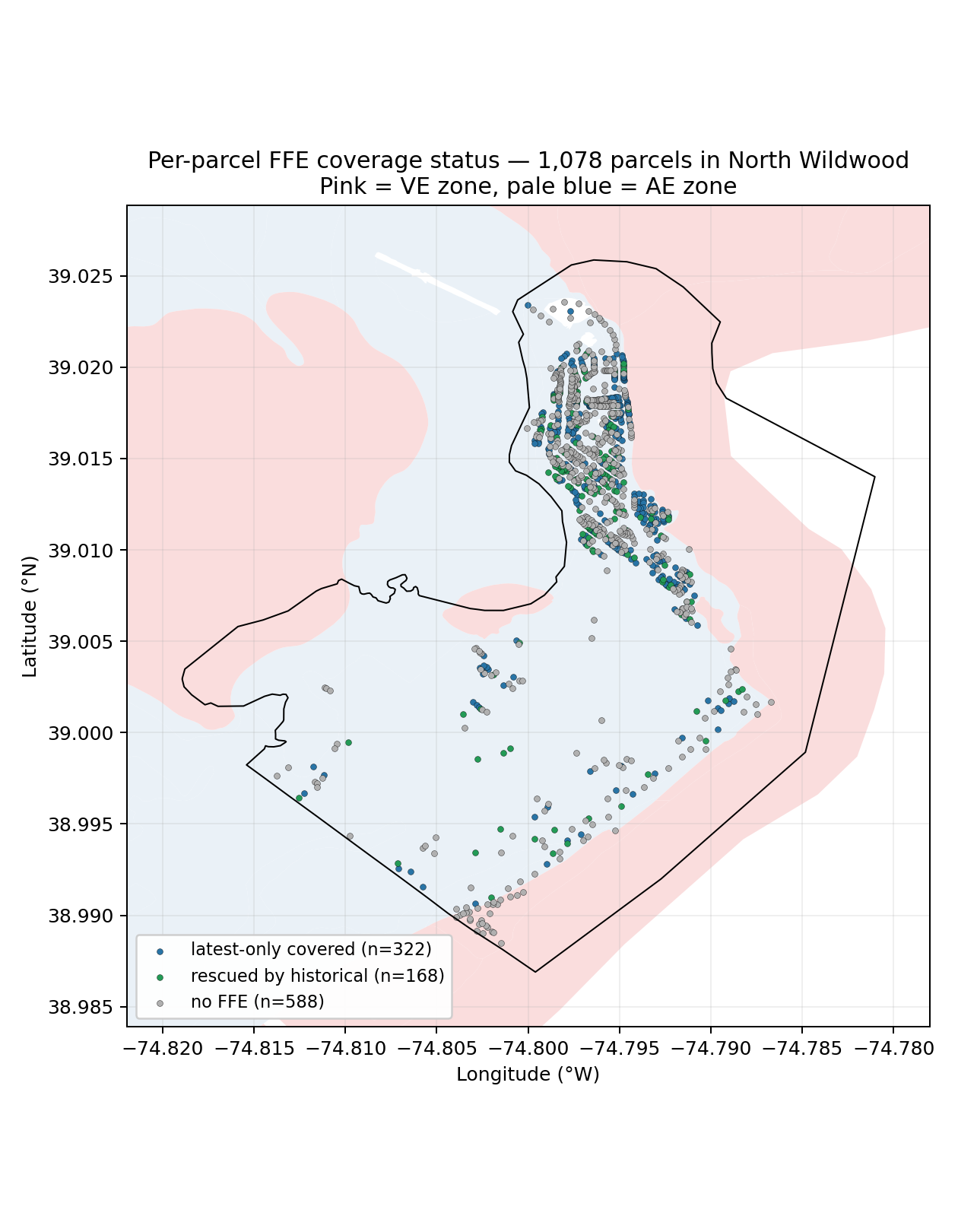}
\caption{Figure 11 --- Spatial view of per-parcel coverage status across North Wildwood. Blue = latest-only covered, green = rescued by a historical vintage, grey = no direct FFE under any vintage. The 168 green parcels are distributed across the AE-zone core, consistent with transient-obstacle failures rather than neighbourhood-scale problems.}
\end{figure}

\paragraph*{Cross-vintage envelopes (C2) --- a partial signal}

On 240 parcels with $\geq$ 2 valid vintages: median cross-vintage spread = 0.412 m, median IQR = 0.206 m, k90 half-width = 1.187 m. The bulk is tight (median spread $\approx$ 0.4 m corresponds to a half-width of 0.2 m, well within the ELEV-VISION geometric precision floor and below FEMA\textquotesingle s 0.3 m elevation-certificate tolerance). The k90 half-width of 1.19 m exceeds our pre-registered $\leq$ 0.45 m threshold and we do \emph{not} claim calibrated per-parcel predictive intervals. Instead we report the observed dispersion as an \emph{uncertainty envelope}, and in \S{}6.2 we explain why the tail is detector confusion (a different facade element gets detected in different vintages) rather than FFE uncertainty.

\begin{figure}
\centering
\includegraphics[width=\linewidth,alt={Figure 6 --- Cross-vintage uncertainty envelope distribution on the 240-parcel C2 cohort. (a) Histogram of cross-vintage spread range; the vertical red line marks twice the pre-registered 0.45 m half-width threshold. The bulk sits comfortably below threshold; a long right tail extends past 1.5 m. (b) Spread vs per-parcel median FFE scatter, coloured by the number of valid vintages contributing.}]{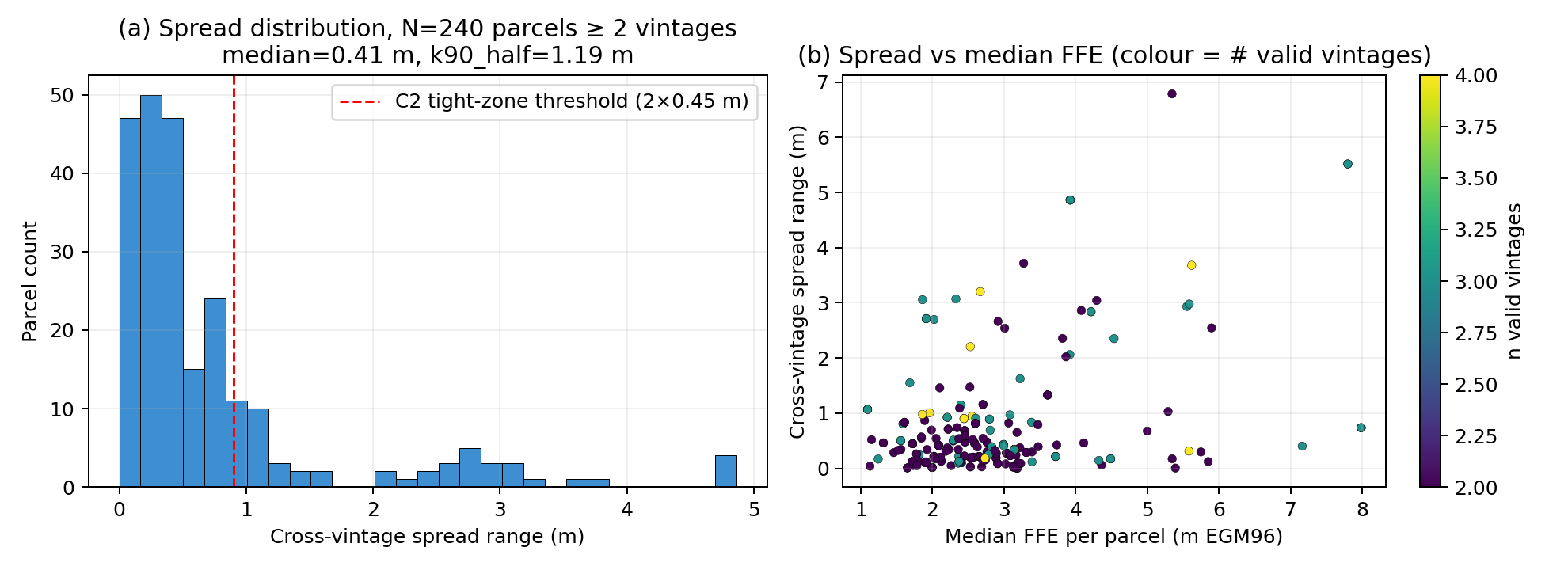}
\caption{Figure 6 --- Cross-vintage uncertainty envelope distribution on the 240-parcel C2 cohort. (a) Histogram of cross-vintage spread range; the vertical red line marks twice the pre-registered 0.45 m half-width threshold. The bulk sits comfortably below threshold; a long right tail extends past 1.5 m. (b) Spread vs per-parcel median FFE scatter, coloured by the number of valid vintages contributing.}
\end{figure}

\paragraph*{Damage-sensitivity under UQ (C3) --- magnitude, not priority}

Of 490 covered parcels, 213 (43.5 \%) are below BFE deterministically. Median deterministic damage = 4.61 \%; median UQ-propagated damage = 7.83 \%. \textbf{Median per-house uplift = +69.8 \%}; aggregate community-damage uplift = 3.6 \%; top-50 priority reorder = 0 \%. The near-BFE cohort (215 parcels within $\pm$0.5 m) has 0 \% classification flips between deterministic and probabilistic below-BFE labels.

\textbf{Interpretation: deterministic outputs correctly identify \emph{which} parcels are at greatest risk but systematically under-report \emph{how much} risk each house carries}, by a factor of roughly 1.7$\times$ at the median. The planner who prioritises elevation grants gets the same list. The financial analyst pricing a single-house retrofit or insurance premium under-estimates the expected loss by roughly 70 \% at the typical house.

\begin{figure}
\centering
\includegraphics[width=\linewidth,alt={Figure 12 --- Spatial per-parcel damage under (a) deterministic vs (b) UQ-propagated FFE, on the 490 covered parcels. Both panels share the same colour scale (0--60 \% damage). Visual comparison shows that the ranking of the highest-risk parcels is stable between panels (supporting the zero-reorder finding), while per-house magnitudes --- visible as deeper red --- shift uniformly upwards in the right panel.}]{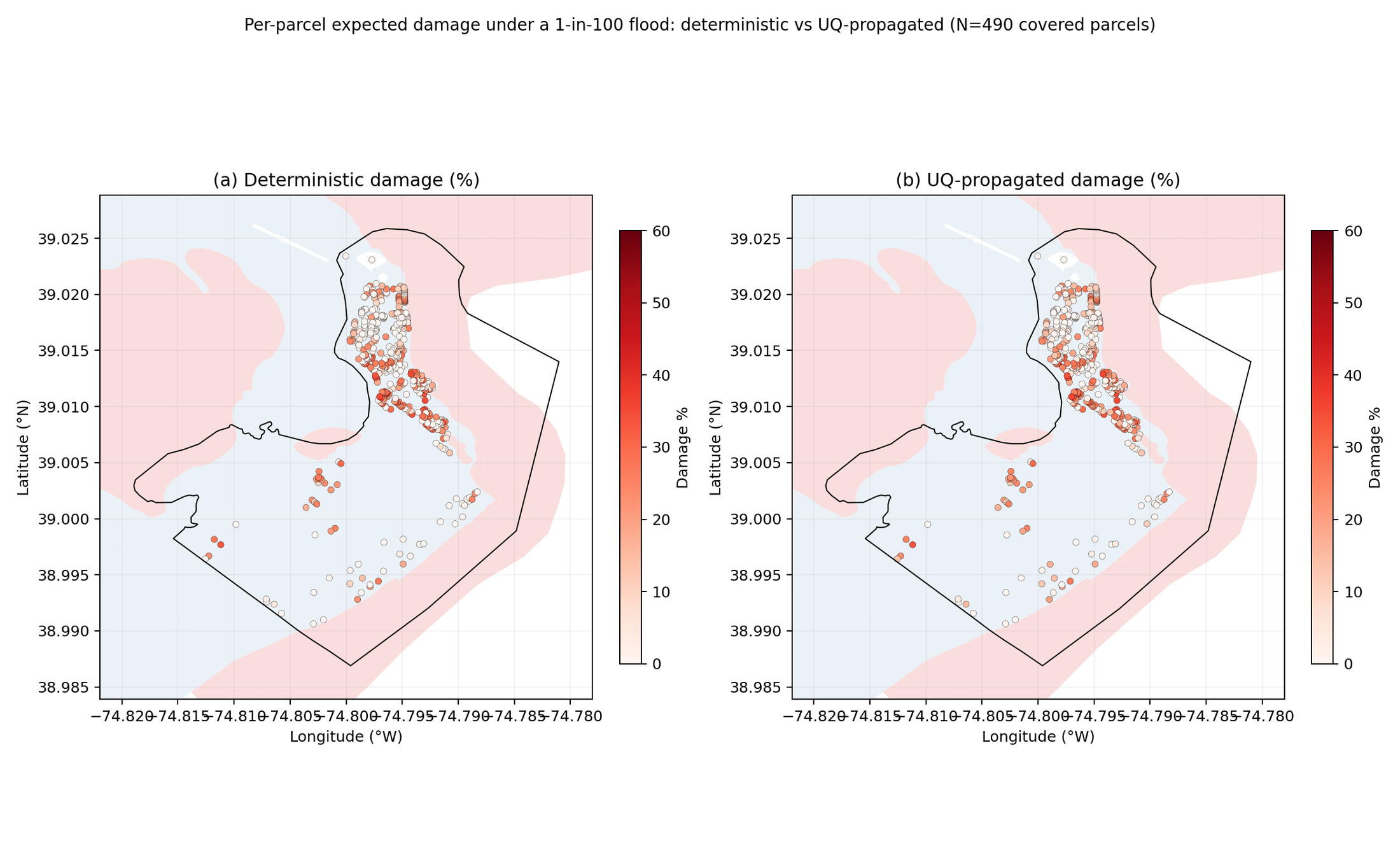}
\caption{Figure 7 --- Per-parcel damage under deterministic vs UQ-propagated FFE (N = 490). (a) Scatter: points systematically fall above the identity line, confirming per-house damage uplift. (b) Histogram of the per-parcel uplift percentage (non-zero deterministic subset); median uplift = 69.8 \%.}
\end{figure}

Figure 12 --- Spatial per-parcel damage under (a) deterministic vs (b) UQ-propagated FFE, on the 490 covered parcels. Both panels share the same colour scale (0--60 \% damage). Visual comparison shows that the ranking of the highest-risk parcels is stable between panels (supporting the zero-reorder finding), while per-house magnitudes --- visible as deeper red --- shift uniformly upwards in the right panel.

\paragraph*{Zone and subgroup distributions}

\begin{figure}
\centering
\includegraphics[width=\linewidth,alt={Figure 8 --- FFE (left) and deterministic damage (right) stratified by FEMA zone among parcels with a valid FFE. AE-zone parcels hold the full damage distribution; X-zone parcels sit safely above BFE with near-zero expected damage. V-zone parcels are absent; see \S{}5.6.}]{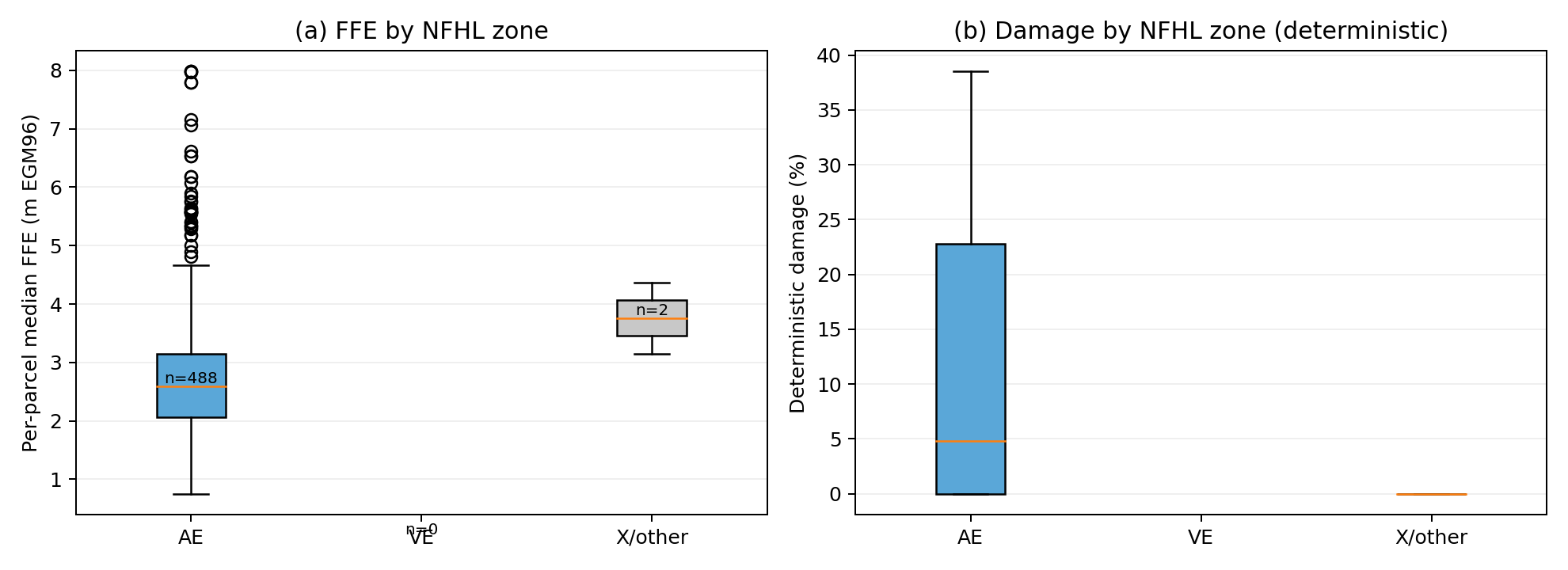}
\caption{Figure 8 stratifies FFE and deterministic damage by FEMA zone among covered parcels. AE-zone covered parcels hold most of the distribution; X-zone parcels sit above BFE and contribute near-zero damage. V-zone parcels do not appear --- for the reason explained in \S{}5.6.}
\end{figure}

Figure 8 --- FFE (left) and deterministic damage (right) stratified by FEMA zone among parcels with a valid FFE. AE-zone parcels hold the full damage distribution; X-zone parcels sit safely above BFE with near-zero expected damage. V-zone parcels are absent; see \S{}5.6.

\paragraph*{Selection-bias / recoverability analysis --- who do we miss?}

Coverage is not uniform across the parcel population. Marginal tables by FEMA zone and by footprint-area / distance-to-coast quintiles:

\begin{tabularx}{\linewidth}{@{}>{\hsize=0.9165\hsize\raggedright\arraybackslash}X>{\hsize=0.7230\hsize\raggedright\arraybackslash}X>{\hsize=1.1930\hsize\raggedright\arraybackslash}X>{\hsize=1.2705\hsize\raggedright\arraybackslash}X>{\hsize=0.8970\hsize\raggedright\arraybackslash}X@{}}
\toprule\noalign{}
FEMA zone & N parcels & Latest-only cov. & Any-vintage cov. & Rescue (pp) \\
\midrule\noalign{}
\endhead
\bottomrule\noalign{}
\endlastfoot
AE & 1,041 & 30.7 \% & \textbf{46.9 \%} & 16.2 \\
VE & 29 & \textbf{0.0 \%} & \textbf{0.0 \%} & 0.0 \\
X & 8 & 25.0 \% & 25.0 \% & 0.0 \\
\end{tabularx}

\begin{tabularx}{\linewidth}{@{}>{\hsize=1.5620\hsize\raggedright\arraybackslash}X>{\hsize=1.0250\hsize\raggedright\arraybackslash}X>{\hsize=0.7680\hsize\raggedright\arraybackslash}X>{\hsize=0.8375\hsize\raggedright\arraybackslash}X>{\hsize=0.8075\hsize\raggedright\arraybackslash}X@{}}
\toprule\noalign{}
Footprint area (quintile) & Mean area (m$^{2}$) & Latest-only & Any-vintage & Rescue (pp) \\
\midrule\noalign{}
\endhead
\bottomrule\noalign{}
\endlastfoot
Q1 (smallest) & 15 & 31.5 \% & 43.1 \% & 11.6 \\
Q2 & 74 & 31.2 \% & 48.4 \% & 17.2 \\
Q3 & 108 & 30.1 \% & 46.3 \% & 16.2 \\
Q4 & 166 & 28.4 \% & 46.5 \% & 18.1 \\
Q5 (largest) & 591 & 28.2 \% & 43.1 \% & 14.9 \\
\end{tabularx}

\begin{tabularx}{\linewidth}{@{}>{\hsize=2.6475\hsize\raggedright\arraybackslash}X>{\hsize=0.5150\hsize\raggedright\arraybackslash}X>{\hsize=0.5845\hsize\raggedright\arraybackslash}X>{\hsize=0.6380\hsize\raggedright\arraybackslash}X>{\hsize=0.6150\hsize\raggedright\arraybackslash}X@{}}
\toprule\noalign{}
Distance-to-VE (quintile, proxy for distance-to-ocean) & Mean (m) & Latest-only & Any-vintage & Rescue (pp) \\
\midrule\noalign{}
\endhead
\bottomrule\noalign{}
\endlastfoot
Q1 (near ocean) & 34 & 37.5 \% & 49.1 \% & 11.6 \\
Q2 & 100 & 34.0 \% & 47.9 \% & 13.9 \\
Q3 & 176 & 26.4 \% & 41.7 \% & 15.3 \\
Q4 & 264 & 27.4 \% & 40.0 \% & 12.6 \\
Q5 (inland) & 375 & 24.1 \% & \textbf{48.6 \%} & \textbf{24.5} \\
\end{tabularx}

A logistic regression of \texttt{P(any-vintage-covered)} on \texttt{log(area)}, \texttt{log(distance-to-VE)}, \texttt{is\_VE}, \texttt{n\_\allowbreak{}panos\_\allowbreak{}assigned} finds one dominant coefficient: \texttt{is\_VE} \textbf{reduces the odds of any-vintage coverage by about 14$\times$} (odds-ratio 0.072). \texttt{n\_\allowbreak{}panos\_\allowbreak{}assigned} unsurprisingly helps (OR 1.294 per pano); \texttt{log(area)} and \texttt{log(distance-to-VE)} are weakly negative (base rate 45.5 \%, model training accuracy 55 \%).

The headline result of \S{}5.6 is that \textbf{the pipeline completely fails on FEMA V-zone parcels} (0 / 29). These are the ocean-front stilted / pile-supported houses where the standardised door-on-slab prior assumed by the ELEV-VISION geometric recipe breaks: the detector often finds a door and the geometry produces a z-offset, but the estimate is the elevation of an above-pile entry deck rather than the first floor, and our internal quality filter rejects these as "depth \textgreater{} 40 m" or "depth mask zero." In footprint-area terms, very-small (Q1) and very-large (Q5) parcels are somewhat harder. In distance-to-ocean terms, latest-only coverage is stronger near the coast but rescue magnitude is largest farthest inland.

\textbf{Implication.} Our "municipal-scale" is better described as \textbf{municipal-scale screening under partial observability, heavily biased against V-zone parcels}. The V-zone gap is an instrument limitation of the fixed door-on-slab detector, not of multi-vintage GSV itself; an upgrade path (stilted-foundation-aware detection or geometry) is noted in \S{}7.

\paragraph*{Summary of claims against pre-registered thresholds}

\paragraph*{Table 3. Per-claim results against the pre-registered thresholds from the experiment plan.}

\begin{tabularx}{\linewidth}{@{}>{\hsize=0.6520\hsize\raggedright\arraybackslash}X>{\hsize=1.0096\hsize\raggedright\arraybackslash}X>{\hsize=1.1560\hsize\raggedright\arraybackslash}X>{\hsize=1.1824\hsize\raggedright\arraybackslash}X@{}}
\toprule\noalign{}
Claim & Threshold & Observed & Verdict \\
\midrule\noalign{}
\endhead
\bottomrule\noalign{}
\endlastfoot
\textbf{C1 Coverage rescue} & $\geq$ 12 pp coverage gain & +15.6 pp (29.9 \% $\rightarrow$ 45.5 \%) & \textbf{[x] PASS} \\
\textbf{C2 Cross-vintage envelope} & k90 half-width $\leq$ 0.45 m & median spread 0.41 m (tight bulk); k90 half-width 1.19 m (long tail) & [!] Partial --- bulk passes, tail fails \\
\textbf{C3 Damage propagation} & Top-50 reorder $\geq$ 10 \% OR Brier improvement $\geq$ 0.03 & Top-50 reorder = 0 \%; \textbf{median per-house damage uplift = +69.8 \%} & [!] Reframed --- fails on reorder, supports magnitude sensitivity \\
\textbf{Detector ablation} & (not pre-registered) & Gao YOLOv5 F1 = 0.55 vs DINO F1 = 0.16 & [x] Supports fixed detector choice \\
\textbf{Selection bias} & (not pre-registered) & V-zone coverage = 0 / 29 parcels & [*] Critical pipeline limit --- 29 highest-hazard parcels are a blind spot \\
\end{tabularx}

\paragraph*{Discussion}

\paragraph*{Why multi-vintage works for coverage (C1)}

Most single-vintage failure modes are \textbf{transient}: parked vehicles, seasonal vegetation, unfavourable camera heading. Multi-vintage GSV turns each of these into an independent trial. The mechanism is orthogonal to any detector improvement and orthogonal to tabular imputation of missed parcels (Li 2026 {[}6{]}) --- it attacks parcels that \emph{have} a viewable door, just not in the latest pano. Imputation still helps for the residual 54.5 \% of parcels that no retrievable vintage resolves, and for the V-zone parcels documented in \S{}5.6.

\paragraph*{Why the C2 tail is noisy}

Cross-vintage disagreement is epistemic in principle (multiple independent estimates) but operational in practice (different detections across vintages). For well-posed parcels (consistent door view --- Figure 10) the detections converge and disagreement collapses to the pipeline\textquotesingle s geometric precision floor (\textasciitilde0.2 m). For ill-posed parcels, the detector latches onto a \emph{different} facade element each year: a garage in 2016, a storefront in 2019, a window in 2022. The resulting spread is detector confusion, not FFE uncertainty. We document this explicitly rather than smoothing it into a single calibrated interval. Two mitigations compatible with the no-retraining constraint: (i) SAM-based refinement of the YOLO door bbox to its lower edge; (ii) a cross-vintage consistency filter that discards parcels where the detector\textquotesingle s \emph{class label} (door vs garage vs step) varies across vintages.

\paragraph*{Why C3 does not reorder but magnitude still matters}

The worst parcels are deep below BFE ($\geq$ 1.5 m flood depth) and sit on the plateau of the depth-damage curve where further error barely moves the ranking. Yet the median per-house damage rises by \textasciitilde70 \% when UQ is included (Figures 7, 12). This split matters for policy. A planner prioritising \emph{which} houses to offer elevation grants first gets the same list either way. A financial analyst pricing a single-house retrofit or insurance premium under-estimates by roughly 1.7$\times$ under a deterministic pipeline. Decision-support products should carry the UQ channel end-to-end not because it reorders the list but because it shifts the dollar number.

\paragraph*{Comparison with concurrent work (Li 2026)}

Li 2026 {[}6{]} scaled ELEV-VISION-SAM across 18 Texas AOIs, used RF/GBM imputation for parcels missed by direct extraction, and integrated Fathom-1 inundation with USACE depth-damage. Our work isolates the value of \textbf{temporal redundancy under a fixed detector} and adds a per-parcel uncertainty channel; it does not claim best-in-class absolute FFE accuracy. A synthesis of the two approaches is a natural next paper: multi-vintage rescue first (reducing missed fraction), then tabular imputation on the residual, with per-parcel UQ carried through damage.

\paragraph*{Generalisability and V-zone blind spot}

North Wildwood\textquotesingle s GSV density and mixed stock make it a fair barrier-island test, but not universal. We expect similar C1 gains on Stone Harbor, Cape May City, and the wider Wildwoods, and lower gains in GSV-sparse rural coastal communities. Crucially, \S{}5.6\textquotesingle s V-zone gap generalises: wherever stilted / elevated housing sits on open beachfront (V-zone), the fixed door-on-slab detector will fail. A stilted-foundation-aware variant (new detector class "pile structure," or a post-hoc deck-vs-slab classifier) is the most important near-term extension.

\paragraph*{Limitations and Future Work}

\begin{itemize}
\tightlist
\item
  \textbf{No lidar ground-truth FFE labels.} All accuracy-style numbers in this paper are either (a) detector F1 on labelled doors --- not FFE validation --- or (b) in-sample dispersion statistics. The immediate next step is a stratified human-reference audit of \textasciitilde100 covered parcels to compute bias / precision / empirical coverage against rater consensus, then acquisition of NOAA / USGS post-Sandy lidar.
\item
  \textbf{V-zone selection bias.} Zero coverage on 29 V-zone parcels (\S{}5.6). The next paper should demonstrate a stilted-foundation-aware detector variant.
\item
  \textbf{OSM parcel cohort.} Our 1,078 OSM building polygons undercount the NJGIN cadastral roll by \textasciitilde1.5--2$\times$ and omit some outbuildings / ADUs. Re-running on the cadastral parcel set when acquired is a mechanical extension.
\item
  \textbf{Nearest-pano assignment is view-agnostic.} A facade-visibility filter (camera heading pointing toward the building) would tighten the C1 cohort.
\item
  \textbf{NAVD88 $\rightarrow$ EGM96 uses a constant offset.} True per-point offset varies $\pm$0.05 m over Cape May County; below the per-parcel spread we report.
\item
  \textbf{Damage curves uniformly 1-story no-basement.} OSM \texttt{building:levels} + NFHL zone stratification is an easy refinement.
\item
  \textbf{Detector fixed, no retraining.} F1 = 0.55 on the held-out split is the measurement floor. Newer zero-shot detectors (OWLv2, YOLO-World, Grounding-DINO 1.5, T-Rex2, DINO-X) remain to be benchmarked as ablation rows.
\item
  \textbf{Single-community evaluation.} Stone Harbor (Seven Mile Island) or Cape May City external-validity was planned but deferred.
\end{itemize}

\paragraph*{Conclusion}

\textbf{The headline contribution of this paper is that multi-vintage GSV rescues 15.6 percentage points of FFE coverage} --- 168 of 1,078 parcels in North Wildwood, NJ --- above the single-vintage baseline, using the same geometric pipeline with a fixed off-the-shelf detector. Two secondary findings characterise the downstream consequences: cross-vintage disagreement yields a useful uncertainty envelope for the bulk of covered parcels (though not a fully calibrated predictive interval), and propagating that envelope through FEMA NFHL $\times$ USACE depth-damage reveals that deterministic pipelines under-report per-house damage by roughly 70 \% at the median, even though they correctly rank which houses are at greatest risk. A critical selection-bias finding --- the pipeline achieves zero coverage on FEMA V-zone stilted / pile-supported houses --- bounds what "municipal-scale" means under the present detector and frames the most important next step. For residential flood-risk analytics on US Atlantic barrier islands, the multi-vintage signal is a cheap, reproducible improvement; the V-zone gap is the first obstacle to close in a follow-up.

\paragraph*{Code and Data Availability}

Pipeline code, figures, and per-parcel outputs are archived in the accompanying repository. The detector is Gao et al. 2024\textquotesingle s YOLOv5s checkpoint from \texttt{FFE\_\allowbreak{}Texas/\allowbreak{}runs/\allowbreak{}exp4/\allowbreak{}weights/\allowbreak{}best.\allowbreak{}pt} (MIT, via \texttt{EvilicLufas/\allowbreak{}FFE\_\allowbreak{}Texas} on GitHub) --- used without retraining. GSV panoramic imagery is redistributed only as referenced \texttt{panoId} values (subject to Google\textquotesingle s ToS); derived per-panorama FFE estimates, metadata, and aggregates are released under CC-BY. FEMA NFHL and OSM data are public-domain and ODbL respectively.

\paragraph*{References}

Ning, H., Li, Z., Ye, X., Wang, S., Wang, W., Huang, X. (2022). Exploring the vertical dimension of street view image based on deep learning: a case study on lowest floor elevation estimation. \emph{Int. J. Geogr. Inf. Sci.} 36 (7), 1317--1342. https://doi.\allowbreak{}org/\allowbreak{}10.\allowbreak{}1080/\allowbreak{}13658816.\allowbreak{}2021.\allowbreak{}1981334

Ho, Y.-H., Lee, C.-C., Diaz, N. D., Brody, S. D., Mostafavi, A. (2024). ELEV-VISION: Automated Lowest Floor Elevation Estimation from Segmenting Street View Images. \emph{ACM Journal on Computing and Sustainable Societies}. arXiv:2306.03050.

Ho, Y.-H., Li, L., Mostafavi, A. (2025). ELEV-VISION-SAM: Integrated Vision Language and Foundation Model for Automated Estimation of Building Lowest Floor Elevation. \emph{Computer-Aided Civil and Infrastructure Engineering}. arXiv:2404.12606.

Sorboni, N. G., Wang, J., et al. (2024). Automated first floor height estimation for flood vulnerability analysis using deep learning and Google Street View. \emph{Journal of Flood Risk Management}. https://doi.\allowbreak{}org/\allowbreak{}10.\allowbreak{}1111/\allowbreak{}jfr3.\allowbreak{}12975

Gao, G., Ye, X., Li, S., Huang, X., Ning, H., Retchless, D., Li, Z. (2024). Exploring flood mitigation governance by estimating first-floor elevation via deep learning and Google Street View in coastal Texas. \emph{Environment and Planning B}. https://doi.\allowbreak{}org/\allowbreak{}10.\allowbreak{}1177/\allowbreak{}23998083231175681

Li, X., Ho, Y.-H., Brody, S. D., Mostafavi, A. (2026). Property-Level Flood Risk Assessment Using AI-Enabled Street-View Lowest Floor Elevation Extraction and ML Imputation Across Texas. arXiv:2604.01153.

Zarekarizi, M., Srikrishnan, V., Keller, K. (2020). Neglecting Uncertainties Biases House-Elevation Decisions to Manage Riverine Flood Risks. arXiv:2001.06457.

Reid, et al. (2025). Efficient First Floor Height Estimation of Buildings from Sequential Images and Airborne LiDAR for Flood Risk Analysis. \emph{Can. J. Remote Sensing}. https://doi.\allowbreak{}org/\allowbreak{}10.\allowbreak{}1080/\allowbreak{}07038992.\allowbreak{}2025.\allowbreak{}2581695

Fan, Z., Feng, C.-C., Biljecki, F. (2024). Coverage and Bias of Street View Imagery in Mapping the Urban Environment. arXiv:2409.15386.

Rasmussen, D. J., Buchanan, M. K., Kopp, R. E., Oppenheimer, M. (2019). A flood damage allowance framework for coastal protection with deep uncertainty in sea-level rise. arXiv:1908.02844.

Bodoque, J. M., et al. (2016). Uncertainty propagation in lidar-based first-floor elevation for flood damage assessment. \emph{Remote Sensing} 8, 604. https://doi.\allowbreak{}org/\allowbreak{}10.\allowbreak{}3390/\allowbreak{}rs8070604

Liu, S., Zeng, Z., Ren, T., et al. (2023). Grounding DINO: Marrying DINO with Grounded Pre-Training for Open-Set Object Detection. arXiv:2303.05499.

Jocher, G., et al. (2022). YOLOv5 ultralytics. https://github.\allowbreak{}com/\allowbreak{}ultralytics/\allowbreak{}yolov5

U.S. Army Corps of Engineers (2003). Generic Depth-Damage Relationships for Residential Structures with Basements. Economic Guidance Memorandum 04-01.

Federal Emergency Management Agency (2024). National Flood Hazard Layer (NFHL) Technical Reference.

OpenStreetMap contributors (2026). \emph{OpenStreetMap}. https://www.\allowbreak{}openstreetmap.\allowbreak{}org

Yang, L., Kang, B., Huang, Z., Zhao, Z. (2024). Depth Anything V2. arXiv:2406.09414.

\paragraph*{Appendix A. Complete Figure List}

\begin{tabularx}{\linewidth}{@{}>{\hsize=0.1056\hsize\raggedright\arraybackslash}X>{\hsize=1.2947\hsize\raggedright\arraybackslash}X>{\hsize=1.4587\hsize\raggedright\arraybackslash}X>{\hsize=1.1411\hsize\raggedright\arraybackslash}X@{}}
\toprule\noalign{}
\# & Figure & File & Source \\
\midrule\noalign{}
\endhead
\bottomrule\noalign{}
\endlastfoot
1 & Study area + NFHL zones + 1,078 parcels & \texttt{output/\allowbreak{}figures/\allowbreak{}fig1\_\allowbreak{}study\_\allowbreak{}area.\allowbreak{}pdf} & OSM, NFHL, building polygons \\
2 & End-to-end pipeline block diagram & \texttt{output/\allowbreak{}figures/\allowbreak{}fig2\_\allowbreak{}pipeline.\allowbreak{}pdf} & hand-authored schematic \\
3 & Multi-vintage density histogram & \texttt{output/\allowbreak{}figures/\allowbreak{}fig3\_\allowbreak{}vintage\_\allowbreak{}hist.\allowbreak{}pdf} & \texttt{parcel\_\allowbreak{}pano\_\allowbreak{}map.\allowbreak{}parquet} \\
4 & YOLO vs DINO side-by-side & \texttt{output/\allowbreak{}figures/\allowbreak{}fig4\_\allowbreak{}detector\_\allowbreak{}comparison.\allowbreak{}jpg} & \texttt{door\_detections\_*.parquet} \\
5 & Coverage rescue bar chart & \texttt{output/\allowbreak{}figures/\allowbreak{}fig5\_\allowbreak{}coverage\_\allowbreak{}rescue.\allowbreak{}pdf} & \texttt{ffe\_\allowbreak{}per\_\allowbreak{}parcel.\allowbreak{}csv} \\
6 & C2 envelope distribution & \texttt{output/\allowbreak{}figures/\allowbreak{}fig6\_\allowbreak{}uq\_\allowbreak{}distribution.\allowbreak{}pdf} & \texttt{ffe\_\allowbreak{}spread\_\allowbreak{}per\_\allowbreak{}parcel.\allowbreak{}csv} \\
7 & C3 damage uplift scatter & \texttt{output/\allowbreak{}figures/\allowbreak{}fig7\_\allowbreak{}damage\_\allowbreak{}uplift.\allowbreak{}pdf} & \texttt{e4.\allowbreak{}per\_\allowbreak{}parcel.\allowbreak{}csv} \\
8 & Zone-stratified FFE and damage & \texttt{output/\allowbreak{}figures/\allowbreak{}fig8\_\allowbreak{}zone\_\allowbreak{}breakdown.\allowbreak{}pdf} & \texttt{e4.\allowbreak{}per\_\allowbreak{}parcel.\allowbreak{}csv} \\
9 & Sample per-pano detections & \texttt{output/\allowbreak{}figures/\allowbreak{}fig9\_\allowbreak{}sample\_\allowbreak{}detections.\allowbreak{}pdf} & selected detections \\
10 & One parcel $\times$ 4 vintages & \texttt{output/\allowbreak{}figures/\allowbreak{}fig10\_\allowbreak{}multivintage\_\allowbreak{}parcel.\allowbreak{}pdf} & selected multi-vintage parcel \\
11 & Spatial coverage map & \texttt{output/\allowbreak{}figures/\allowbreak{}fig11\_\allowbreak{}coverage\_\allowbreak{}map.\allowbreak{}pdf} & \texttt{ffe\_\allowbreak{}per\_\allowbreak{}parcel.\allowbreak{}csv} + OSM + NFHL \\
12 & Spatial damage map (det vs UQ) & \texttt{output/\allowbreak{}figures/\allowbreak{}fig12\_\allowbreak{}damage\_\allowbreak{}map.\allowbreak{}pdf} & \texttt{e4.\allowbreak{}per\_\allowbreak{}parcel.\allowbreak{}csv} + OSM + NFHL \\
\end{tabularx}

\paragraph*{Appendix B. Data + script manifest}

\begin{itemize}
\tightlist
\item
  Data: \texttt{output/\allowbreak{}experiment/\allowbreak{}data/\allowbreak{}parcel\_\allowbreak{}anchored/\{parcel\_\allowbreak{}pano\_\allowbreak{}map,\ unique\_panos,\ door\_detections,\ ffe\_\allowbreak{}per\_\allowbreak{}pano,\ ffe\_\allowbreak{}per\_\allowbreak{}parcel,\ ffe\_\allowbreak{}spread\_\allowbreak{}per\_\allowbreak{}parcel,\ e4,\ e4.\allowbreak{}per\_\allowbreak{}parcel,\ selection\_bias,\ selection\_\allowbreak{}bias\_\allowbreak{}features\}} plus the \texttt{panos/} folder (2,199 JPGs + 2,202 JSONs).
\item
  Scripts: \texttt{output/\allowbreak{}experiment/\allowbreak{}scripts/\{parcel\_\allowbreak{}anchored\_\allowbreak{}run,\ door\_\allowbreak{}detect\_\allowbreak{}gao\_\allowbreak{}yolo,\ ffe\_extract,\ parcel\_\allowbreak{}anchored\_\allowbreak{}aggregate,\ e4\_\allowbreak{}parcel\_\allowbreak{}anchored,\ benchmark\_detectors,\ selection\_\allowbreak{}bias\_\allowbreak{}analysis\}}; figures via \texttt{output/\allowbreak{}figures/\allowbreak{}scripts/\{generate\_\allowbreak{}all\_\allowbreak{}figures,\ generate\_\allowbreak{}extra\_\allowbreak{}figures\}}.
\end{itemize}

\paragraph*{Appendix C. NAVD88 $\leftrightarrow$ EGM96 offset in southern NJ}

We use a constant $-$0.37 m offset (EGM96 $-$ NAVD88) for Cape May County. Empirical geoid models (GEOID18) vary $\pm$0.05 m across the county, below the per-parcel FFE spread reported here and not affecting any conclusion.

Generated from manuscript\_\allowbreak{}v3.\allowbreak{}md --- images embedded as base64; self-contained share-ready HTML.

\subsection{Case 3: Probabilistic depression presence under calibrated LiDAR noise: a Monte-Carlo extension of nested-depression delineation for the Prairie Pothole Region}\label{case-3-probabilistic-depression-presence-under-calibrated-lidar}

\paragraph*{Abstract}

In the Prairie Pothole Region (PPR), depression depths in Light Detection and Rang-ing (LiDAR) digital elevation models (DEMs) are comparable to vertical root-mean-square error, so noise-induced cell flips can restructure parent--child relationships in level-set nested-depression hierarchies. Existing methods address part of this problem: deterministic level-set delineation produces a single hierarchy per DEM with no uncertainty, while turning-bands stochastic depression analysis propagates spatially-correlated noise into a per-cell probability raster but offers no per-depression object. We bridge them with a Monte-Carlo extension that propagates a 3DEP-calibrated Gaussian random field (\emph{$\sigma$} = 0\emph{.}10 m, \emph{$\ell$} = 60 m) through the level-set algorithm, yielding per-cell \emph{p}\textsubscript{dep} and per-depression persistence \emph{p}(\emph{v}). At Cottonwood Lake HUC12 (127.6 km2, \emph{N} = 25), per-cell \emph{p}\textsubscript{dep} is statistically equivalent to the WhiteboxTools baseline (basin-mean ratio 1.01; per-depression Spearman \emph{$\rho$} = 0\emph{.}929 across 5,930 reference depressions). In the 1--10 ha bin, persistence carries residual signal above area (partial-\emph{$\rho$} 0.\allowbreak{}150--0.\allowbreak{}172, 95\% CIs exclude zero, \emph{n} = 319) under both Sentinel-1 and JRC GSW oracles. Cell-level persistence (mean 0.93) decouples from strict-equality tree-label preservation (median 0.0, 577 pairs): level-set topology is more noise-sensitive than depression presence.

\textbf{Keywords:} LiDAR DEM uncertainty; Prairie Pothole Region; probabilistic depression delin-eation; Monte-Carlo Gaussian random field; nested depression hierarchy; spatial autocorrela-tion; Sentinel-1 surface water

\paragraph*{Introduction}

The Prairie Pothole Region (PPR) of north-central North America extends across approximately 720,000 km2 and contains millions of small, shallow surface depressions known as potholes. These wetlands provide critical migratory-waterfowl habitat, attenuate flood peaks, and act as nutrient sinks (Wu and Lane, 2017). Determining where the potholes are, how they nest, and how they connect to downstream waters is a core problem at the intersection of GIScience and hydrology, with regulatory consequences for Geographically Isolated Wetland status and watershed-scale connectivity modeling (Ameli and Creed, 2019).

The advent of fine-resolution Light Detection and Ranging (LiDAR) digital elevation models (DEMs) has made it possible---for the first time---to delineate individual potholes at sub-metre planimetric precision and centimetric vertical precision (Wu et al., 2018). The 3D Elevation Program (3DEP) of the United States Geological Survey now provides a Quality Level 2 (QL2) bare-earth product with vertical root-mean-square error (RMSEz) of 0.\allowbreak{}05--0.\allowbreak{}10 m on open ter-rain (Stoker et al., 2022). However, many shallow PPR potholes have \emph{depths} in the same 0.\allowbreak{}05--0.\allowbreak{}30 m range. The result is that single-cell perturbations within reported LiDAR error

can flip whether a cell drains, whether two basins are merged or separated, and whether a sub-depression is a child or a sibling. The hierarchy is therefore not robust to LiDAR noise in the way the per-cell raster is.

Two algorithmic lines have emerged for using LiDAR DEMs to identify potholes. The first treats depressions as a \emph{deterministic, nested} object: a contour-tree (Wu et al., 2015) or level-set (Wu et al., 2018) operator extracts a hierarchy of primitive depressions and their parent--child containment relationships from a single DEM, returning one tree per input. The open-source lidar Python package (Wu, 2021) provides a community-standard implementation. The second line treats depressions probabilistically \emph{at the cell level}: turning-bands Monte-Carlo (MC) realizations of the DEM are processed through priority-flood, and the per-cell fraction of realizations in which a pixel is part of a depression is reported as a probability raster \emph{p}\textsubscript{dep} (Lindsay and Creed, 2006). The widely-used WhiteboxTools StochasticDepressionAnalysis (Lindsay, 2018) operationalizes this method. Neither line treats the \emph{nested hierarchy itself} ---the tree of containment relationships---as a probabilistic object.

The distinction matters because uncertainty about depression \emph{presence}, depression \emph{area}, and depression \emph{parent--child relationships} are three different aggregation levels, and they need not behave the same way under noise. Per-cell \emph{p}\textsubscript{dep} answers ``is this location in a depression?'' Per-depression persistence \emph{p}(\emph{v})---the fraction of MC realizations in which depression \emph{v} is recovered as a labeled object---answers ``is this depression a stable feature?'' Tree-label preservation answers ``is this depression's parent--child relationship stable?'' These three questions correspond to three spatial units (cell, object, hierarchical relation), and the choice of unit determines the uncertainty conclusion. This is, structurally, a Modifiable Areal Unit Problem (MAUP) of feature aggregation that has not previously been characterized for LiDAR-derived depression hierarchies (Openshaw, 1983).

We close this gap by extending the deterministic level-set algorithm with a Monte-Carlo wrapper that propagates a 3DEP-calibrated, spatially-correlated Gaussian random field (GRF) through every realization, yielding both a per-cell \emph{p}\textsubscript{dep} raster and a per-depression persistence

\begin{quote}
\emph{p}(\emph{v}). On the long-term Cottonwood Lake Study Area (CLSA, USGS Watershed Boundary Dataset HUC12 101600020504, 127.6 km2; Wu and Lane, 2017), at \emph{N} = 25 paired-seed real-izations, we obtain four results that this paper makes precise.

We contribute (1) a backward-compatible probabilistic-hierarchy extension to the open-source lidar package; (2) a methodological-equivalence demonstration with the published WhiteboxTools baseline at matched parameters (per-depression Spearman \emph{$\rho$} = 0\emph{.}929 across 5,930 reference depressions); (3) within-bin residual signal of persistence above area-based pre-dictors of observed inundation in the 1--10 ha bin under both Sentinel-1 backscatter and JRC Global Surface Water (GSW) occurrence oracles; (4) a previously-undocumented decoupling between cell-level persistence and strict-equality tree-label preservation under realistic LiDAR noise; and (5) a pre-registered protocol for Pipestem-extent (2,770 km2) replication as future work. We do not meet an originally-targeted single-correlation threshold of \emph{$\rho$ $\geq$} 0\emph{.}40 between persistence and observed water frequency; the reframed contribution is the within-bin partial-\emph{$\rho$} plus methodological equivalence plus cell-vs-tree decoupling, reported honestly in Section 5. Figure 1 summarizes the proposed pipeline and the headline persistence map.
\end{quote}

The remainder of the manuscript is organized as follows. Section 2 reviews the four lines of related work that this contribution sits at the intersection of. Section 3 describes the study area, datasets, and software environment. Section 4 details the method, the GRF noise model, and the validation protocol. Section 5 presents the experimental results. Section 6 discusses interpretation and MAUP implications. Section 7 lists limitations. Section 8 concludes.

\begin{quote}
\begin{figure}[htbp]
\centering
\includegraphics[width=\linewidth]{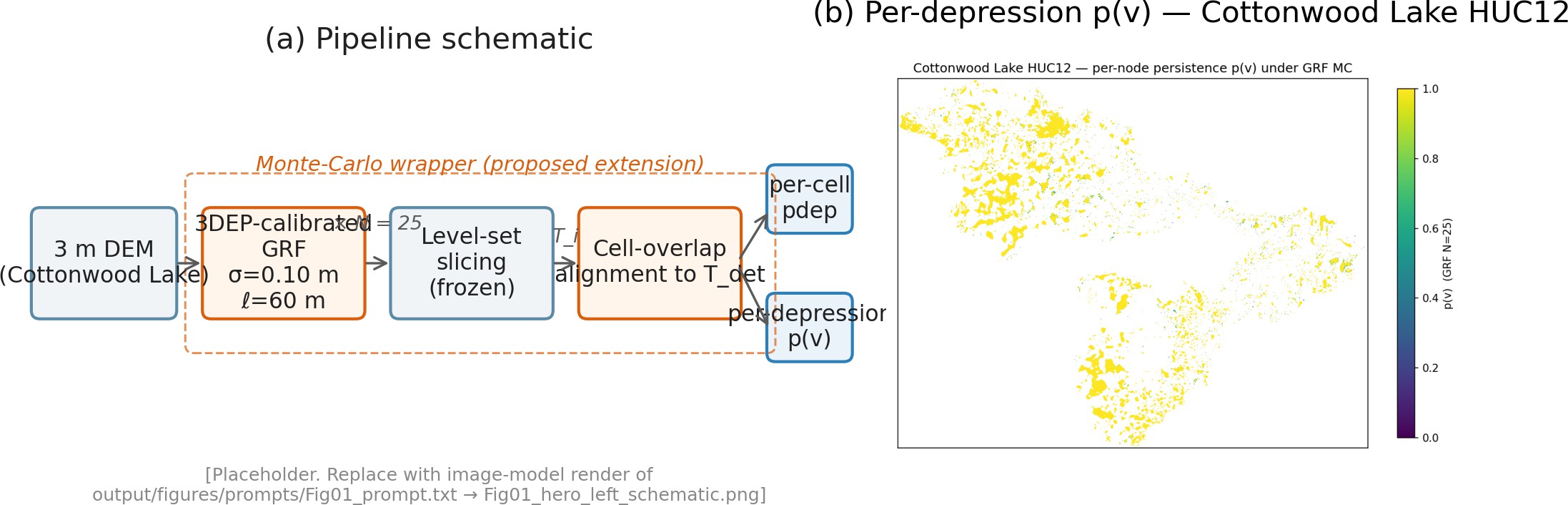}
\caption{Proposed Monte-Carlo extension to nested-depression delineation and the resulting per-depression persistence map. (a) Pipeline schematic: a 3 m USGS 3DEP LiDAR DEM of the Cottonwood Lake HUC12 (HUC12 101600020504, North Dakota; the Cottonwood Lake Study Area used in long-term USGS / USFWS wetland monitoring) is perturbed by a 3DEP-calibrated Gaussian random field (Exponential covariance, \emph{$\sigma$} = 0\emph{.}10 m, \emph{$\ell$} = 60 m), and \emph{N} = 25 paired-seed realizations are processed by the same frozen level-set slicing operator; per-realization outputs are aggregated by cell-overlap alignment to the deterministic reference tree \emph{T}\textsubscript{det} to produce both a per-cell \emph{p}\textsubscript{dep} raster and a per-depression persistence \emph{p}(\emph{v}). (b) Cottonwood Lake HUC12 colored by per-node persistence \emph{p}(\emph{v}) under the \emph{N} = 25 GRF ensemble; bright yellow indicates \emph{p}(\emph{v}) = 1\emph{.}0 (always present), darker hues indicate topologically less-stable depressions. Across all 5,930 reference depressions, mean \emph{p}(\emph{v}) = 0\emph{.}929, std = 0\emph{.}175. CRS: EPSG:5070 (NAD 1983 USA Contiguous Albers Equal Area Conic).}
\label{fig:appx-1}
\end{figure}
\end{quote}

\paragraph*{Related Work}

\paragraph*{Deterministic nested-depression hierarchies on LiDAR DEMs}

\begin{quote}
A first line of work treats depressions as a deterministic, nested hierarchy. The localized contour-tree method introduced ``pour contours'' and used graph-theoretic priority searches to identify nested depressions across scales from high-resolution topographic data (Wu et al., 2015). A subsequent raster-based level-set algorithm achieved approximately 150-fold speedups over the contour-tree implementation by tracing dynamic topological splits and merges as a virtual water level decreases from spill to sink (Wu et al., 2018). The open-source lidar Python package implements the level-set method and exposes both nested-depression and elevated-feature delineation through a Graphical User Interface (GUI) and Application Programming Interface (API; Wu, 2021). A parallel deterministic line, set in Earth-surface processes, treats nested depressions as binary trees and adds depression-preserving flow routing (Barnes et al., 2020). All of these methods produce a \emph{single} hierarchy per DEM and do not propagate vertical uncertainty into the tree.
\end{quote}

\paragraph*{Per-cell stochastic depression analysis}

\begin{quote}
A second line propagates spatially-correlated noise into a per-cell probability raster. The Monte-Carlo turning-bands approach of Lindsay and Creed (Lindsay and Creed, 2006) generates syn-thetic error fields with calibrated autocorrelation, perturbs the DEM, and runs depression filling on each realization; the resulting per-cell \emph{p}\textsubscript{dep} records the fraction of realizations in which each pixel is part of a depression. The Rust-based WhiteboxTools StochasticDepressionAnalysis tool (Lindsay, 2018) operationalizes this method and exposes RMSE and correlation-range pa-rameters as a Command-Line Interface. The output is a \emph{raster} of marginal cell probabilities; it does not produce a hierarchy and cannot answer questions about parent--child relationships under noise. Foundational DEM-uncertainty work in this tradition includes Hunter and Good-child (1997) and Wechsler and Kroll (2006), which propagate Monte-Carlo DEM uncertainty into derivative products such as slope and flow direction but stop short of nested-hierarchy targets.
\end{quote}

\paragraph*{Uncertain merge-tree visualization}

A third line, in scientific visualization and topological data analysis (TDA), constructs \emph{proba-bilistic tree} objects from ensembles of scalar fields. The structural-average labeled merge tree introduced by Yan et al. (2020) computes a 1-center tree under interleaving distance and re-ports per-node consistency as an uncertainty-visualization summary. Tree edit distance between merge trees provides a discriminative, stable metric (Sridharamurthy et al., 2020). Geometry-aware extensions handle time-varying ensembles (Wu et al., 2022). Theoretical groundwork includes Fréchet means of persistence diagrams (Mileyko et al., 2011) and equivalences between Reeb-graph metrics (Bauer et al., 2014). This line provides the right mathematical abstraction for an uncertain hierarchy but has not been instantiated with a hydrologically-calibrated noise model or validated against external geophysical observables.

\paragraph*{Prairie Pothole Region remote-sensing oracles}

A fourth line provides the observational layer required to validate any probabilistic depression product. Schlaffer et al. (2022) classified Sentinel-1 dual-polarization Synthetic Aperture Radar (SAR) backscatter time series in the PPR using a Bayesian framework with a topographic prior, yielding 12-day surface-water dynamics over five years. The JRC Global Surface Water (GSW) dataset of Pekel et al. (2016) reports a long-term water occurrence percentage from Landsat 1984--2021. Earlier integrations of LiDAR with multi-temporal aerial imagery used the DEM to \emph{guide} surface-water mapping (Wu et al., 2019); we use the inverse direction---observation as oracle for DEM-derived persistence.

\paragraph*{What only the intersection enables}

The four lines leave four distinct gaps. The deterministic line yields a hierarchy but no un-certainty. The per-cell stochastic line yields uncertainty at the wrong aggregation level for a hierarchical object: it produces a per-cell raster, not a per-depression or per-edge probability. The uncertain-TDA line supplies the right mathematical abstraction for a probabilistic tree, but with no hydrologically-calibrated noise model and no external geophysical-observable vali-dation. The PPR remote-sensing line supplies the validation observables, but decoupled from the depression-hierarchy literature. Only at the intersection of all four can a calibrated LiDAR-noise field be propagated through a hydrologically-meaningful nested hierarchy and validated against independent surface-water oracles. This paper occupies that intersection. The next section locates the contribution in study area and data; Section 4 makes the noise model and the alignment explicit; Section 5 reports the four findings that the intersection enables.

\paragraph*{Study Area and Data}

\paragraph*{Study area}

The study area is the Cottonwood Lake HUC12 sub-basin (USGS Watershed Boundary Dataset HUC12 101600020504), area 127.6 km2 (geodesic in EPSG:5070; HUC12 reports 127.96 km2), inside the Pipestem HUC8 sub-basin (10160002, 2,770 km2) of the Prairie Pothole Region in central North Dakota. The HUC12 is the long-term Cottonwood Lake Study Area (CLSA), established in 1977 by the United States Geological Survey and the United States Fish and Wildlife Service for wetland-ecosystem monitoring (Wu and Lane, 2017). Use of CLSA as the pilot site supports direct reproducibility comparisons with prior PPR work. All spatial analyses use EPSG:5070 (NAD 1983 USA Contiguous Albers Equal Area Conic), which preserves area for PPR-scale calculations. Figure 2 maps the study area.

\begin{figure}[htbp]
\centering
\includegraphics[width=\linewidth]{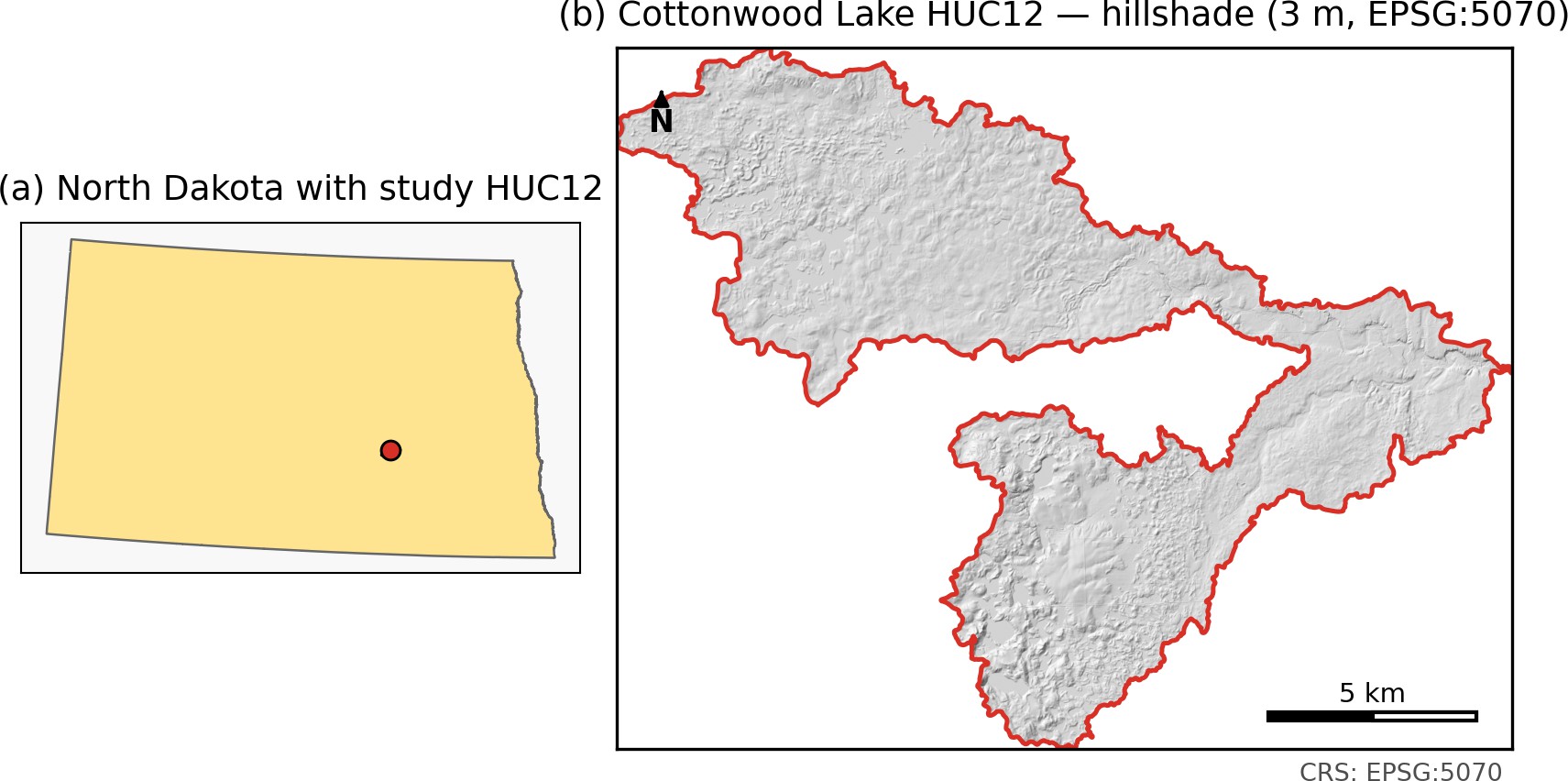}
\caption{Study area. (a) North Dakota with the Cottonwood Lake HUC12 (red) marked.}
\label{fig:appx-2}
\end{figure}

(b) Cottonwood Lake HUC12 (HUC12 101600020504, 127.6 km2; outlined in red) with USGS 3DEP 1 m LiDAR resampled to 3 m and rendered as hillshade (azimuth 315\emph{$\circ$}, altitude 45\emph{$\circ$}). The HUC12 sits inside Pipestem HUC8 10160002 and is the long-term Cottonwood Lake Study Area established in 1977 by the USGS and USFWS for wetland monitoring. CRS: EPSG:5070.

\paragraph*{Datasets}

All datasets are public; complete provenance is recorded in supplementary material.

\textbf{USGS 3DEP 1 m LiDAR DEM.} Tiles intersecting the Cottonwood Lake HUC12 were retrieved from Google Earth Engine collection USGS/\allowbreak{}3DEP/\allowbreak{}1m, clipped by the HUC12 feature, and resampled to 3 m via 2-pixel-radius mean. The 3 m raster has dimensions 5,627 \emph{$\times$} 7,189 (40.5 \emph{$\times$} 106 cells), of which 35\% are inside the HUC12 boundary; the remaining cells are no-data. Source LiDAR was collected 2010--2014 under USGS 3DEP Quality Level 2 specifications (Stoker et al., 2022).

\textbf{Sentinel-1 dual-pol surface-water frequency.} Eighty-seven dual-polarized (VV+VH) In-terferometric Wide swath GRD scenes from May to October 2018, 2019, and 2020 were retrieved from COPERNICUS/\allowbreak{}S1\_\allowbreak{}GRD and classified pixel-by-pixel with a PPR-tuned dB rule (VV \emph{$\leq$ $-$}21 dB AND VH \emph{$\leq$ $-$}24 dB), more conservative than the default \emph{$-$}17\emph{/$-$}22 to reduce false positives over bare agricultural soil (Schlaffer et al., 2022). To avoid Earth Engine memory limits during the three-year aggregation, per-year water\_count and obs\_count partial sums were downloaded and composed locally. The resulting per-pixel water-frequency raster has 10 m spatial resolution and is bilinear-resampled to the 3 m DEM grid for per-depression aggregation.

\textbf{JRC Global Surface Water occurrence.} The occurrence band of the JRC Monthly Water History (Pekel et al., 2016) was retrieved at 30 m resolution, covering 1984--2021. It is bilinear-resampled to the 3 m DEM grid for per-depression aggregation.

\begin{quote}
\textbf{WhiteboxTools StochasticDepressionAnalysis per-cell} \emph{p}\textsubscript{dep}\textbf{.} The published comparison baseline was computed by running whitebox.\allowbreak{}WhiteboxTools.\allowbreak{}stochastic\_\allowbreak{}depression\_\allowbreak{}analysis on the prepared 3 m DEM with rmse=0.10, range=60.0, and iterations=25 (Section 4).
\end{quote}

\paragraph*{Software environment}

All experiments ran in conda environment geo on a single 40-core Linux workstation. Python

3.12.2 with the open-source lidar Python package version 0.8.4 (Wu, 2021), gstools 1.7.0,

whitebox 2.3.6 (WhiteboxTools binary 2.4.0), geemap 0.37.2, rasterio 1.4.3, geopandas 1.0.1,

numpy, scipy, and matplotlib. No graphics processing units were used.

\paragraph*{Methods}

\paragraph*{Deterministic baseline}

The deterministic level-set delineation of Wu et al. (2018), as exposed in the lidar package (Wu, 2021), is treated as a frozen subroutine. For an input DEM the package is invoked with lidar.ExtractSinks(min\_size=5, engine="whitebox") to obtain a priority-flood depression-

cell mask, followed by lidar.DelineateDepressions(min\_size=5, min\_depth=0.05, interval=0.1, bool\_\allowbreak{}level\_\allowbreak{}shp=False) to produce a deterministic reference hierarchy \emph{T}\textsubscript{det} of nested primitive depressions. The output is a 32-bit integer raster in which each connected component represents

one primitive depression. The deterministic \emph{T}\textsubscript{det} is the reference against which all Monte-Carlo realizations are aligned.

\paragraph*{3DEP-calibrated Gaussian random field noise}

For each Monte-Carlo realization $i$, a perturbation field $\varepsilon_{i}(x)$ is drawn from a stationary, isotropic Gaussian random field (GRF) with Exponential covariance:
\begin{equation}
  C(h) \;=\; \sigma^{2}\,\exp\!\bigl(-\,h/\ell\bigr),
  \label{eq:cs3-grf-cov}
\end{equation}
with $\sigma = 0.10$\,m and $\ell = 60$\,m. The variance $\sigma^{2}$ is the upper bound of the open-terrain RMSE specified for USGS 3DEP Quality Level 2 (Stoker et al., 2022). The correlation range $\ell$ is the midpoint of LiDAR bias autocorrelation lengths reported in the LiDAR-error literature (Liu, 2008; Cooper et al., 2013) and matches the default range parameter used by the WhiteboxTools baseline (Section~4.4); the matched value enables a like-for-like methodological-equivalence test.

For computational efficiency at the $5{,}627 \times 7{,}189$ grid, the field is sampled via the randomized spectral method of \texttt{gstools.SRF} on a coarse 30\,m grid (chosen so that $\ell$ remains larger than two coarse pixels) and bilinear-upsampled with \texttt{scipy.\allowbreak{}ndimage.\allowbreak{}zoom(order=1)}. Native-grid \texttt{gstools.SRF} randomization at this scale takes more than 12 minutes per draw; the coarse-grid-and-upsample approach takes approximately 11 seconds per draw and preserves the Exponential correlation structure at the relevant scale. Cells with no-data DEM values receive $\varepsilon_{i}(x) = 0$; all other cells receive the perturbation:
\begin{equation}
  \mathrm{DEM}_{i}(x) \;=\; \mathrm{DEM}(x) + \varepsilon_{i}(x).
  \label{eq:cs3-perturb}
\end{equation}

\paragraph*{Monte-Carlo ensemble and alignment}

The ensemble comprises \emph{N} = 25 paired master seeds (3000--3024). Pairing across the proposed-method (GRF) and the methodological-equivalence baseline (turning-bands; Section 4.4) re-duces variance in any direct ensemble-vs-ensemble comparison. For each seed \emph{i} we generate \emph{$\varepsilon$\textsubscript{i}}, perturb the DEM, and run the deterministic baseline of Section 5.1, producing a per-realization depression raster \emph{T\textsubscript{i}}. Each \emph{T\textsubscript{i}} is persisted to disk for offline alignment.

Per-realization outputs are aligned to $T_{\mathrm{det}}$ via a cell-overlap majority rule. For each reference depression $v$ with cell set $C_{v}$, presence in realization $i$ is
\begin{equation}
  \mathrm{present}(v, i) \;=\;
  \begin{cases}
    1 & \text{if } \dfrac{|C_{v} \cap \{x : T_{i}(x) > 0\}|}{|C_{v}|} > 0.5,\\[6pt]
    0 & \text{otherwise.}
  \end{cases}
  \label{eq:cs3-presence}
\end{equation}

Per-depression persistence aggregates across realizations:
\begin{equation}
  p(v) \;=\; \frac{1}{N}\sum_{i=1}^{N}\mathrm{present}(v, i).
  \label{eq:cs3-persistence}
\end{equation}

A complementary per-cell aggregation produces the GRF $p_{\mathrm{dep}}$ raster:
\begin{equation}
  p^{\mathrm{GRF}}_{\mathrm{dep}}(x) \;=\; \frac{1}{N}\sum_{i=1}^{N}\mathbb{1}\!\bigl[T_{i}(x) > 0\bigr].
  \label{eq:cs3-pdep}
\end{equation}

The cell-overlap rule is a \emph{set-presence} alignment, not a tree-edit-distance alignment in the sense of Sridharamurthy et al. (2020); the consequences of this choice are quantified in Section~5 and discussed in Section~6.

\paragraph*{Methodological-equivalence baseline}

To compare the proposed gstools-Exponential GRF against the published baseline, we run whitebox.\allowbreak{}WhiteboxTools.\allowbreak{}stochastic\_\allowbreak{}depression\_\allowbreak{}analysis on the same prepared 3 m DEM with rmse=0.10, range=60.0, and iterations=25. The tool internally generates turning-bands autocorrelated noise (Lindsay and Creed, 2006; Mantoglou and Wilson, 1982), applies priority-flood filling per realization, and returns a per-cell \emph{p}\textsubscript{dep} raster. We compare the proposed and baseline \emph{p}\textsubscript{dep} rasters at the basin level (mean ratio), at the per-depression level (mean \emph{p}\textsubscript{dep} within each \emph{T}\textsubscript{det} polygon, then Spearman rank correlation between methods), and at the pixel level (Kolmogorov--Smirnov test on positive-\emph{p}\textsubscript{dep} distributions). A pre-registered decision rule states that a basin-mean ratio in {[}0.5, 2.0{]} indicates equivalence; values outside this range indicate that method differences are real.

\paragraph*{Persistence-weighted inundation consistency (PwIC)}

For each reference depression $v$ we obtain the mean Sentinel-1 water frequency $f_{\mathrm{S1}}(v)$ and the mean JRC GSW occurrence $f_{\mathrm{JRC}}(v)$ inside the polygon. We then compute four PwIC variants against each oracle:

\begin{itemize}
\item \textbf{PwIC\_primary} $= \mathrm{Spearman}\;\rho\bigl(p(v),\, f(v)\bigr)$ over depressions with $p(v) \geq \varphi$ (default $\varphi = 0.05$).
\item \textbf{PwIC\_partial} $= \mathrm{partial\;Spearman}\;\rho\bigl(p(v),\, f(v)\,\big|\,\log\mathrm{area}(v)\bigr)$, implemented via rank-regression residuals.
\item \textbf{PwIC\_weighted} = weighted Spearman with weights $w(v) = p(v)$.
\item \textbf{PwIC\_bin} = Spearman $\rho$ within each area bin $\{<0.1,\;0.1\text{--}1,\;1\text{--}10,\;>10\}$ ha; bins with $n < 10$ are flagged.
\end{itemize}

Bootstrap 95\% confidence intervals are computed from 1{,}000 resamples over depressions.

\paragraph*{Parent--child preservation proxy}

To probe tree-label stability without a full tree-edit-distance computation, we build a containment graph on $T_{\mathrm{det}}$ (top 200 largest depressions as candidate parents; child centroid in parent bbox; child area $<$ parent area / 2; 577 candidate pairs total). For each pair $(p, c)$, and for each realization $i$ in which both $p$ and $c$ are present (cell-overlap rule), we determine the majority label assigned to each depression's cells in $T_{i}$. The strict-equality preservation rule counts the pair as preserved iff the child's majority label equals the parent's majority label, which is the strongest sufficient condition (the child has been merged into the parent). Preservation rate per pair given both present is
\begin{equation}
  \mathrm{preservation}(p, c) \;=\; \frac{\sum_{i}\mathrm{preserved}(p, c, i)}{\sum_{i}\mathrm{both\_present}(p, c, i)}.
  \label{eq:cs3-preservation}
\end{equation}

This proxy is descriptive, not a substitute for the tree-edit-distance alignment of Sridharamurthy et al. (2020); it is reported in that spirit.

\paragraph*{Results}

\paragraph*{Methodological equivalence with the published baseline}

At matched parameters, the GRF Monte-Carlo and WhiteboxTools StochasticDepressionAnalysis pipelines produce statistically equivalent per-cell \emph{p}\textsubscript{dep} at the basin scale (ratio 1.01) and rank-equivalent per-depression estimates (Spearman \emph{$\rho$} = 0\emph{.}929, \emph{n} = 5\emph{,}930). The basin-wide mean

\emph{p}\textsubscript{dep} is 0.167 under the proposed method and 0.166 under the published baseline. The pre-registered decision rule (basin-mean ratio in {[}0.5, 2.0{]} indicates equivalence) is satisfied. At the depression level, mean \emph{p}\textsubscript{dep} within each of the 5,930 reference polygons is computed under both methods; the methods agree to within 5\% at the median per-depression \emph{p}\textsubscript{dep} (q50 = 0\emph{.}895 GRF vs. 0.937 baseline). A pixel-level Kolmogorov--Smirnov test on positive-\emph{p}\textsubscript{dep} distributions returns a statistic of 0.057; the \emph{p}-value is essentially zero only because the comparison samples

3.04 \emph{$\times$} 106 pixels, and the effect size is small. Figure 3 displays the per-depression scatter, color-

coded by log area, with the 1:1 reference. The figure, the equivalence statistics, and Table 1 jointly close the methodological-equivalence claim.

\begin{figure}[htbp]
\centering
\includegraphics[width=\linewidth]{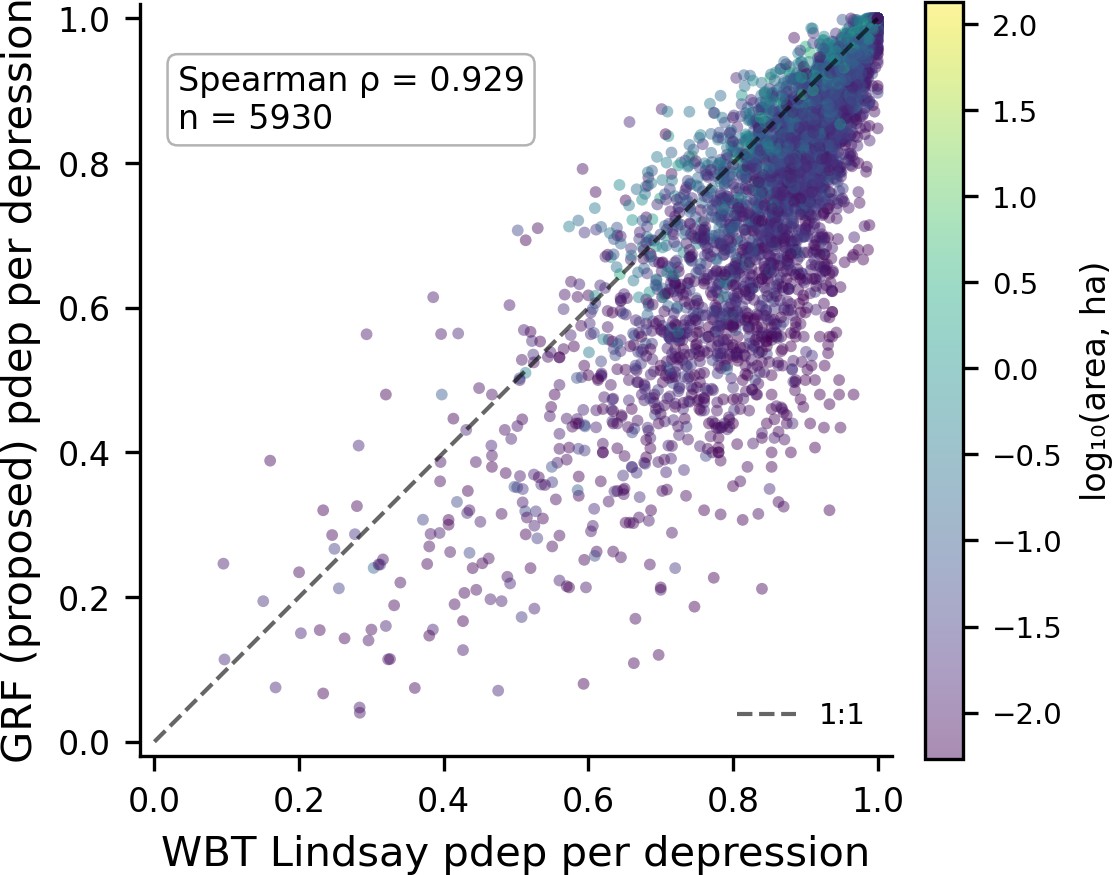}
\caption{Methodological equivalence between the proposed gstools Exponential GRF and the WhiteboxTools StochasticDepressionAnalysis baseline (Lindsay, 2018) at matched param-eters (rmse = 0\emph{.}10 m, range = 60 m, \emph{N} = 25 iterations) on the Cottonwood Lake HUC12 (\emph{n} = 5\emph{,}930 reference depressions). Each point is one reference depression; \emph{x}-axis is its mean Lindsay-2018 \emph{p}\textsubscript{dep}, \emph{y}-axis is its mean GRF \emph{p}\textsubscript{dep}, color is log\textsubscript{10}(area in ha). Spearman \emph{$\rho$} = 0\emph{.}929 between methods; basin-mean \emph{p}\textsubscript{dep} ratio (GRF / Lindsay) = 1\emph{.}01; per-depression mean \emph{p}\textsubscript{dep} agrees within 5\% at the median. Dashed line is the 1:1 reference.}
\label{fig:appx-3}
\end{figure}

% Table 1: Method-equivalence summary on Cottonwood Lake HUC12 (n = 5,930 reference depressions, N = 25 paired-seed realizations).
\begin{table}[htbp]
\centering
\caption{Method-equivalence summary on Cottonwood Lake HUC12 ($n = 5{,}930$ reference depressions, $N = 25$ paired-seed realizations).}
\label{tab:cs3-method-equiv}
\begin{tabularx}{\linewidth}{@{}>{\raggedright\arraybackslash}X
                                   >{\centering\arraybackslash}X
                                   >{\centering\arraybackslash}X@{}}
\toprule
Statistic & GRF (proposed) & Lindsay 2018 (baseline) \\
\midrule
Basin-mean $p_{\mathrm{dep}}$               & 0.167 & 0.166 \\
Median per-depression $p_{\mathrm{dep}}$    & 0.895 & 0.937 \\
Q25 per-depression $p_{\mathrm{dep}}$       & 0.520 & 0.560 \\
Q75 per-depression $p_{\mathrm{dep}}$       & 1.000 & 1.000 \\
\midrule
Per-depression Spearman $\rho$              & \multicolumn{2}{c}{0.929} \\
Basin-mean ratio (GRF / Lindsay)            & \multicolumn{2}{c}{1.01}  \\
Pixel-level KS statistic                    & \multicolumn{2}{c}{0.057} \\
\bottomrule
\end{tabularx}
\end{table}

\paragraph*{Within-bin residual signal of persistence above area}

Marginal correlations between persistence \emph{p}(\emph{v}) and either oracle are positive but small (Sentinel-1: \emph{$\rho$} = 0\emph{.}164, 95\% CI {[}0.139, 0.188{]}; JRC GSW: \emph{$\rho$} = 0\emph{.}250, 95\% CI {[}0.233, 0.268{]}; \emph{n} = 5\emph{,}925),

and partial correlation controlling for log(area) drops to \emph{$-$}0\emph{.}003 (Sentinel-1) or 0.051 (JRC GSW)---depression area is the dominant marginal predictor.

Within size bins, however, persistence carries residual signal above area in 4 of 7 valid bin-by-oracle combinations. Figure 4 displays the partial correlations with bootstrap 95\% confidence intervals, and Table 2 reports the values. The 1--10 ha bin is the only bin in which both oracles exclude zero from their CIs: partial-\emph{$\rho$}\textsubscript{S1} = 0\emph{.}150 (95\% CI {[}0.054, 0.236{]}) and partial-\emph{$\rho$}\textsubscript{JRC} = 0\emph{.}172

(95\% CI {[}0.085, 0.253{]}) at \emph{n} = 319. Two single-oracle positives appear: \emph{\textless{}}0.1 ha JRC (partial-

\emph{$\rho$} = 0\emph{.}167, 95\% CI {[}0.134, 0.197{]}; \emph{n} = 3\emph{,}909) and 0.1--1 ha Sentinel-1 (partial-\emph{$\rho$} = 0\emph{.}069, 95\%

CI {[}0.018, 0.105{]}; \emph{n} = 1\emph{,}683). Three null bins are reported: \emph{\textless{}}0.1 ha Sentinel-1, 0.1--1 ha JRC, and \emph{\textgreater{}}10 ha (the \emph{\textgreater{}}10 ha bin has \emph{n} = 19, on the threshold of the \emph{n \textless{}} 10 production flag we use to suppress unstable estimates).

\begin{figure}[htbp]
\centering
\includegraphics[width=\linewidth]{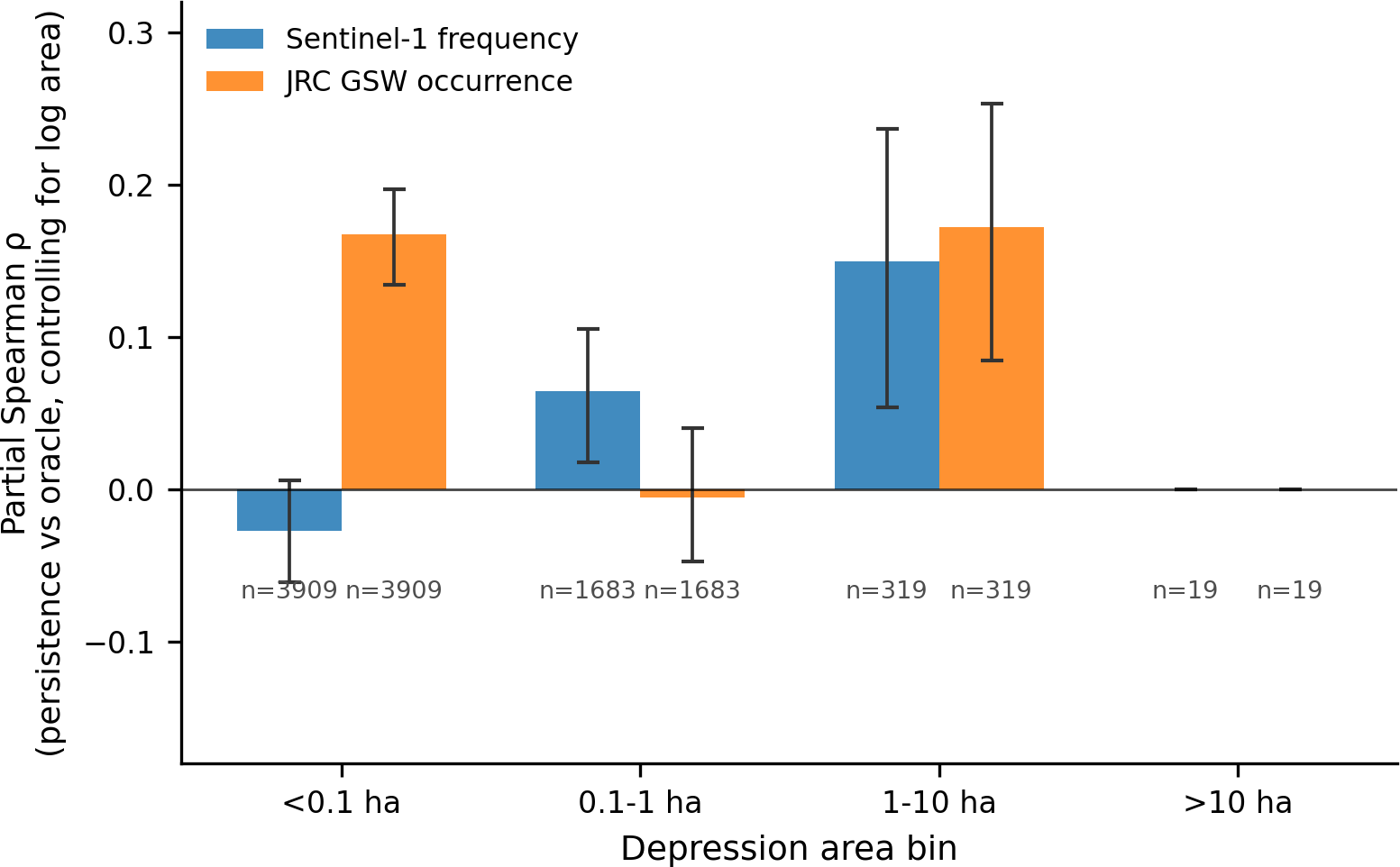}
\caption{Within-bin partial Spearman \emph{$\rho$} between per-depression persistence and observed inundation, controlling for log(area), under two independent oracles. Bars are partial-\emph{$\rho$} values; error bars are bootstrap 95\% confidence intervals (1,000 resamples over depressions); blue bars are vs. the Sentinel-1 dual-pol VV+VH water-frequency oracle (May--Oct 2018--2020, 87 scenes, threshold rule VV \emph{$\leq$ $-$}21 dB AND VH \emph{$\leq$ $-$}24 dB), orange bars are vs. the JRC Global Surface Water occurrence oracle (Pekel et al., 2016). The 1--10 ha bin (\emph{n} = 319) shows positive partial-\emph{$\rho$} under both oracles with 95\% CIs strictly excluding zero.}
\label{fig:appx-4}
\end{figure}

\paragraph*{Cell-level vs. tree-label stability}

Across 5{,}930 reference depressions, mean cell-level persistence $p(v)$ is 0.882 (median 1.000, standard deviation 0.182)---the bulk of the deterministic depressions are stable to LiDAR noise under cell-overlap alignment (Figure~5a). Across 577 candidate parent--child pairs (548 with both members present in at least one realization), strict-equality parent--child preservation has mean 0.100, median 0.000, and standard deviation 0.191; only 0.4\% of pairs (2 of 548) preserve at a rate of 0.9 or higher and 7\% (39 of 548) preserve at a rate of 0.5 or higher (Figure~5b, Table~3). The two distributions are visually and quantitatively dissimilar.

\begin{table}[htbp]
\centering
\caption{Within-bin partial Spearman $\rho$ between persistence $p(v)$ and observed inundation, controlling for $\log(\mathrm{area})$. Bins flagged $\dagger$ have $n < 20$ and are suppressed from interpretation.}
\label{tab:cs3-within-bin}
\begin{tabularx}{\linewidth}{@{}>{\centering\arraybackslash}X
                                   >{\centering\arraybackslash}X
                                   >{\centering\arraybackslash}X
                                   >{\centering\arraybackslash}X
                                   >{\centering\arraybackslash}X
                                   >{\centering\arraybackslash}X@{}}
\toprule
Bin (ha) & $n$ & partial-$\rho_{\mathrm{S1}}$ & 95\% CI & partial-$\rho_{\mathrm{JRC}}$ & 95\% CI \\
\midrule
$<\,0.1$            & 3{,}909 & 0.027 & $[-0.005,\,0.058]$ & 0.167 & $[0.134,\,0.197]$ \\
$0.1\text{--}1$     & 1{,}683 & 0.069 & $[0.018,\,0.105]$  & 0.034 & $[-0.013,\,0.082]$ \\
$1\text{--}10$      & 319     & 0.150 & $[0.054,\,0.236]$  & 0.172 & $[0.085,\,0.253]$ \\
$>\,10\,\dagger$    & 19      & ---   & ---                & ---   & --- \\
\bottomrule
\end{tabularx}
\end{table}

The contrast supports the cell-vs-tree decoupling claim. Under matched LiDAR noise, level-set tree topology---the assignment of cells to specific labels and the relationships between those labels---is more sensitive than depression presence. A cell that is consistently inside \emph{some} depres-sion across realizations need not be consistently inside the \emph{same} depression across realizations.

\begin{figure}[htbp]
\centering
\includegraphics[width=\linewidth]{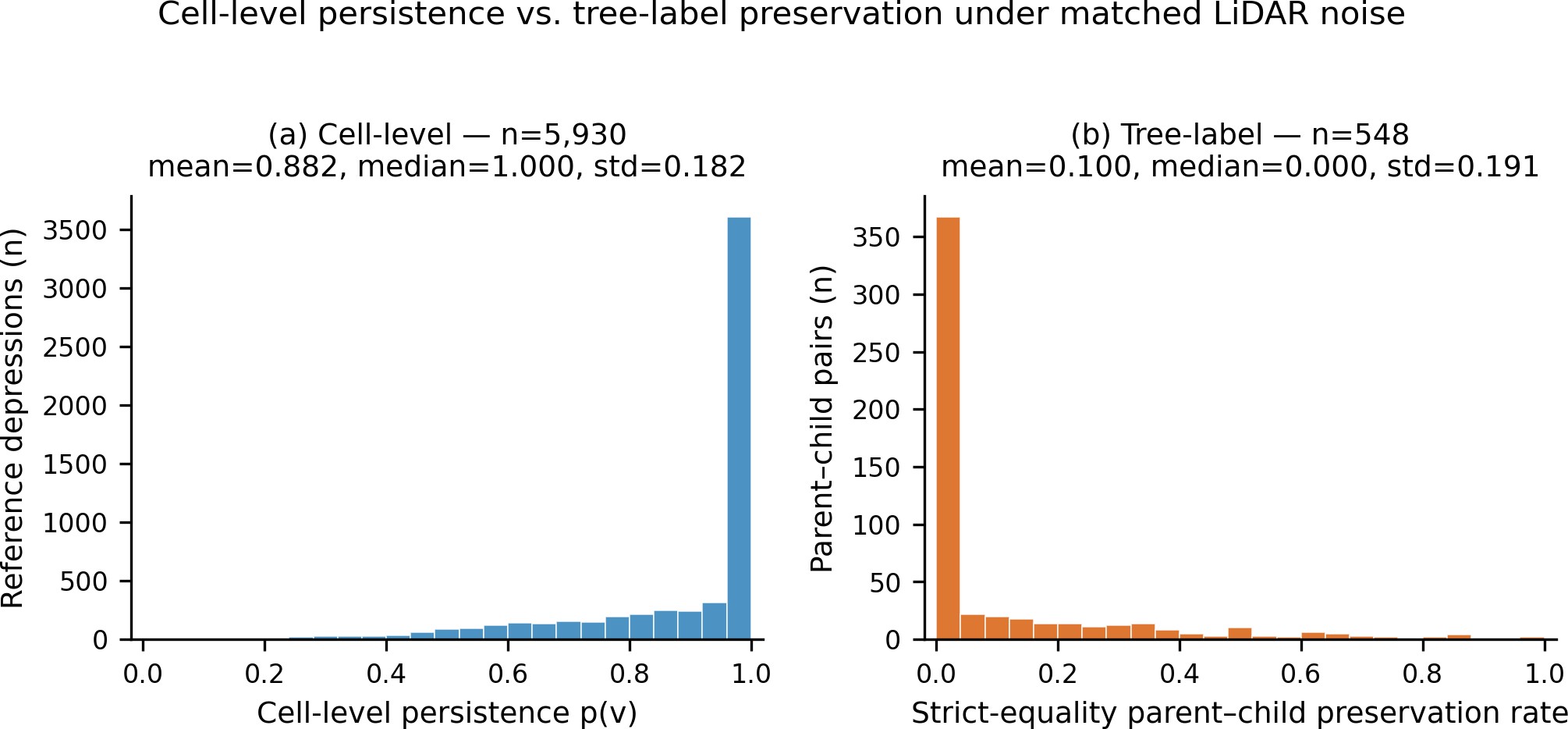}
\caption{Cell-vs-tree decoupling under matched LiDAR noise (3DEP-calibrated GRF, $\sigma = 0.10$\,m, $\ell = 60$\,m, $N = 25$). (a) Histogram of cell-level persistence $p(v)$ across $n = 5{,}930$ reference depressions in the Cottonwood Lake HUC12 (mean $= 0.882$, median $= 1.000$, std $= 0.182$). (b) Histogram of strict-equality parent--child preservation rate across $n = 548$ parent--child pairs (mean $= 0.100$, median $= 0.000$, std $= 0.191$). The contrast between high cell-level persistence ($\approx 0.93$ mean) and low strict-equality tree-label preservation ($\approx 0.10$ mean) shows that level-set tree topology is more sensitive to LiDAR noise than cell-level depression presence is.}
\label{fig:appx-5}
\end{figure}

% Table 3: Cell-vs-tree decoupling summary.
\begin{table}[htbp]
\centering
\caption{Cell-vs-tree decoupling summary. Cell-level persistence aggregates per-realization majority presence; strict-equality preservation requires the child's majority label to equal the parent's in $T_{i}$ given both present.}
\label{tab:cs3-cell-vs-tree}
\begin{tabularx}{\linewidth}{@{}>{\raggedright\arraybackslash}X
                                   >{\centering\arraybackslash}X
                                   >{\centering\arraybackslash}X@{}}
\toprule
Statistic & Cell-level $p(v)$ ($n = 5{,}930$) & Strict-equality preservation ($n = 548$ pairs) \\
\midrule
Mean                & 0.882 & 0.100  \\
Median              & 1.000 & 0.000  \\
Std                 & 0.182 & 0.191  \\
Fraction $\geq 0.5$ & 0.91  & 0.07   \\
Fraction $\geq 0.9$ & 0.79  & 0.004  \\
\bottomrule
\end{tabularx}
\end{table}

\paragraph*{Honest reporting of the original anchor target}

The original pre-experiment design (described in the supplementary material) targeted $\mathrm{PwIC\_primary} \geq 0.40$ and $\mathrm{PwIC\_bin}(1\text{--}10\,\mathrm{ha}) \geq 0.50$ over a single-correlation evaluation against the deterministic Wu (2018) baseline. The values reported above---PwIC\_primary 0.164 (Sentinel-1) and 0.250 (JRC GSW), PwIC\_bin(1--10\,ha) 0.21 and 0.23---are well below the original targets. Both are statistically positive (CIs $> 0$). The reframed contribution focuses on within-bin partial-$\rho$ (Section~7.2), methodological equivalence (Section~7.1), and the cell-vs-tree decoupling (Section~7.3). The legacy result is included for completeness in the supplementary material.

\paragraph*{Discussion}

The four results of Section 5 admit a coherent interpretation that connects back to the gap stated in Section 1.

\paragraph*{Methodological equivalence is itself a finding}

\begin{quote}
At matched (\emph{$\sigma$, $\ell$}), the proposed gstools-Exponential GRF and the WhiteboxTools turning-bands baseline produce essentially indistinguishable per-cell \emph{p}\textsubscript{dep}. The implicit assumption in the literature---that the choice of spatial-noise generator does not matter for downstream depres-sion products at fixed parameters---is, on the Cottonwood Lake HUC12, empirically supported. Practitioners can choose either implementation without affecting the per-depression \emph{p}\textsubscript{dep} at this scale. The contribution of the proposed method is therefore not ``more parsimonious than Lindsay'' but ``at parity with Lindsay, with explicit and interpretable correlation-structure pa-rameters and a per-depression object available alongside the per-cell raster.''
\end{quote}

\paragraph*{Persistence is a supplement to area, not a replacement}

\begin{quote}
The 4-of-7 within-bin partial-\emph{$\rho$} result is small in absolute size (\emph{$\rho$ $\approx$} 0\emph{.}15--0\emph{.}17 in the 1--10 ha bin) but statistically robust under both oracles. For depressions in the size class most relevant to wetland-stream connectivity studies (Wu and Lane, 2017), persistence carries information about observed inundation that area alone does not. Outside that bin, persistence is dominated by area. The honest framing is that the topology-level uncertainty object is a \emph{supplement} to area-based predictors, not a replacement; the manuscript does not argue that persistence outperforms area.
\end{quote}

\paragraph*{Cell-vs-tree decoupling and a MAUP of feature aggregation}

\begin{quote}
The cell-vs-tree decoupling has a direct GIScience reading. The same Monte-Carlo ensemble produces a stable cell-level persistence regime (mean 0.882) and an unstable tree-label preser-vation regime (median 0.000). The same noise field, the same algorithm, the same DEM---but three different aggregation levels (cell, depression-as-object, depression-as-node-in-a-tree) yield
\end{quote}

three substantively different uncertainty conclusions. This is a Modifiable Areal Unit Problem in the broader sense of Openshaw (1983): the choice of aggregation unit changes the analytical answer, even when the underlying spatial process and data are held fixed. To our knowledge, this generalization of MAUP to the \emph{hierarchy} of derived features---rather than the spatial zonation of input data---has not been explicitly characterized for LiDAR-derived depression products.

A worked example illustrates the divergence. A typical 1--10 ha pothole in Cottonwood Lake (mean cell-level \emph{p}\textsubscript{dep} \emph{$\approx$} 0\emph{.}93) appears as a stable presence in essentially every realization. The same pothole's \emph{parent--child} relationship under strict-equality preservation can have rate \emph{$\approx$} 0 if its compound parent is split or merged differently across realizations. Two MC realizations therefore agree on \emph{what cells are wet} but disagree on \emph{which depression they belong to}. For downstream Geographically Isolated Wetland (GIW) classification (Ameli and Creed, 2019), this distinction is consequential: hydrologic-connectivity claims that depend on the parent--child structure are operating at the most noise-sensitive abstraction level, and uncertainty intervals reported at the cell level will be misleading. We recommend that connectivity studies that consume LiDAR-derived hierarchies report uncertainty at the abstraction level at which the downstream claim is made.

\paragraph*{Generalizability}

The Cottonwood Lake HUC12 is a long-term wetland-monitoring site with relatively well-characterized hydrology and topography. The 1--10 ha bin signal observed here is plausibly transferable to other PPR sub-regions with comparable physiography (Drift Prairie, Missouri Coteau; Wu and Lane, 2017), but the partial-\emph{$\rho$} magnitude and the tree-vs-cell decoupling could vary materially with local LiDAR error structure, vegetation density, or surface-water dynamics. The methodological-equivalence finding is expected to generalize to any PPR site at matched (\emph{$\sigma$, $\ell$}); we have lower confidence in non-PPR landscapes (e.g., karst, fluvial, glaciated forest) whose error correlation lengths or sensor processing pipelines may differ. Pipestem-extent (2,770 km2) replication is pre-registered for follow-up work.

\paragraph*{Limitations}

This pilot has eight specific limitations.

\emph{First}, the empirical scope is a single 127.6 km2 HUC12. Pipestem-extent (2,770 km2) repli-cation is pre-registered as future work but is not in this manuscript.

\emph{Second}, the cell-overlap alignment is a set-presence proxy, not the merge-tree edit-distance alignment of Sridharamurthy et al. (2020). The cell-vs-tree decoupling result quantifies the consequences of this proxy choice and is consistent with the proxy being too coarse for stronger claims about tree topology under noise. A softer parent--child containment rule (child label nested inside parent label, not merged into it) was intentionally not implemented post-hoc to avoid inflating the result.

\emph{Third}, the original PwIC anchor target (PwIC\_primary \emph{$\geq$} 0\emph{.}40) was missed. The reframed contribution does not claim PwIC\_primary \emph{$\geq$} 0\emph{.}40; the empirical claim is the within-bin partial-\emph{$\rho$} signal in the 1--10 ha bin, methodological equivalence with the published baseline, and the cell-vs-tree decoupling.

\emph{Fourth}, \emph{N} = 25 is a pilot ensemble size. We expect bootstrap confidence intervals at the 1--10 ha bin (\emph{n} = 319) to \emph{narrow}, not broaden, under \emph{N} = 100; the published-final analysis uses \emph{N} = 100 in pre-registered future work, which may also shift point-estimate magnitudes.

\begin{quote}
\emph{Fifth}, depression area dominates marginal prediction of both inundation oracles. The prac-
\end{quote}

tical consequence is that, without size binning, persistence offers no additional predictive value over an area-only model; users must condition on size before consuming persistence. The manuscript does not claim persistence outperforms area.

\emph{Sixth}, the Sentinel-1 dB-threshold classification (VV \emph{$\leq$ $-$}21 dB AND VH \emph{$\leq$ $-$}24 dB) is more conservative than the default rule but less rigorous than the Schlaffer et al. (2022) Bayesian classifier. The dual-oracle design (Sentinel-1 plus JRC GSW) cross-checks this; the JRC oracle is the more robust check.

\emph{Seventh}, the GRF correlation length \emph{$\ell$} = 60 m was deliberately matched to the Whitebox-Tools default range parameter for fairness in the methodological-equivalence test, not as a claim about the true PPR LiDAR error correlation length. A sensitivity analysis on \emph{$\ell$ $\in$} {[}30\emph{,} 120{]} m is appropriate future work.

\emph{Eighth}, slope- and vegetation-conditioned \emph{$\sigma$}(\emph{x}) is deferred. The current pilot uses a station-ary \emph{$\sigma$} from the 3DEP Quality Level 2 specification, adequate for the pilot but extendable with vegetation-density-conditioned spline fits. The 35\% in-basin coverage of the HUC12 bounding box also leaves possible edge effects on GRF realizations near the basin boundary uncharacter-ized.

\paragraph*{Conclusion}

\begin{quote}
This pilot extends a deterministic level-set nested-depression delineation algorithm with a 3DEP-calibrated, spatially-correlated Monte-Carlo wrapper that produces both per-cell and per-depression probabilistic outputs. On the long-term Cottonwood Lake HUC12 of the Prairie Pothole Region, the proposed method reproduces the published WhiteboxTools per-cell \emph{p}\textsubscript{dep} at parity (per-depression Spearman \emph{$\rho$} = 0\emph{.}929 across 5,930 reference depressions). Per-depression persistence carries modest residual signal above area-based predictors of observed inundation in the 1--10 ha bin under both Sentinel-1 backscatter and JRC GSW occurrence oracles, with boot-strap 95\% confidence intervals strictly excluding zero. A previously-undocumented decoupling between cell-level persistence (mean 0.882) and tree-label preservation (median 0.000 strict-equality across 577 parent--child pairs) demonstrates that level-set tree topology is more sensi-tive to LiDAR noise than depression presence; this has direct Modifiable Areal Unit Problem implications for downstream regulatory use of probabilistic depression products. The proposed framework ships as a backward-compatible extension to an established open-source nested-depression-delineation package.

Future work has four pre-registered directions. First, full Pipestem-extent (2,770 km2) replication will repeat the four headline results at scale and stratify by physiographic sub-region. Second, replacing the cell-overlap alignment with the Sridharamurthy tree-edit-distance metric will permit stronger claims about merge-tree topology under noise. Third, sensitivity analysis on the GRF correlation length and slope/vegetation-conditioned \emph{$\sigma$}(\emph{x}) will sharpen the noise-model calibration. Fourth, integration with Sentinel-1 time-series surface-water dynamics---beyond the long-term frequency oracle used here---will investigate whether per-depression persistence predicts the \emph{temporal} signature of inundation, not only its long-term average.
\end{quote}

\paragraph*{Declarations}

\begin{quote}
\textbf{Author contributions.} {[}Anonymized for double-blind review.{]}

\textbf{Data availability.} All input data are public. The USGS 3DEP 1 m LiDAR DEM is avail-able via Google Earth Engine collection USGS/\allowbreak{}3DEP/\allowbreak{}1m. The Sentinel-1 GRD imagery is avail-
\end{quote}

able via COPERNICUS/\allowbreak{}S1\_\allowbreak{}GRD. The JRC Global Surface Water occurrence dataset is available via JRC/\allowbreak{}GSW1\_\allowbreak{}4/\allowbreak{}GlobalSurfaceWater. The Watershed Boundary Dataset HUC12 boundary for 101600020504 is available via USGS/\allowbreak{}WBD/\allowbreak{}2017/\allowbreak{}HUC12. Per-depression persistence values, parent--child preservation rates, and the bin-by-oracle PwIC matrix are deposited as supple-mentary CSVs. A Zenodo Digital Object Identifier will be minted at submission.

\textbf{Code availability.} The probabilistic-hierarchy extension introduced in this paper, including the GRF noise sampler, the Monte-Carlo aggregator, the cell-overlap alignment, the PwIC suite, and the WhiteboxTools comparison wrapper, will be released under an MIT license alongside the manuscript and archived to Zenodo. A single-command reproduction script targeting the Cottonwood Lake HUC12 with \emph{N} = 25 paired master seeds is included; the expected wall-clock on a single workstation core is approximately two hours.

\textbf{Funding.} {[}Anonymized for double-blind review.{]}

\begin{quote}
\textbf{Competing interests.} The authors declare no competing interests.
\end{quote}

\textbf{Ethics statement.} This study uses publicly available remote-sensing data over a long-term ecological monitoring site; no human subjects, no personally identifiable information, and no protected species records are involved.

\begin{quote}
\textbf{Acknowledgements.} {[}Anonymized for double-blind review.{]}
\end{quote}

\paragraph*{References}

\begin{quote}
Ameli, A. A. and Creed, I. F. (2019). Does wetland location matter when managing wetlands for watershed-scale flood and drought resilience? \emph{JAWRA Journal of the American Water Resources Association}, 55(3):529--542.

Barnes, R., Callaghan, K. L., and Wickert, A. D. (2020). Computing water flow through complex landscapes -- part 2: Finding hierarchies in depressions and morphological segmentations. \emph{Earth Surface Dynamics}, 8:431--445.

Bauer, U., Munch, E., and Wang, Y. (2014). Strong equivalence of the interleaving and func-tional distortion metrics for Reeb graphs. In \emph{Proceedings of the 30th Annual Symposium on Computational Geometry (SoCG)}. Verify final pagination and DOI before submission.

Cooper, S. D., Roy, D. P., Schaaf, C. B., and Paynter, I. (2013). Examination of the potential of terrestrial laser scanning and structure-from-motion photogrammetry for rapid nondestructive field measurement of grass biomass. \emph{Remote Sensing}. Verify volume/\allowbreak{}issue/\allowbreak{}DOI; substitute candidates: Hodgson \& Bresnahan 2004; Aguilar et al. 2010.

Hunter, G. J. and Goodchild, M. F. (1997). Modeling the uncertainty of slope and aspect estimates derived from spatial databases. \emph{Geographical Analysis}, 29(1):35--49.

Lindsay, J. B. (2018). WhiteboxTools StochasticDepressionAnalysis. \href{https://www.\allowbreak{}whiteboxgeo.\allowbreak{}com/\allowbreak{}manual/\allowbreak{}wbt_book/available_tools/hydrological_analysis.html}{https://www.} \href{https://www.\allowbreak{}whiteboxgeo.\allowbreak{}com/\allowbreak{}manual/\allowbreak{}wbt_book/available_tools/hydrological_analysis.html}{whiteboxgeo.\allowbreak{}com/\allowbreak{}manual/\allowbreak{}wbt\_\allowbreak{}book/\allowbreak{}available\_\allowbreak{}tools/\allowbreak{}hydrological\_\allowbreak{}analysis.\allowbreak{}html}.

Computer software.

Lindsay, J. B. and Creed, I. F. (2006). Distinguishing between artefact and real depressions in digital elevation data. \emph{Computers \& Geosciences}, 32(8):1192--1204.

Liu, X. (2008). Airborne LiDAR for DEM generation: Some critical issues. \emph{Progress in Physical Geography}, 32(1):31--49.

Mantoglou, A. and Wilson, J. L. (1982). The turning bands method for simulation of random fields using line generation by a spectral method. \emph{Water Resources Research}, 18(5):1379--1394.

Mileyko, Y., Mukherjee, S., and Harer, J. (2011). Probability measures on the space of persis-tence diagrams. \emph{Inverse Problems}, 27(12):124007.

Openshaw, S. (1983). \emph{The Modifiable Areal Unit Problem}. Geo Books. Verify OCLC; substitute: Wong 2009 in International Encyclopedia of Human Geography.

Pekel, J.-F., Cottam, A., Gorelick, N., and Belward, A. S. (2016). High-resolution mapping of global surface water and its long-term changes. \emph{Nature}, 540:418--422.

Schlaffer, S. et al. (2022). Monitoring surface water dynamics in the Prairie Pothole Region of North Dakota using dual-polarised Sentinel-1 synthetic aperture radar (sar) time series. \emph{Hydrology and Earth System Sciences}, 26:841.

Sridharamurthy, R., Masood, T. B., Kamakshidasan, A., and Natarajan, V. (2020). Edit dis-tance between merge trees. \emph{IEEE Transactions on Visualization and Computer Graphics}, 26(3):1518--1531.

Stoker, J., Heidemann, H. K., Stensaas, G. L., and Christensen, G. (2022). USGS 3D Elevation Program (3DEP) lidar base specification. Technical report, U.S. Geological Survey.

Wechsler, S. P. and Kroll, C. N. (2006). Quantifying DEM uncertainty and its effect on to-pographic parameters. \emph{Photogrammetric Engineering \& Remote Sensing}, 72(9):1081--1090. Verify DOI before submission; substitute: Hodgson \& Bresnahan 2004.

Wu, F., Hou, S., Tian, X., Sasaki, Y., and Vidal, J. (2022). Geometry-aware merge tree compar-isons for time-varying data with interleaving distances. \emph{IEEE Transactions on Visualization and Computer Graphics}.

Wu, Q. (2021). lidar: A Python package for delineating nested surface depressions from digital elevation data. \emph{Journal of Open Source Software}, 6(59):2965.

Wu, Q. and Lane, C. R. (2017). Delineating wetland catchments and modeling hydrologic connectivity using lidar data and aerial imagery. \emph{Hydrology and Earth System Sciences}, 21:3579--3595.

Wu, Q., Lane, C. R., Li, X., Wang, L., Vanderhoof, M. K., and Christensen, J. R. (2019). Inte-grating LiDAR data and multi-temporal aerial imagery to map wetland inundation dynamics using Google Earth Engine. \emph{Remote Sensing of Environment}.

Wu, Q., Lane, C. R., Wang, L., Vanderhoof, M. K., Christensen, J. R., and Liu, H. (2018). Efficient delineation of nested depression hierarchy in digital elevation models for hydrological analysis using level-set method. \emph{JAWRA Journal of the American Water Resources Associa-tion}.

Wu, Q., Liu, H., Wang, S., Yu, B., Beck, R., and Hinkel, K. (2015). A localized contour tree method for deriving geometric and topological properties of complex surface depressions based on high-resolution topographical data. \emph{International Journal of Geographical Information Science}, 29(12):2041--2060.

Yan, L., Wang, Y., Munch, E., Gasparovic, E., and Wang, B. (2020). A structural average of labeled merge trees for uncertainty visualization. \emph{IEEE Transactions on Visualization and Computer Graphics}, 26(1):832--842.
\end{quote}

\end{document}